\pdfoutput=1

\documentclass[11pt]{article}

\usepackage[preprint]{acl}
\usepackage{xurl}
\usepackage{longtable}
\usepackage{booktabs}

\usepackage{array}
\usepackage{times}
\usepackage{latexsym}
\usepackage{algorithm}
\usepackage{algpseudocode}
\usepackage{tcolorbox}
\usepackage{multirow}
\usepackage{booktabs}
\usepackage{makecell}
\usepackage{bbold}
\usepackage{array}
\usepackage{booktabs}
\usepackage{subcaption}
\usepackage[T1]{fontenc}

\usepackage[utf8]{inputenc}
\usepackage{amsfonts}

\usepackage{microtype}

\usepackage{inconsolata}

\usepackage{graphicx}

\usepackage{enumitem}

\usepackage{amsmath}
\usepackage{multicol}
\usepackage{amssymb}
\usepackage[normalem]{ulem}
\setlist[itemize]{itemsep=1pt, topsep=2pt, partopsep=0pt, parsep=0pt}
\setlist[enumerate]{itemsep=1pt, topsep=2pt, partopsep=0pt, parsep=0pt}

\title{Mechanism Shift During Post-training from Autoregressive to Masked Diffusion Language Models} 

\author{
  \textbf{Injin Kong}$^{*}$ \quad
  \textbf{Hyoungjoon Lee}$^{*,\S}$ \quad
  \textbf{Yohan Jo}$^{\dag}$ \\
  Graduate School of Data Science, Seoul National University \\
  \texttt{mtkong77@snu.ac.kr, hjoon721@snu.ac.kr, yohan.jo@snu.ac.kr}
}

\begin{document}
\maketitle

\def\thefootnote{\fnsymbol{footnote}}
\footnotetext[1]{Equal contribution.}
\footnotetext[4]{Visiting intern from the Department of Biosystems \& Biomaterials Science and Engineering, Seoul National University.}
\footnotetext[2]{Corresponding author.}
\def\thefootnote{\arabic{footnote}}

\begin{abstract} 
Post-training pretrained autoregressive models (ARMs) into masked diffusion models (MDMs) provides an efficient route to diffusion language modeling, but it remains unclear whether the resulting models reuse inherited autoregressive computation or reorganize it for non-autoregressive generation. We compare two 7B ARM--MDM families across four controlled diagnostic tasks and find a task-dependent mechanism shift. On prefix-dominant tasks, MDMs largely preserve inherited high-attribution pathways or exhibit only modest changes in where computation occurs. On globally constrained tasks, the reorganization is substantially stronger, with task-relevant computation shifting toward earlier layers. This depth-wise pattern persists across prompt resampling, circuit budgets, and tested inference budgets, while targeted ablations support the functional importance of the identified structures under the tested intervention protocols. At the component level, diagnostic probes suggest that ARMs rely more strongly on sharply specialized components, whereas MDMs exhibit weaker single-component specialization and more diffuse output-space alignment. Together, these results suggest that diffusion post-training selectively preserves or reorganizes inherited computation according to task structure, rather than uniformly replacing autoregressive mechanisms. 
\end{abstract}

\section{Introduction}
\label{sec:introduction}

Large language models have achieved near-human performance across diverse linguistic tasks \cite{touvron2023llama2openfoundation, qwen2025qwen25technicalreport}. Yet the autoregressive framework imposes structural limitations \cite{welleck2019neuraltextgenerationunlikelihood}: causal masking cannot revise earlier tokens, so early errors propagate throughout the sequence \cite{gu2018nonautoregressive, NIPS2015_e995f98d, ranzato2016sequenceleveltrainingrecurrent}. Moreover, many reasoning and planning tasks require global reasoning, where early decisions must satisfy constraints on the entire sequence \cite{ye2025beyond}.

Masked diffusion models (MDMs) have gained interest as a non-autoregressive paradigm well-suited to overcoming these limitations \cite{austin2021structured,sahoo2024simple}, but training them from scratch remains computationally expensive due to slower convergence \cite{gong2025scaling}. To mitigate this cost, recent work post-trains pretrained autoregressive models (ARMs) into the diffusion paradigm \cite{gong2025scaling,ye2025dream7bdiffusionlarge}, with models such as Dream \cite{ye2025dream7bdiffusionlarge} achieving strong performance at a fraction of the compute needed for training from scratch.

Despite the empirical success of post-training ARMs with diffusion objectives \cite{gong2025scaling}, the resulting algorithmic changes remain poorly understood. Since adopting a new training objective does not by itself guarantee a corresponding internal mechanism \cite{related10}, a central mechanistic question arises: does diffusion post-training uniformly preserve inherited autoregressive pathways, or does it selectively rewire them depending on the structural demands of the task?

In this work, we address this question through circuit-level analysis \cite{bhaskar2024finding} of ARMs and MDMs post-trained from the same autoregressive backbones. We first investigate \textbf{where} algorithmic changes occur by comparing task-specific circuit structures across paired ARM--MDM models and then examine \textbf{how} these changes are realized through component-wise logit-lens probes and neuron-level activation analyses.

These considerations motivate two empirical questions. First, when an ARM is post-trained into an MDM, which inherited high-attribution pathways remain useful and which are reorganized? Second, when reorganization occurs, does it appear primarily in pathway identity, depth-wise localization, or the diagnostic organization of task-critical components? We study these questions in four controlled settings where the target behavior and clean--corrupted contrast are sufficiently well defined for circuit analysis. Our conclusions are therefore scoped to these tasks and two 7B ARM--MDM families, rather than to open-ended generation or diffusion language models generally.

\section{Related Works}
\label{sec:related_works}

\subsection{Masked Diffusion Models}

MDMs generate text by reversing a corruption process that stochastically replaces tokens with a [MASK] symbol \cite{chang2022maskgit, austin2021structured}, enabling non-autoregressive generation with full bidirectional context. While training such models from scratch \cite{nie2025largelanguagediffusionmodels} is computationally expensive, recent work has shown that pretrained ARMs can be effectively post-trained into MDMs \cite{gong2025scaling}. Rather than learning diffusion dynamics from scratch, these approaches initialize from a pretrained ARM and post-train it to iteratively denoise partially masked inputs.

From DiffuLLaMA \cite{gong2025scaling} to Dream \cite{ye2025dream7bdiffusionlarge}, this line of work shows that such post-training retains the practical advantages of diffusion—parallel decoding, iterative refinement, and bidirectional attention—while reducing training cost, and achieves strong performance on directionality-sensitive tasks. However, these works focus on performance and efficiency, leaving open the question of how diffusion objectives reshape the underlying computational mechanisms.

\subsection{Mechanistic Interpretability and Circuits}

Mechanistic interpretability aims to identify the internal components and algorithms responsible for specific model behaviors \cite{related1, related2}. 
A central concept is the \textbf{circuit}, defined as a subgraph of the computational graph connecting inputs to the unembedding projection that is sufficient to produce a target behavior \cite{related1, bhaskar2024finding}. 
Nodes correspond to components such as attention heads and MLPs, while directed edges represent causal dependencies between components, where the output of one node contributes to the input of another \cite{related12, hanna2024have}.

Empirical studies show that behaviors can be explained by sparse circuits involving a small fraction of model connections \cite{bhaskar2024finding, related4}. 
Methods such as Edge Attribution Patching identify these subgraphs via gradient-based attribution, enabling circuit discovery for tasks including indirect object identification and numerical comparison \cite{related3, related7, bhaskar2024finding}. 
Across models and scales, similar circuits, such as induction heads recur consistently, suggesting that they implement stable algorithmic functions rather than incidental patterns \cite{related10, related11, related12, related13}. 
Recent automated approaches, including ACDC and EAP, enable scalable circuit discovery without manual inspection \cite{related8, bhaskar2024finding}.

\section{Method}
\label{sec:method}

\begin{figure*}[t]
  \includegraphics[width=0.75\linewidth]{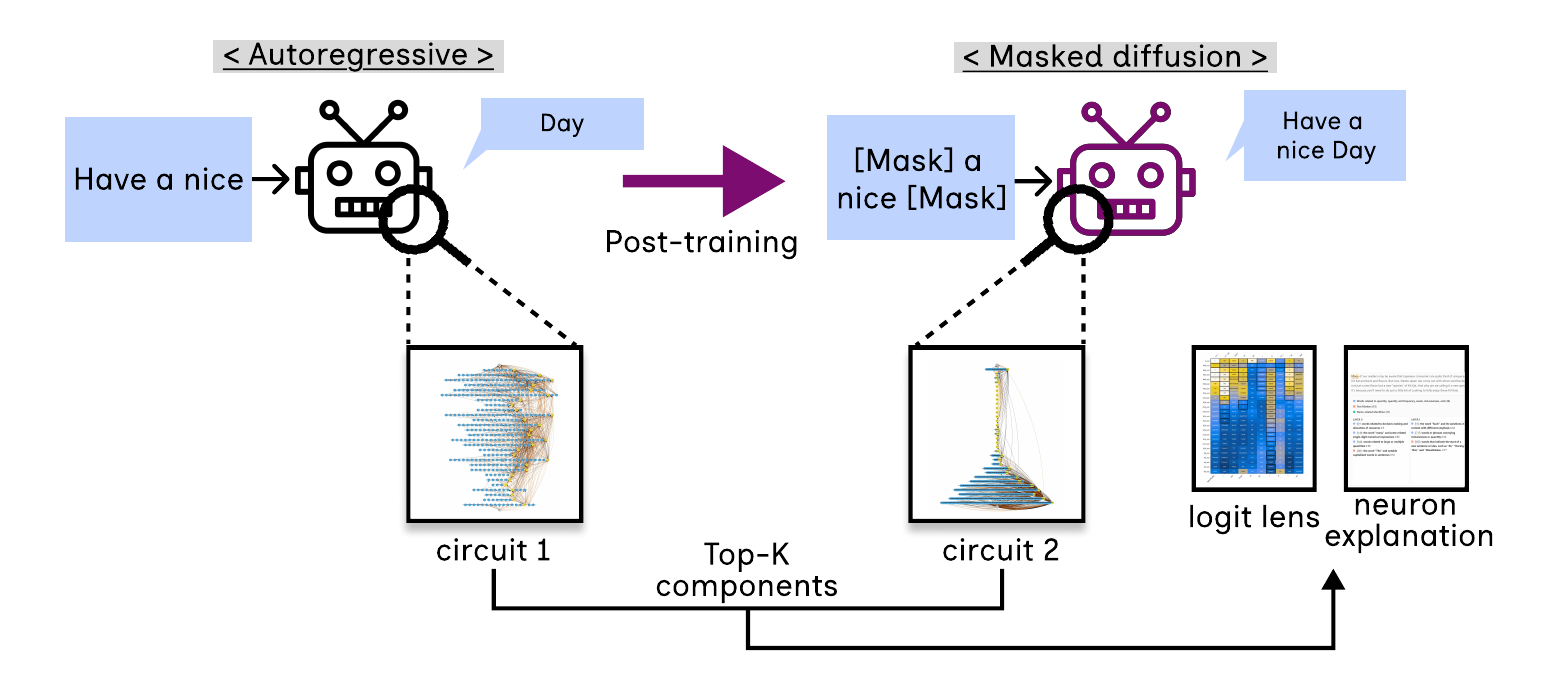}
  \centering
\caption{\textbf{Overview of the top-attribution-subgraph analysis pipeline.} 
We extract equal-budget EAP-IG subgraphs for each ARM and post-trained MDM, compare their retained edges, components, and depth-wise attribution, and analyze high-attribution components using diagnostic logit-lens and neuron-level probes. The retained subgraphs are sparse attribution summaries rather than complete task circuits.}
    \label{fig:mechanism-shift}
\end{figure*}

Figure~\ref{fig:mechanism-shift} illustrates our comparative framework for investigating the mechanistic shift from ARMs to MDMs. By comparing pretrained ARMs with MDMs post-trained from the same autoregressive backbones, we reduce confounds from model scale, initialization, and base architecture. This within-family comparison allows us to characterize computational changes associated with the overall ARM-to-MDM adaptation process. Because the diffusion objective, denoising interface, post-training data, and optimization trajectory are jointly changed during this process, we interpret the observed differences as effects of holistic adaptation rather than attributing them to any single factor.

\subsection{Models and Configuration}

To test whether the qualitative pattern replicates across model families, we evaluate Qwen2.5-7B~\cite{qwen2025qwen25technicalreport}/Dream-Base-7B~\cite{ye2025dream7bdiffusionlarge} and LLaMA-2-7B~\cite{touvron2023llama2openfoundation}/DiffuLLaMA-7B~\cite{gong2025scaling}. Comparisons within each pair control model architecture and initialization. Comparisons of effect magnitude across the two families remain confounded by backbone capability, pretraining history, post-training data, and diffusion-training recipe. We therefore treat cross-family magnitude differences descriptively and base our primary claim on the within-pair pattern reproduced in both families.

\subsection{Tasks and Datasets}

To examine whether circuit reorganization depends on task structure, we distinguish between causal and global regimes. In causal tasks, the target can largely be inferred from left-to-right prefix context, whereas in global tasks it depends on constraints across the full sequence. To operationalize this distinction beyond informal task descriptions, we use an MDM-based conditional-entropy gap, $\Delta H := H_{\mathrm{MDM}}(x_t \mid x_{<t}) - H_{\mathrm{MDM}}(x_t \mid x_{\setminus t})$, where $x_{\setminus t}$ denotes the sequence with $x_t$ masked. Positive $\Delta H$ indicates a global regime, while non-positive $\Delta H$ indicates a causal or prefix-dominant regime. We compute the primary grouping with Dream and verify its robustness with LLaDA, which recovers the same sign-based grouping across all four tasks. This entropy criterion is used only to operationalize task grouping and does not enter circuit extraction, overlap, CoG, component-level probes, or ablation analyses. Details are provided in Appendix~\ref{sec:task-dataset}.

\paragraph{Behavioral coverage and shared-success protocol.}
For each task and ARM--MDM pair $p=(A_p,M_p)$, we first evaluate both models on the complete task set $\mathcal{D}$ and identify the pair-specific common-correct pool $\{x\in\mathcal{D}:A_p(x)=y(x)\land M_p(x)=y(x)\}$. We then select 500 examples from this pool to form $\mathcal{D}^{(p)}_{\mathrm{circuit}}$, where $y(x)$ denotes the task-defined correct output. All EAP-IG attribution, edge-overlap, component-overlap, Center-of-Gravity (CoG), and downstream component-level analyses are computed exclusively on $\mathcal{D}^{(p)}_{\mathrm{circuit}}$. Consequently, these metrics characterize successful computation on the pair-specific shared-success distribution; they should not be interpreted as describing failure cases or the models' computation over the complete task distribution.

\begin{table}[t]
\centering
\small
\setlength{\tabcolsep}{4pt}
\begin{tabular}{@{}lccc@{}}
\toprule
Task & Pair & Acc. (ARM/MDM) & \shortstack{Common-correct\\$n$ (\%)} \\
\midrule
IOI       & Q/D  & 0.956 / 0.902 & 603 (86.1\%) \\
          & L/DL & 0.972 / 0.876 & 594 (84.9\%) \\
GT        & Q/D  & 0.962 / 0.918 & 619 (88.4\%) \\
          & L/DL & 0.943 / 0.832 & 552 (78.9\%) \\
Countdown & Q/D  & 0.236 / 0.477 & 507 (20.3\%) \\
          & L/DL & 0.253 / 0.322 & 502 (20.1\%) \\
SI        & Q/D  & 0.632 / 0.753 & 513 (51.3\%) \\
          & L/DL & 0.640 / 0.695 & 511 (51.1\%) \\
\bottomrule
\end{tabular}
\caption{
Behavioral performance and circuit-analysis coverage.
Q/D denotes Qwen2.5-7B/Dream-Base-7B and L/DL denotes
LLaMA-2-7B/DiffuLLaMA-7B.
Acc. reports full-set accuracy for ARM/MDM, respectively.
Common-correct $n$ (\%) reports the number and proportion of full-set
examples correctly solved by both models. We use 500 examples from
each common-correct pool for circuit analysis.
}
\label{tab:behavioral_coverage}
\end{table}

\paragraph{Indirect Object Identification (\textsc{IOI}):}
IOI is a canonical circuit-analysis task in which the model identifies the correct indirect object in a short narrative with repeated names~\cite{related4}. Since the answer can be resolved from preceding context, IOI tests whether inherited autoregressive pathways are preserved under diffusion post-training.

\paragraph{Greater-Than (\textsc{GT}):}
GT is a numerical comparison task in which the model predicts a year greater than a reference year given in the prompt~\cite{related3}. The constraint can be resolved directly from the prefix, making GT a complementary test of prefix-bound numerical reasoning.

\subsubsection{Global Tasks}

\paragraph{Countdown:}
Countdown requires constructing a valid arithmetic expression from a set of source integers to match a target value~\cite{ye2025beyond}. 
Since each intermediate choice must satisfy a final global constraint, Countdown tests the model's ability to plan beyond left-to-right generation.

\paragraph{Semantic Infilling (\textsc{SI}):}
SI requires predicting missing intermediate phrases conditioned on both preceding and subsequent text blocks. 
Because the missing span must be coherent with context on both sides, SI probes bidirectional coordination and non-sequential integration.

Generation budgets are aligned to task-specific target lengths and prediction settings; details are provided in Appendix~\ref{sec:task-dataset}.

\subsection{Circuit Discovery and Analysis Pipeline}
To trace where and how high-attribution computation changes, we compare matched top-attribution subgraphs in three stages: extraction, attribution-guided comparison, and component-level diagnostics. Extraction ranks edges by EAP-IG attribution and retains an equal-cardinality subgraph for each model. We then compare retained edge identity, influential components, and depth-wise attribution. These sparse subgraphs summarize the highest-attribution computation under a controlled edge budget; they are not assumed to recover the models' complete task circuits.

\subsubsection{Discovery}
We use Edge Attribution Patching with Integrated Gradients (EAP-IG) to score edges by their contribution to the task-defined clean--corrupted output contrast. We construct each clean--corrupted pair using the same semantic intervention for the paired ARM and MDM while preserving each model's native prediction interface. For MDMs, \texttt{[MASK]} tokens are not themselves used as the semantic corruption: the clean and corrupted examples preserve the target positions, mask configuration, denoising schedule, and diffusion-step aggregation, and differ only in the unmasked semantic context that determines the correct output. For ARMs, we apply the same semantic intervention directly to the observed context without introducing mask tokens. Exact task-specific corruption rules are provided in Appendix~\ref{sec:task-dataset}.

For each model and task, we retain the 1,000 edges with the greatest absolute attribution, yielding a top-attribution subgraph $\mathcal{S}^{(1000)}_{m,t}\subset\mathcal{G}_{m}$. The common edge count is a matched-sparsity operating point: it enables ARM--MDM comparisons at equal subgraph cardinality and prevents overlap measures from being directly confounded by different retained edge counts. We do not interpret 1,000 edges as a model-specific optimum, a minimally faithful circuit, or a complete task circuit. Sensitivity to equal budgets from 50 to 1,000 edges is reported in Appendix~\ref{app:budget-sweep}.

\subsubsection{Attribution-Guided Circuit Comparison}

\begin{table*}[!t]
\centering
\small
\begin{tabular}{lll}
\toprule
Level & Metric & Interpretation \\
\midrule
Edge  
& $J(\mathcal{E}^{\mathrm{top}}_{\mathrm{ARM}}, \mathcal{E}^{\mathrm{top}}_{\mathrm{MDM}})$  
& Pathway reuse \\
\addlinespace[0.5em] 

Component 
& $s(v)=\sum_{(u,v)\in \mathcal{E}^{\mathrm{top}}} a_{u\rightarrow v}
+\sum_{(v,u)\in \mathcal{E}^{\mathrm{top}}} a_{v\rightarrow u}$ 
& Influential source/sink components \\
\addlinespace[0.8em] 

Layer 
& $\mathrm{CoG}=\frac{\sum_l l A_l}{\sum_l A_l}$ 
& Depth-wise localization \\
\bottomrule
\end{tabular}
\caption{
Summary metrics for attribution-guided circuit comparison.
Here, \(\mathcal{E}^{\mathrm{top}}\) denotes the retained top-attribution edges,
\(a_{u\rightarrow v}\) denotes the EAP-IG attribution score of edge \(u\rightarrow v\),
and \(A_l = \sum_{v:\ell(v)=l} s(v)\) denotes the total attribution mass assigned to layer \(l\).
}
\label{tab:circuit-comparison-metrics}
\end{table*}

After extracting a matched-size top-attribution subgraph for each
model, we compare whether the MDM retains the same high-attribution
edges as its ARM, assigns attribution to different components, or
changes the depth-wise distribution of attribution. We conduct these
comparisons at the edge, component, and layer levels.

\paragraph{Edge-level overlap.}
For each ARM--MDM pair, we select the highest-attribution edges under EAP-IG~\cite{hanna2024have} on identical prompts and compute their Jaccard overlap. 
High overlap indicates pathway reuse, whereas low overlap suggests that diffusion post-training recruits different computational routes. 
We use the top 1000 edges in all experiments, balancing attribution coverage and circuit sparsity; details are provided in Appendix~\ref{sec:circuit-details}.

\paragraph{Component-level importance.}
Edge overlap tells us whether two models use similar connections, but not which components dominate the computation. 
We therefore aggregate EAP-IG attribution scores over the incoming and outgoing high-attribution edges incident to each node $v$, yielding a component score $s(v)$.
Components with large $s(v)$ repeatedly appear as sources or sinks of important information flow. 
We define the Top-$K$ components as the $K=100$ nodes with the largest scores and use them as the focus of downstream logit-lens and neuron-level analyses. 
The choice of $K$ is discussed in Appendix~\ref{sec:circuit-details}.

\paragraph{Layer-wise localization.}
Finally, to summarize where the circuit is concentrated across depth, we compute a layer-wise Center of Gravity (CoG). 
Lower CoG indicates that the attribution mass is concentrated in earlier layers, while higher CoG indicates greater reliance on middle or late layers. 
This provides a scalar summary of whether diffusion post-training preserves the depth profile of the ARM circuit or shifts computation toward earlier layers. 
Formal details are provided in Appendix~\ref{sec:metrics}.

\subsubsection{Mechanism Interpretation}

After identifying where circuits differ, we next ask how the computation carried by these changed components differs. 
We focus on the Top-$K$ EAP-IG components from the attribution-guided comparison and analyze them using two complementary probes: component-wise logit lens and neuron-level explanation.

\paragraph{Component-wise Logit Lens.}
Following the logit lens framework~\cite{nostalgebraist2020logitlens}, 
we project intermediate component activations into vocabulary space using each model's native unembedding matrix. 
Because each MDM is post-trained from the same autoregressive backbone as its paired ARM and retains the same tokenizer, 
the paired models share an identical vocabulary space. 
We compute diagnostic logits as 
\(z_{\mathrm{AR}}^{(c,t)}=(W_U^{\mathrm{AR}})^\top h_{\mathrm{component}}^{(c,t)}\) for ARMs and 
\(z_{\mathrm{MDM}}^{(c,t,s)}=(W_U^{\mathrm{MDM}})^\top h_{\mathrm{component}}^{(c,t,s)}\) for MDMs, 
where \(c\) indexes the component, \(t\) the token position, and \(s\) the diffusion timestep. 
We trace ARM alignments across token positions and MDM alignments across diffusion timesteps at fixed positions.

These projections are used only as diagnostic output-basis probes, not as calibrated predictions or basis-invariant evidence. 
Thus, they indicate which vocabulary directions a component aligns with, rather than proving that the component directly predicts those tokens.

To quantify these patterns, we compute summary statistics from component-wise logit distributions. 
For \textsc{IOI}, we measure name-token concentration with \textit{NameFrac@K}. 
Across tasks, we measure specialization using \textit{LogitGap}, $\Delta$LME, and top-$K$ entropy. 
Details are provided in Appendix~\ref{sec:metrics}.

\paragraph{Neuron Explanation.}
To complement output-space probes, we examine the internal features encoded by high-attribution components, focusing on the early layers where MDM circuits place substantial attribution mass. We define a neuron as a single scalar coordinate in the residual stream at a given transformer layer. For each task, we record activations over neuron indices and token positions in layers \(0\)--\(4\), restricted to components in the Top-\(K\) EAP-IG circuit, and collect the maximally activating input contexts for qualitative and automated explanation, following \citet{bills2023language}. For MLPs we inspect top-activating residual coordinates; for attention components we analyze head output features. Because individual coordinates are not basis-invariant, we interpret neuron explanations only at the aggregate category level rather than as definitive labels for single neurons.

Together, component-wise logit lens and neuron-level explanation provide complementary views of the mechanism shift: the former characterizes how influential components align with output vocabulary directions, while the latter examines what internal features support those alignments.

\section{Results \& Analysis}
\label{sec:Results & Analysis}

\subsection{Circuit-Level Differences}
\begin{figure*}[!t]
    \centering
    \includegraphics[width=0.24\linewidth]{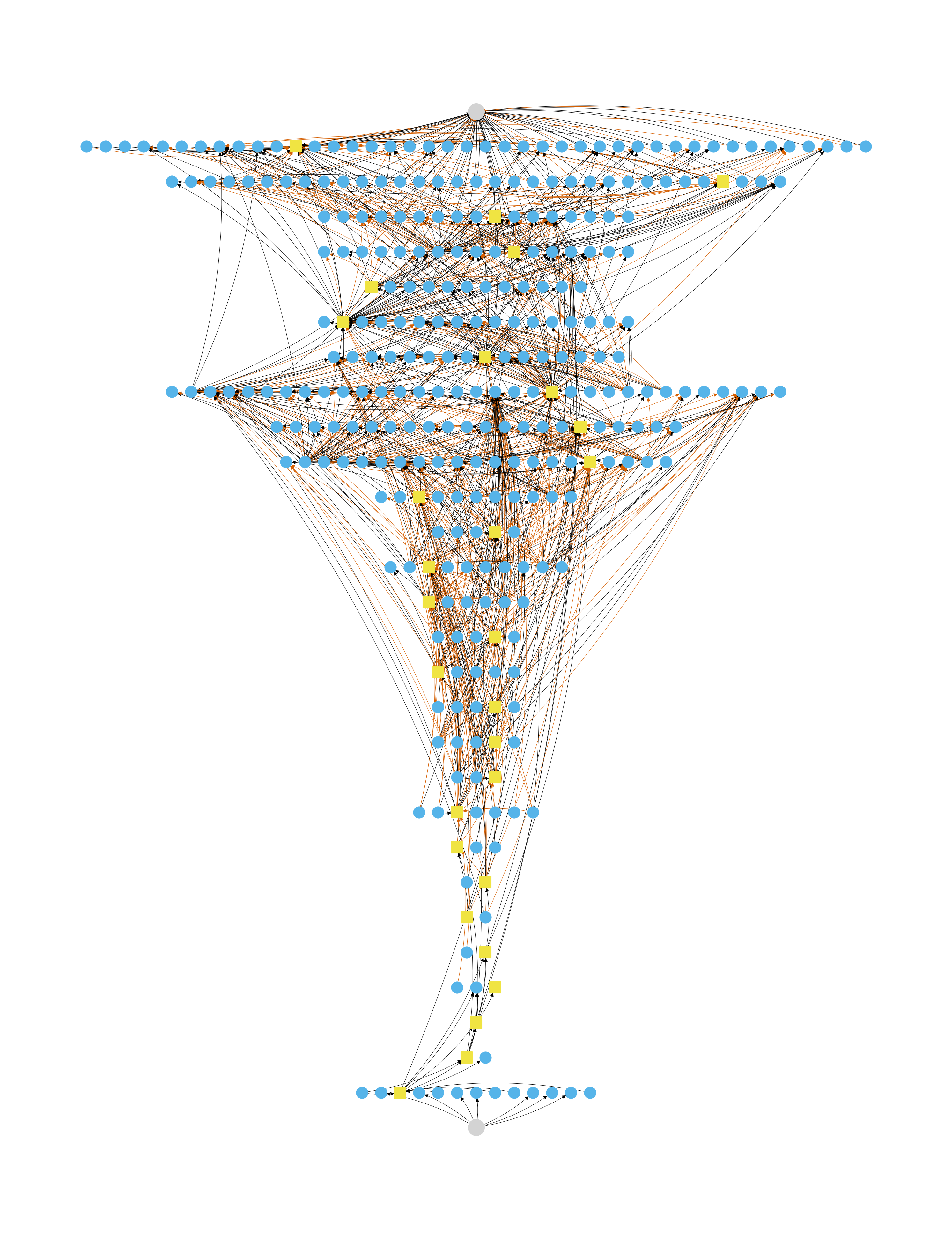}
    \hfill
    \includegraphics[width=0.24\linewidth]{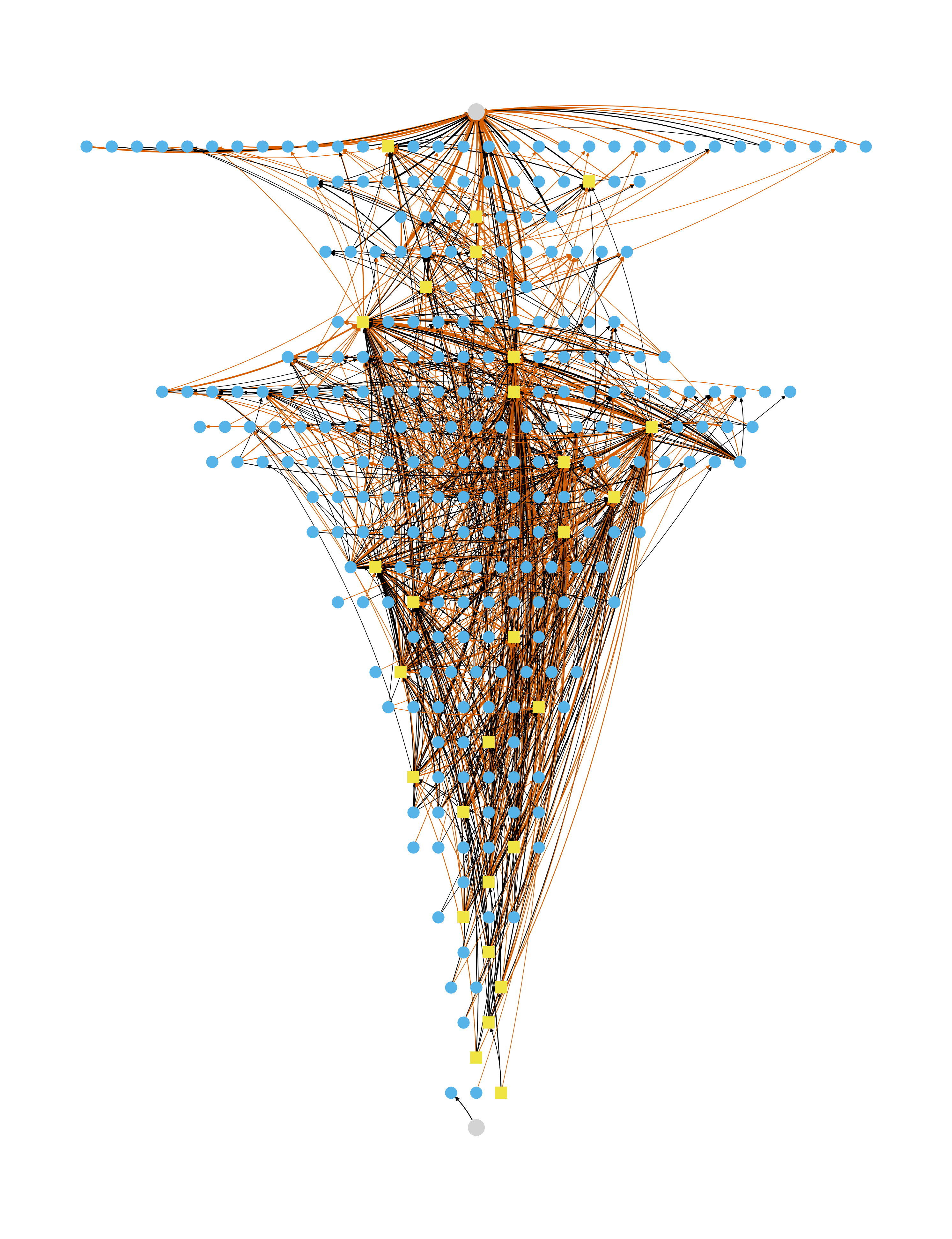}
    \hfill
    \includegraphics[width=0.24\linewidth]{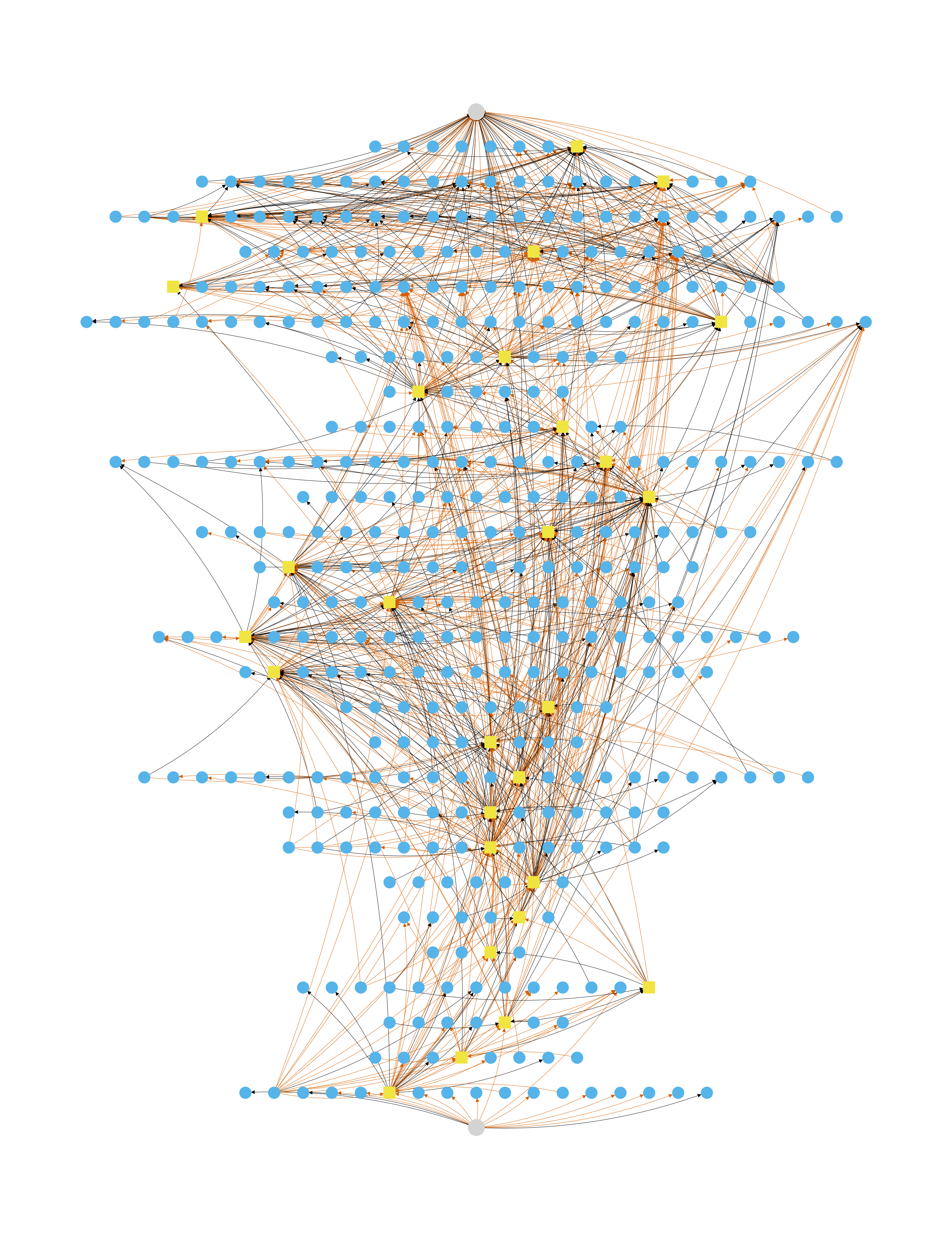}
    \hfill
    \includegraphics[width=0.24\linewidth]{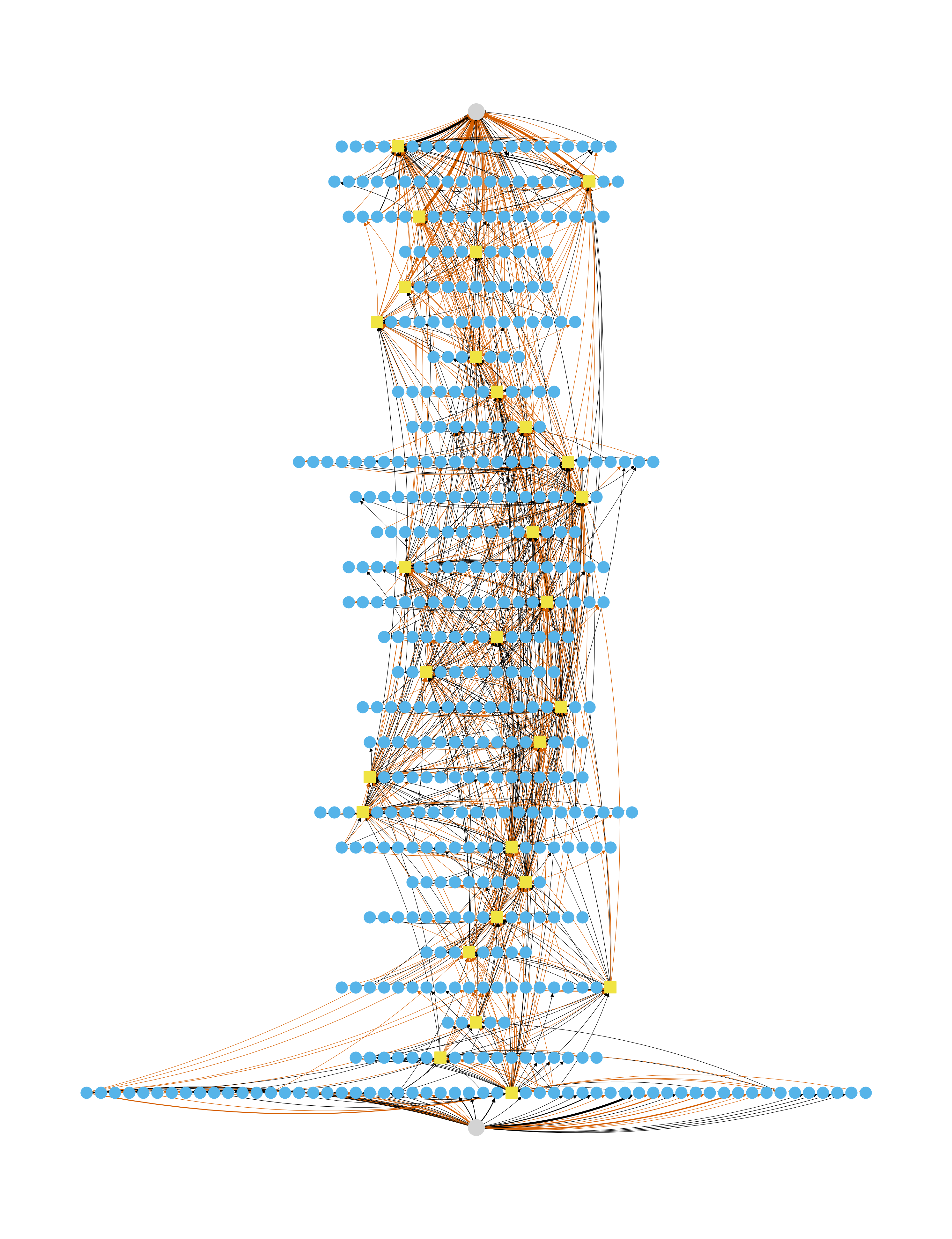}

    \caption{
    Matched top-1,000 attribution subgraphs for IOI and Countdown.
From left to right: Qwen2.5-7B and Dream-Base-7B on IOI, followed by Qwen2.5-7B and Dream-Base-7B on Countdown.
Node and edge density show the retained high-attribution structure under a common sparsity budget, not complete task circuits.
Higher-resolution versions and remaining tasks are in Appendix Figure~\ref{fig:circuit_comparison_expanded}.
    }
    \label{fig:circuit-difference}
\end{figure*}

Figure~\ref{fig:circuit-difference} shows circuits for ARMs and MDMs on \textsc{IOI} and \textsc{Countdown}. The corresponding circuit visualizations for \textsc{Greater-Than} and \textsc{Semantic Infilling} are provided in Appendix~\ref{sec:circuit-details}. Combined with the overlap and CoG metrics in Table~\ref{tab:circuit_comprehensive}, including the \textsc{Greater-Than(GT)} and \textsc{Semantic Infilling(SI)} evaluations, the results reveal a task-dependent pattern of circuit reuse and reorganization.

\begin{table}[t]
\centering
\scriptsize
\setlength{\tabcolsep}{2.5pt}
\renewcommand{\arraystretch}{1.0}

\begin{tabular}{@{}llcccc@{}}
\toprule
\multirow{2}{*}{\textbf{Task}} &
\multirow{2}{*}{\textbf{Pair}} &
\multicolumn{2}{c}{\textbf{Overlap}} &
\multicolumn{2}{c}{\textbf{CoG}} \\
\cmidrule(lr){3-4} \cmidrule(l){5-6}
& & \textbf{Edge} & \textbf{Top-$K$} & \textbf{ARM} & \textbf{MDM} \\
\midrule

\multicolumn{6}{l}{\textit{Causal regime}} \\

\multirow{2}{*}{\textsc{IOI}}
& Q/D
& \shortstack{.193 \\ {\tiny [.174,.214]}}
& \shortstack{.105 \\ {\tiny [.091,.121]}}
& \shortstack{17.5 \\ {\tiny [17.2,17.9]}}
& \shortstack{20.4 \\ {\tiny [20.0,20.9]}} \\

& L/DL
& \shortstack{.088 \\ {\tiny [.074,.104]}}
& \shortstack{.124 \\ {\tiny [.108,.143]}}
& \shortstack{18.2 \\ {\tiny [17.8,18.6]}}
& \shortstack{19.8 \\ {\tiny [19.3,20.3]}} \\

\multirow{2}{*}{\textsc{GT}}
& Q/D
& \shortstack{.290 \\ {\tiny [.268,.313]}}
& \shortstack{.515 \\ {\tiny [.489,.542]}}
& \shortstack{18.7 \\ {\tiny [18.3,19.1]}}
& \shortstack{14.4 \\ {\tiny [14.0,14.9]}} \\

& L/DL
& \shortstack{.054 \\ {\tiny [.044,.066]}}
& \shortstack{.117 \\ {\tiny [.101,.135]}}
& \shortstack{17.3 \\ {\tiny [16.9,17.8]}}
& \shortstack{11.7 \\ {\tiny [11.2,12.1]}} \\

\midrule
\multicolumn{6}{l}{\textit{Global regime}} \\

\multirow{2}{*}{\textsc{Countdown}}
& Q/D
& \shortstack{.008 \\ {\tiny [.005,.012]}}
& \shortstack{.093 \\ {\tiny [.079,.109]}}
& \shortstack{16.5 \\ {\tiny [16.1,16.9]}}
& \shortstack{4.8 \\ {\tiny [4.5,5.1]}} \\

& L/DL
& \shortstack{.032 \\ {\tiny [.025,.041]}}
& \shortstack{.081 \\ {\tiny [.067,.098]}}
& \shortstack{17.1 \\ {\tiny [16.6,17.5]}}
& \shortstack{5.3 \\ {\tiny [4.9,5.7]}} \\

\multirow{2}{*}{\textsc{SI}}
& Q/D
& \shortstack{.225 \\ {\tiny [.191,.263]}}
& \shortstack{.429 \\ {\tiny [.386,.471]}}
& \shortstack{13.7 \\ {\tiny [13.0,14.4]}}
& \shortstack{3.8 \\ {\tiny [3.4,4.3]}} \\

& L/DL
& \shortstack{.018 \\ {\tiny [.010,.031]}}
& \shortstack{.124 \\ {\tiny [.096,.158]}}
& \shortstack{13.1 \\ {\tiny [12.4,13.8]}}
& \shortstack{3.9 \\ {\tiny [3.4,4.5]}} \\

\bottomrule
\end{tabular}

\caption{Edge-set similarity and depth-wise Center of Gravity for matched top-1,000 attribution subgraphs. Values are point estimates with 95\% bootstrap confidence intervals in brackets. Higher Jaccard overlap indicates greater reuse of retained high-attribution edges, while lower CoG indicates earlier attribution concentration. Q/D denotes Qwen/Dream and L/DL denotes LLaMA/DiffuLLaMA. These metrics describe a matched-sparsity operating point rather than recovery of a complete circuit.}
\label{tab:circuit_comprehensive}
\end{table}

\paragraph{Uncertainty across prompt resampling.}
Paired bootstrap resampling preserves the qualitative depth pattern across tasks. The $\Delta$CoG intervals remain strongly negative for \textsc{Countdown} ($-11.7$ [$-12.2,-11.1$] for Q/D; $-11.8$ [$-12.4,-11.2$] for L/DL) and \textsc{SI} ($-9.9$ [$-10.8,-9.0$]; $-9.2$ [$-10.1,-8.3$]), while \textsc{IOI} shows no early-layer relocation ($+2.9$ [$+2.4,+3.5$]; $+1.6$ [$+1.0,+2.1$]). This indicates that the qualitative depth-localization pattern
remains stable across prompt resampling.

\paragraph{Depth-level sensitivity to the edge budget.}
We recomputed the comparisons using equal-size top-$n$ attribution
subgraphs for
$n\in\{50,100,\ldots,1000\}$. Across this tested range, Countdown and
Semantic Infilling retain substantially earlier MDM localization in
both model families, Greater-Than retains a smaller earlier shift, and
IOI shows no consistent early-layer MDM relocation. The exact retained
edges and numerical Jaccard values vary with $n$, particularly when
overlap is close to zero. We therefore interpret this sweep as evidence
of depth-level robustness within the tested 50--1,000-edge range, not
as threshold invariance of edge identity. We make no robustness claim
for budgets above 1,000, below 50, or attribution-mass-based
thresholds. Moreover, stability of the depth profile does not imply
that any retained subgraph constitutes a complete task circuit. Full
results are reported in Appendix~\ref{app:budget-sweep}.

For \textsc{IOI}, MDMs show substantial preservation of high-attribution structure inherited from their ARM initializations. Table~\ref{tab:circuit_comprehensive} summarizes this pattern using edge overlap and CoG. Although the overlap score is modest (e.g., 0.193 for Qwen/Dream), targeted ablation indicates that the shared edges are functionally important under the tested intervention protocol: ablating the intersection reduces \textsc{IOI} accuracy to 28.1\%, whereas ablating a random subset of non-shared edges of equal size retains $92.4 \pm 1.2$\% accuracy across five random seeds. The CoG remains relatively stable in the mid-to-late layers (17.5 vs.\ 20.4), consistent with greater reuse of inherited computation rather than a large depth-wise relocation.

In contrast, for \textsc{Countdown}, the inherited autoregressive circuitry is much less reused. Circuit overlap drops sharply (e.g., to 0.008 for Qwen/Dream), indicating structural decoupling. This change is accompanied by a strong shift in depth, with the CoG moving from layer 16.5 in the ARM to 4.8 in the MDM. This depth shift is also consistent with the aggregated layer-wise component usage statistics, which show increased concentration in earlier layers (Appendix~\ref{sec:circuit-details} Figure~\ref{fig:circuit-example}).

The complementary \textsc{GT} and \textsc{SI} results corroborate this regime-level dichotomy while revealing variation across model families. \textsc{GT}, like \textsc{IOI}, remains closer to the causal regime than the global tasks, through milder depth shifts, higher overlap, or both. \textsc{SI}, like \textsc{Countdown}, exhibits a strong shift toward earlier layers, consistent with greater reliance on bidirectional context. Within this regime-level pattern, the magnitude of reorganization varies across model families: Dream tends to preserve more of its Qwen backbone on causal tasks, whereas DiffuLLaMA shows greater structural decoupling from its LLaMA backbone across regimes. Nevertheless, both model families exhibit the same qualitative split, with greater preservation on causal tasks and stronger early-layer relocation on global tasks. Because the two pairs differ in backbone capability, pretraining history, post-training data, and diffusion-training recipe, we interpret these cross-family differences descriptively rather than attributing them to any single factor.

Taken together, these findings show that masked diffusion post-training does not simply overwrite autoregressive mechanisms. Instead, it preserves inherited circuitry where it remains task-compatible, while recruiting early-layer components when global or bidirectional reasoning requires stronger non-sequential integration.

\subsection{Semantic Reorganization of Task-Critical Components and Early Layers}

\begin{table}[t]
\centering
\scriptsize
\setlength{\tabcolsep}{3pt}
\renewcommand{\arraystretch}{0.95}
\resizebox{0.94\columnwidth}{!}{%
\begin{tabular}{@{}lrrrr@{}}
\toprule
\textbf{Category} & \textbf{Dream} & \textbf{DiffuLLaMA} & \textbf{Qwen} & \textbf{LLaMA} \\
\midrule
PERSON   & 1,300 & 337 & 298 & 528 \\
OTHER    & 8,348 & 3,251 & 2,965 & 14,606 \\
ORG      & 62 & 30 & 81 & 418 \\
LOC      & 55 & 33 & 31 & 343 \\
DIGIT    & 32 & -- & 2 & 55 \\
OPERATOR & 20 & 2 & 210 & 124 \\
MISC     & 3 & 2 & 2 & -- \\
\midrule
\textbf{Total} & \textbf{9,820} & \textbf{3,933} & \textbf{3,775} & \textbf{15,619} \\
\midrule
\textbf{NameFrac@10} & \textbf{13.2\%} & \textbf{8.6\%} & \textbf{7.9\%} & \textbf{3.4\%} \\
\bottomrule
\end{tabular}%
}
\caption{NER category distribution among the top-10 aligned tokens ($T_{10}$) for each model. NameFrac@10 represents the percentage of PERSON tokens relative to the total top-10 tokens extracted.}
\label{tab:ner_distribution}
\end{table}

To understand how circuit-level differences translate into computational strategies, we analyze model behavior at two complementary granularities. First, we examine \emph{task-critical components}—attention heads and MLPs that receive high EAP-IG attribution—using component-wise logit lens analysis. This allows us to characterize how high-attribution components align with task-relevant output-space directions and how concentrated these alignments are across components. Second, we analyze \emph{early-layer neuron activations} to characterize aggregate activation patterns in regions where MDMs emphasize computation but component-level output-space alignment appears diffuse.

\subsubsection{Task-Critical Components: Component-wise Logit Lens Analysis}

\begin{table*}[t]
\centering
\footnotesize
\begin{tabular}{lllccl}
\toprule
\textbf{Task} & \textbf{Model} & \textbf{Component} & \textbf{Mean Logit} & \textbf{Top Tokens} & \textbf{Role} \\
\midrule
\textsc{IOI} &
Qwen & a27.h5 & 21.14 & Dan & Person-name-related Component \\
& Qwen & a23.h11 & 6.84 & Ben & Person-name-related Component \\
& LLaMA & a24.h15 & 2.75 & Jerry & Person-name-related Component \\
& LLaMA & a21.h1 & 1.73 & Carol & Person-name-related Component \\
& Dream & m25 & 3.14 & Browser & Proper-noun component \\
& Dream & m22 & 2.33 & Kremlin & Proper-noun component \\
& DiffuLLaMA & a26.h21 & 2.81 & Marian & Person-name-related Component \\
& DiffuLLaMA & a22.h19 & 1.80 & Grace & Person-name-related Component \\
\midrule
\textsc{Countdown} &
Qwen & m25 & 51.29 & 3 & Digit-related Component \\
& Qwen & m20 & 20.67 & 1, 2 & Broad Numerical Component \\
& LLaMA & m29 & 6.20 & pick & Instruction-related Component \\
& LLaMA & a22.h13 & 2.10 & four & Numerical--Lexical Component \\
& Dream & m27 & 26.06 & 1, 2, 3 & Broad Numerical Component \\
& Dream & m23 & 5.34 & 5, 1, 4 & Broad Numerical Component \\
& DiffuLLaMA & m31 & 12.32 & 0, 1, 2 & Broad Numerical Component \\
& DiffuLLaMA & m4 & 1.40 & -- & Symbol-related Component \\
\bottomrule
\end{tabular}
\caption{Representative components exhibiting high logit concentration across tasks and model families.
Roles are descriptive labels summarizing observed logit distribution patterns rather than definitive functional assignments.}
\label{tab:component-roles}
\end{table*}

We analyze task-critical components identified by high EAP-IG attribution using a component-wise logit lens. This allows us to test whether task-relevant information is concentrated in a few specialized components or distributed across many. We report probes of semantic alignment and dominance in the main text, and defer representative top-token examples to Tables~\ref{tab:component-roles} and~\ref{tab:logit_lens}.

\paragraph{Causal reasoning task (\textsc{IOI}).}
For \textsc{IOI}, ARMs exhibit \emph{sharply localized semantic specialization}: a small set of high-attribution components aligns decisively with person-name tokens. As shown in Table~\ref{tab:ner_distribution}, MDMs can place person-name tokens relatively often among their top aligned candidates---Dream reaches a higher \textsc{NameFrac}@10 (13.2\%) than Qwen (7.9\%) and LLaMA (3.4\%), with DiffuLLaMA at 8.6\%---but this does not imply sharper specialization. ARMs exhibit substantially stronger selective amplification once a name token appears, with median $\Delta\mathrm{LME}$ values of 0.90 for Qwen and 0.89 for LLaMA, compared with 0.31 for Dream and 0.05 for DiffuLLaMA (Appendix Table~\ref{tab:dominance_metrics}). ARMs also show larger logit gaps, while DiffuLLaMA exhibits higher entropy. Thus, raw name frequency does not imply specialization: ARMs remain dominated by a few decisive name-specific components, whereas MDMs spread name-related evidence across broader component sets.

Representative components mirror this pattern (Table~\ref{tab:component-roles}). High-attribution attention heads in Qwen and LLaMA align strongly with person-name tokens, consistent with pointer-like behavior, and DiffuLLaMA largely preserves this pattern. Dream, however, departs from it: its high-attribution components are less clearly name-specific, often aligning with non-person proper nouns or broader semantic tokens. Even when circuit topology is similar, diffusion post-training can alter output-space alignment patterns and weaken the dominance of individual name-aligned heads.

\paragraph{Global reasoning task (\textsc{Countdown}).}
A contrasting pattern emerges in \textsc{Countdown}. Whereas ARMs rely on components with \emph{strong numerical selectivity}---sharply aligned with specific digits or operators and suggesting that global planning is approximated through sequential, component-centric heuristics (Table~\ref{tab:component-roles})---MDMs dissolve this concentration. In Dream, the sharp specialization collapses entirely: instead of a single dominant numerical component, multiple components exhibit moderate responses across numerical tokens with no individual component exerting decisive control. DiffuLLaMA occupies an intermediate regime, retaining numerical associations but with reduced magnitude and increased dispersion. This redistribution is also architectural---both MDMs front-load task-relevant computation, with CoG dropping to 4.8 for Dream and 5.3 for DiffuLLaMA from mid-network ARM values (Table~\ref{tab:circuit_comprehensive}). To test whether this early-layer localization is functionally relevant beyond the logit-lens diagnostics, we perform targeted ablations on the identified structures. Ablating these early-layer structures substantially reduces \textsc{Countdown} accuracy relative to matched random and late-layer controls, providing interventional evidence that they contribute to \textsc{Countdown} performance under the tested ablation protocol (see Appendix~\ref{sec:countdown_ablation_details} for full metrics and control baselines).

\paragraph{Expanded \textsc{GT}/\textsc{SI} probes.}
The completed Top-100 component-filtered logit-lens metrics for \textsc{GT} and \textsc{SI} show the same broad pattern of reduced selective amplification in MDMs (Appendix Table~\ref{tab:gt_si_logitlens_metrics}). Qwen remains sharply concentrated on both tasks, with large \textsc{LogitGap} values on \textsc{GT} (3.107) and \textsc{SI} (3.050), whereas Dream's gaps are much smaller (0.085 and 0.042). LLaMA and DiffuLLaMA are both more diffuse than Qwen, but DiffuLLaMA remains at the low-gap, high-entropy end of the spectrum, especially on \textsc{SI} (\textsc{LogitGap} 0.076; entropy 2.263). Thus, the component-level reorganization observed on \textsc{IOI} and \textsc{Countdown} generalizes across both regimes: even when circuit overlap is nontrivial, diffusion post-training weakens single-component dominance and pushes task evidence toward broader component ensembles.

Taken together, these results suggest a component-level reorganization in the diagnostic output-space patterns induced by diffusion post-training. Whereas ARMs exhibit stronger single-component specialization, MDMs show weaker single-component dominance and more distributed output-space alignment across high-attribution components, accompanied on global tasks by a pronounced shift of attribution toward earlier layers.

\subsubsection{Early-layer Representation: Neuron-level Analysis}

While component-wise logit lens analysis captures semantic specialization among task-critical components, it does not fully explain the behavior of early-layer regions emphasized by MDM circuits. To characterize how task-relevant information is implemented there, we analyze neuron activations in the lowest transformer layers.

This analysis is especially important for \textsc{Countdown}, where the CoG shift in Table~\ref{tab:circuit_comprehensive} indicates that post-training moves substantial computation into the bottom of the network. In ARMs, early-layer neuron activation is \emph{highly task-dependent}. In \textsc{IOI}, many early-layer neurons are strongly activated by descriptive adjectives and modifier-related tokens (e.g., 2,821 neurons in LLaMA; Appendix Table~\ref{tab:neuron_explanations}), suggesting these layers encode syntactic features before resolving entity identity. In \textsc{Countdown}, this pattern shifts toward neurons associated with numerical or technical content (over 1,000 neurons). Despite this task-dependent reallocation, ARMs consistently exhibit strong concentration within specific semantic categories.

MDMs display a qualitatively different pattern. Across both tasks, the number of active neurons in early layers remains relatively stable (typically $\sim$50--200 active neurons per top category; Appendix Table~\ref{tab:neuron_explanations}), and these neurons tend to correspond to broad, genre-level cues rather than sharply defined task-specific categories. This suggests MDMs rely less on category-specific early-layer specialization and instead maintain a more uniform, task-agnostic representational regime. Combined with the CoG results, this pattern implies that MDM early layers absorb more computation without relying on sharply specialized neurons.

\subsubsection{Connecting Component-level and Neuron-level Perspectives}

Taken together, the component-level and neuron-level analyses reveal a mechanistic pattern. Component-wise logit lens characterizes how outputs are distributed across influential components, while neuron-level explanation provides aggregate evidence about early-layer representations. Because neuron labels from automated explanation are noisy, we interpret them only at the category-distribution level rather than as functional assignments for specific neurons.

In ARMs, automated explanations concentrate on a smaller set of interpretable units (266) but distribute unevenly across semantic categories ($\sigma = 369.78$ in per-category counts): a few categories, such as descriptive adjectives or numerical values, account for a large fraction of explained early-layer neurons, while most categories attract only a handful (Appendix Table~\ref{tab:neuron_explanations}). This pattern suggests dominance by a limited set of highly specialized units. MDMs instead spread interpretable labels across more units (426) with lower variance ($\sigma = 202.52$), reflecting more even coverage across semantic categories rather than concentration in a few. Both the larger unit count and the flatter category distribution indicate broader, less concentrated category-level activation patterns.

This component-level reorganization is consistent with the circuit-level evidence: the front-loaded computation observed via CoG and the reduced dominance of interpretable components in logit-lens probes reflect complementary aspects of the same pattern. These findings suggest that diffusion post-training is associated with weaker component-level specialization and more distributed diagnostic alignment, a pattern that accompanies circuit-level shift toward earlier layers.

\section{Conclusion}
\label{sec:Conclusion}

In this paper, we examined how internal computation changes when pretrained autoregressive models are post-trained into masked diffusion models. Across two ARM--MDM families and four controlled tasks, we find a task-dependent pattern: prefix-dominant tasks preserve more inherited high-attribution structure, whereas globally constrained tasks show stronger reorganization and earlier depth-wise localization. Component-level diagnostics further suggest weaker single-component dominance and more distributed output-space alignment in MDMs.

These findings support a \emph{task-compatibility hypothesis}: inherited autoregressive computation may remain useful for prefix-dominant behaviors, while stronger bidirectional or global coordination may require greater reorganization. We view this as a working explanation rather than a complete theory of ARM-to-MDM adaptation. The results also suggest that, as task-relevant alignment becomes less concentrated in individual components, mechanistic analyses may need to complement localized probes with ensemble-, interaction-, and routing-level views.

Several directions remain open. Our paired-model design captures holistic ARM-to-MDM adaptation but does not disentangle the diffusion objective, denoising interface, post-training data, and optimization trajectory. Future work should separate these factors, extend analysis to naturalistic and open-ended generation, and test settings requiring stronger non-local coordination, including verb-final languages such as Japanese, Korean, Turkish, and Basque. It will also be important to test whether the same task-dependent patterns arise in diffusion language models trained from scratch.

\newpage

\section*{Limitations}
\label{sec:limitations}
Our evaluation covers four benchmarks (IOI, GREATER-THAN, COUNTDOWN, and SEMANTIC INFILLING) that contrast causal tracking with global planning. While this controlled setup isolates the mechanism shift along a clear regime axis, whether the same patterns generalize to more open-ended linguistic tasks remains to be tested. Second, we study a holistic adaptation regime in which the diffusion objective, iterative inference schedule, and post-training data exposure are coupled; our inference-budget sensitivity analysis reduces, but does not fully eliminate, the coupling between task structure and denoising budget. Because the two model families also differ in backbone capability, pretraining history, post-training data, and diffusion-training recipe, cross-family differences in reorganization magnitude cannot be attributed to any single factor. Finally, our attribution-guided pipeline isolates the most behaviorally influential pathways but does not exhaustively characterize all task-relevant computation; scaling the analysis to richer circuits and longer reasoning chains is left to future work.

\section*{Acknowledgments}
This work was supported by the National Research Foundation of Korea (NRF) grant (RS-2024-00333484) and by the Institute of Information \& Communications Technology Planning \& Evaluation (IITP) grant (RS-2024-00338140, Development of Learning and Utilization Technology to Reflect Sustainability of Generative Language Models and Up-to-dateness over Time), both funded by the Korean government (MSIT). This work was also supported by the Institute of Information \& Communications Technology Planning \& Evaluation (IITP) grant (No. RS-2026-25551943, Military-Industry-Academia Cooperation Center (Seoul (Yongsan))) funded by the Korean government (MND), and by the National Supercomputing Center with supercomputing resources including technical support (KSC-2025-CRE-0332).

\bibliography{custom}
\clearpage
\appendix
\section{Task Details}
\label{sec:task-dataset}

\paragraph{Datasets}
We evaluate the full behavioral pools before constructing the pair-specific common-correct subsets used for circuit analysis. The full evaluation sets contain 700 examples for \textsc{IOI} and \textsc{Greater-Than}, 2,500 for \textsc{Countdown}, and 1,000 for \textsc{Semantic Infilling}; from each ARM--MDM common-correct pool, we select 500 examples for circuit analysis. For \textsc{Countdown}, LLaMA-series models use 13 diffusion steps (13 tokens), while Qwen-series models use 12 diffusion steps (12 tokens). For \textsc{IOI}, we use a single-step setting with one target token. Because the task requires predicting only a single token, additional denoising steps would repeatedly refine the same target position, making the single-step setting the most direct comparison. For \textsc{Greater-Than}, we align the diffusion budget to the two-token target continuation and average MDM attribution over 2 diffusion steps. For \textsc{Semantic Infilling}, the diffusion budget follows the tokenizer-specific target-span length: Qwen-series models use 3 diffusion steps, while LLaMA-series models use 4 diffusion steps.

\paragraph{Task Templates and Examples}
To ground our mechanistic evaluations, we detail the specific structural templates and input-output formats for each of the tasks below:

\begin{itemize}
    \item \textbf{\textsc{IOI} (Indirect Object Identification):} This task tests local, directional linguistic structures by requiring the model to identify which name is the indirect object. 
    \begin{itemize}
        \item \textit{Template:} ``[Name A] and [Name B] went to the store. [Name A] gave a toy to [Target]''
        \item \textit{Example:} ``John and Mary went to the store. John gave a toy to \textbf{Mary}''
    \end{itemize}
    
    \item \textbf{\textsc{Countdown}:} This task evaluates global planning and arithmetic constraints. The model is given a target number and a set of allowed digits, and must generate a valid mathematical expression that evaluates exactly to the target.
    \begin{itemize}
        \item \textit{Template:} ``Target: [Target]. Numbers: [Num 1], [Num 2], [Num 3]. Equation: [Expression]''
        \item \textit{Example:} ``Target: 24. Numbers: 4, 6, 8. Equation: \textbf{6 * (8 - 4)}''
    \end{itemize}
    
    \item \textbf{\textsc{GT} (Greater-Than):} This task isolates sequential and localized relational tracking, forcing the model to complete a mathematical inequality based on a context year.
    \begin{itemize}
        \item \textit{Template:} ``The movie was released in the year [Year]. It became popular in the year [Target]''
        \item \textit{Example:} ``The movie was released in the year 1975. It became popular in the year \textbf{1976}''
    \end{itemize}
    
    \item \textbf{\textsc{SI} (Semantic Infilling):} This task requires predicting missing intermediate phrases conditioned on both preceding and subsequent text blocks. Because the missing span must be coherent with context on both sides, SI probes bidirectional coordination and non-sequential integration.
    \begin{itemize}
        \item \textit{Template:} ``[Prefix] [Target span] [Suffix]''
        \item \textit{Example:} ``The chef chopped the \textbf{fresh vegetables} and added them to the boiling soup.'' (where ``fresh vegetables'' is the target span constrained by both the preceding action and the subsequent context).
    \end{itemize}
\end{itemize}

\subsection{Clean/Corrupted Intervention Details}
\label{sec:clean-corrupted-interventions}

Because EAP-IG attribution depends on the semantic corruption defining
the clean--corrupted counterfactual pair, we specify the task-specific
interventions used in our experiments. Across paired ARMs and MDMs, we
apply the same semantic intervention while preserving each model's
native prediction interface. For MDMs, clean and corrupted examples use
the same target positions, mask configuration, denoising schedule, and
timestep aggregation; only the unmasked semantic context is changed.
The \texttt{[MASK]} tokens themselves are therefore not used as the
semantic corruption.

\paragraph{\textsc{IOI}.}
We apply the standard name/role corruption by changing the name
assignment that determines the correct indirect object while preserving
the sentence structure. For ARMs, EAP-IG is applied directly to the
paired observed contexts and scored at the target name position. For
MDMs, the same target position and mask configuration are used for the
clean and corrupted inputs, with only the unmasked name/role context
changed. We use a single diffusion step for the single-token target.

\paragraph{Sensitivity to the IOI corruption.}
To assess whether the extracted structure depends on the semantic
corruption, we additionally evaluate a harder IOI intervention that
swaps the relevant name/role assignment using the other in-context
name. Under this corruption, ARM--MDM edge Jaccard is 0.136 at
Top-50 and 0.390 at Top-1000, while the ARM and MDM CoGs are 16.21
and 19.10, respectively. Thus, exact edge identity is sensitive to the
chosen counterfactual, but the depth-level conclusion remains
qualitatively unchanged: IOI still shows no pronounced early-layer
relocation in the MDM. This analysis concerns sensitivity to the
semantic corruption; sensitivity to alternative MDM masking policies
is a separate question.

\paragraph{\textsc{Greater-Than}.}
We change the reference year within the same century
(e.g., $1352 \rightarrow 1301$) while preserving the target continuation
format. For ARMs, EAP-IG is applied to the paired observed prefixes and
scored on the target continuation. For MDMs, the same target span,
mask realization, diffusion-step budget, and timestep aggregation are
used for both inputs; only the reference-year context is changed.

\paragraph{\textsc{Semantic Infilling}.}
We replace the constraining context item with a matched semantic
contrast while keeping the target infill span fixed. For ARMs, EAP-IG
is applied to the paired clean and corrupted contexts and scored on the
same target span. For MDMs, the same masked target span, mask
realization, denoising schedule, and timestep aggregation are retained,
while only the unmasked semantic context is changed.

\paragraph{\textsc{Countdown}.}
We shift the target by $+1$ and shuffle the source-number order,
retaining only solvable corrupted instances with the same output-span
length. For ARMs, EAP-IG is applied to the paired prompts and scored on
the corresponding solution-expression output. For MDMs, the same
output-span mask configuration, diffusion-step budget, and timestep
aggregation are used for the clean and corrupted inputs; only the
target and source-number context is changed.

\section{Detailed Causal Ablation Framework and Results for Countdown}
\label{sec:countdown_ablation_details}

To test the functional relevance of the front-loaded computation observed in Masked Diffusion Models (MDMs) on \textsc{Countdown}, we conduct a targeted ablation study on early-layer components (Layers 2--8). This intervention complements the projection-based logit-lens analysis by directly measuring the performance effect of ablating the identified structures.

\subsection{Experimental Setup}
We utilize the rank-based activation masks extracted via the protocol described in Section 3. Instead of merely observing activation values, we systematically ablate the top-ranked task-specific attention heads and MLP dimensions by overwriting their outputs with zero vectors during the forward pass. As a control baseline, we perform an identical ablation procedure on randomly selected dimensions and middle-to-late layer groups (Layers 16--24) across 5 independent random seeds.

\subsection{Quantitative Results}
The performance changes under the targeted interventions reveal a clear difference between the identified early-layer structures and the control ablations:
\begin{itemize}
    \item \textbf{Early-Layer Target Ablation:} Ablating the selected early-layer components in the MDM reduces final task completion accuracy from an unablated 91.2\% to \textbf{35.2\%}.
    \item \textbf{Control Ablation:} Ablating an equivalent number of randomly selected dimensions or middle-to-late layer components yields substantially smaller degradation, with accuracy remaining at \textbf{89.4\%}.
\end{itemize}

The large difference between targeted and control ablations provides interventional evidence that the selected early-layer structures make a substantial contribution to \textsc{Countdown} performance under this intervention protocol. We do not interpret this result as establishing a general causal role for early layers across global planning tasks.

\section{Experimental Details and Additional Results}
\label{sec:circuit-details}

\begin{figure}[H]
    \centering
    \includegraphics[width=1\linewidth]{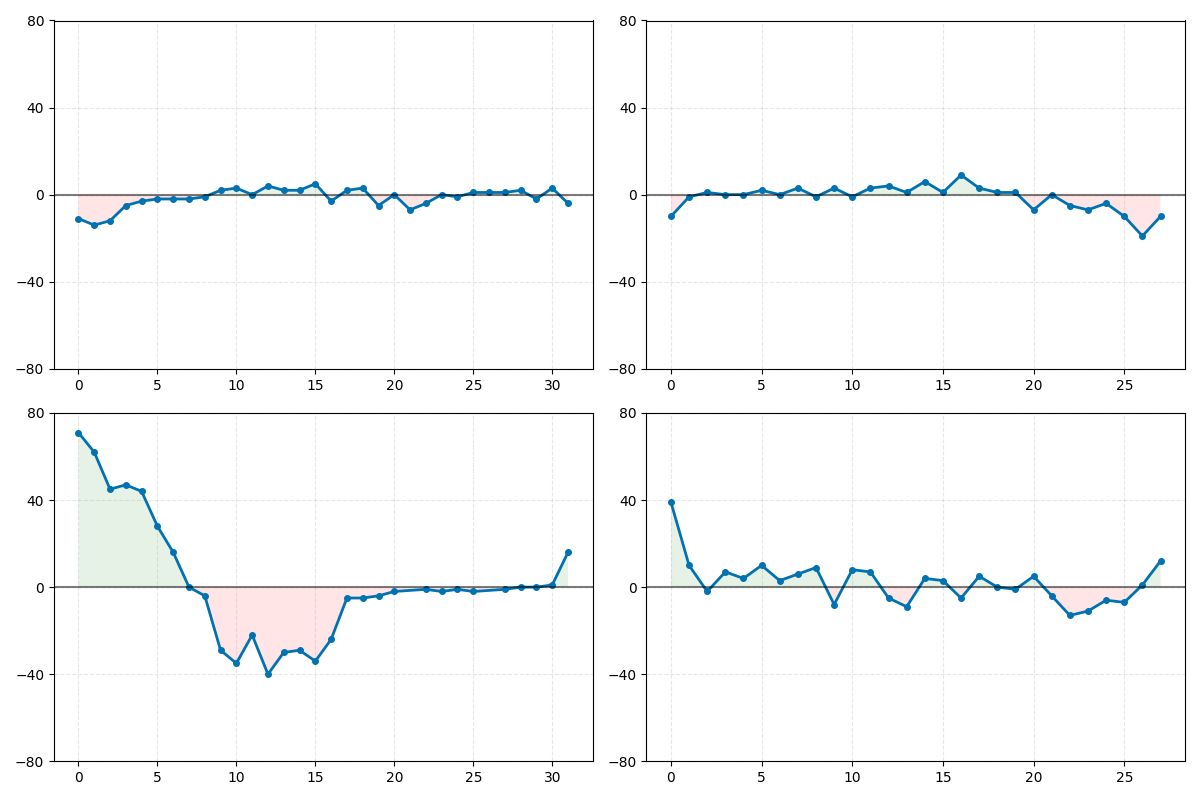}
    \caption{
    Layer-wise difference in unique attention component usage (MDM$-$ARM). Rows correspond to \textsc{IOI} and \textsc{Countdown}; columns compare DiffuLLaMA vs.\ LLaMA-2 and Dream vs.\ Qwen. Green indicates greater usage in MDMs, and red indicates greater usage in ARMs.
    }
    \label{fig:circuit-example}
\end{figure}

\paragraph{Computational Resources}
All experiments were conducted on NVIDIA A6000 GPUs. Circuit extraction and analysis required less than 10 GPU-hours per model. No additional pre-training was performed.

\paragraph{Implementation Details}
We use HuggingFace Transformers for model loading and inference. Circuit discovery is implemented with Edge Attribution Patching with Integrated Gradients (EAP-IG). All primary ARM--MDM comparisons retain the top 1,000 edges by
absolute EAP-IG attribution. This value is used as a common
matched-sparsity operating point, not as a model-specific optimum,
minimal faithful size, or estimate of a complete task circuit.
Equal-sized subgraphs ensure that Jaccard overlap is not directly
confounded by different retained edge counts. We evaluate depth-level
sensitivity over
$n\in\{50,100,\ldots,1000\}$ and do not claim robustness outside this
range. In particular, the present experiments do not evaluate
2,000-edge subgraphs, arbitrary attribution-mass thresholds, or
model-specific iso-faithfulness thresholds.

Attribution scores are then aggregated at the component level to identify the Top-K components. We set $K=100$, the smallest value for which the selected components form a well-connected and interpretable subgraph while preserving end-to-end connectivity. Importantly, these top 100 components account for more than 65\% of the total attribution mass, indicating that a relatively small subset of components captures most of the task-relevant computation. Logit-lens projections follow the standard unembedding-based formulation. All other parameters use default library settings.

\subsection{Equal-Budget Sensitivity}
\label{app:budget-sweep}

Table~\ref{tab:budget-sweep} summarizes the equal-budget sweep. Countdown and Semantic Infilling retain strong earlier MDM localization throughout the tested range in both families. Greater-Than retains a smaller earlier shift, whereas IOI exhibits no early-layer MDM relocation. Because exact edge overlap varies with the retained-edge budget, the sweep supports robustness of the qualitative depth profile rather than invariance of edge identity.

\begin{table*}[t]
\centering
\small
\setlength{\tabcolsep}{3pt}
\begin{tabular}{llrrlrrr}
\toprule
Pair & Task
& Jaccard range
& CV
& Depth trend
& $J_{1000}$
& CoG$_{\mathrm{ARM}}$
& CoG$_{\mathrm{MDM}}$ \\
\midrule
Q/D & IOI
& .193--.227 & .048
& No consistent shift
& .193 & 17.5 & 20.4 \\

Q/D & Countdown
& .008--.041 & .554
& Earlier MDM
& .008 & 16.5 & 4.8 \\

Q/D & Greater-Than
& .290--.428 & .102
& Earlier MDM (mild)
& .290 & 18.7 & 14.4 \\

Q/D & Semantic Infilling
& .176--.288 & .116
& Earlier MDM
& .225 & 13.7 & 3.8 \\

L/D & IOI
& .088--.123 & .101
& Later MDM
& .088 & 18.2 & 19.8 \\

L/D & Countdown
& .010--.032 & .395
& Earlier MDM
& .032 & 17.1 & 5.3 \\

L/D & Greater-Than
& .010--.054 & .400
& Earlier MDM (mild)
& .054 & 17.3 & 11.7 \\

L/D & Semantic Infilling
& .010--.018 & .208
& Earlier MDM
& .018 & 13.1 & 3.9 \\
\bottomrule
\end{tabular}
\caption{
Sensitivity across equal-size top-$n$ attribution subgraphs for
$n\in\{50,100,\ldots,1000\}$.
Q/D denotes Qwen/Dream and L/D denotes LLaMA/DiffuLLaMA.
The qualitative depth profile persists across matched sparsity levels,
although exact edge identity varies with the retained-edge budget.
}
\label{tab:budget-sweep}
\end{table*}

\subsection{Inference-Budget Sensitivity}
\label{app:inference-budget}

To test whether the observed depth relocation can be explained by the
number of denoising iterations, we vary only the inference budget
within \textsc{Greater-Than} and \textsc{Semantic Infilling}, while
keeping the evaluation examples and circuit-extraction procedure fixed.

\begin{table}[h]
\centering
\small
\setlength{\tabcolsep}{3pt}
\begin{tabular}{lrrrrr}
\toprule
Task & Steps & Edge Ov. & ARM CoG & MDM CoG & Gap \\
\midrule
GT & 2 & .2900 & 18.65 & 14.36 & 4.29 \\
GT & 4 & .1682 & 18.65 & 14.86 & 3.79 \\
GT & 8 & .1648 & 18.65 & 14.81 & 3.84 \\
SI & 2 & .1662 & 13.68 & 10.57 & 3.12 \\
SI & 4 & .1969 & 13.68 & 10.65 & 3.03 \\
\bottomrule
\end{tabular}
\caption{Sensitivity to the MDM inference budget. The MDM remains
earlier than the ARM at every tested budget, while the magnitude of the
depth shift does not increase monotonically with the number of
denoising steps.}
\label{tab:inference-budget}
\end{table}

Across all tested budgets, MDM CoG remains lower than ARM CoG.
Moreover, increasing the number of denoising steps does not
monotonically increase the CoG gap, and edge overlap changes in
different directions across the two tasks. Thus, inference budget
alone is insufficient to explain the observed depth relocation,
although this analysis does not establish task structure as its sole
driver.

\paragraph{Licenses and Terms of Use}
All pretrained models and tools used in this work are publicly released research artifacts. We use them solely for research and analysis in accordance with their respective licenses, and do not redistribute any models or derived data.

\begin{figure*}[!t]
    \centering
    \includegraphics[width=0.24\linewidth]{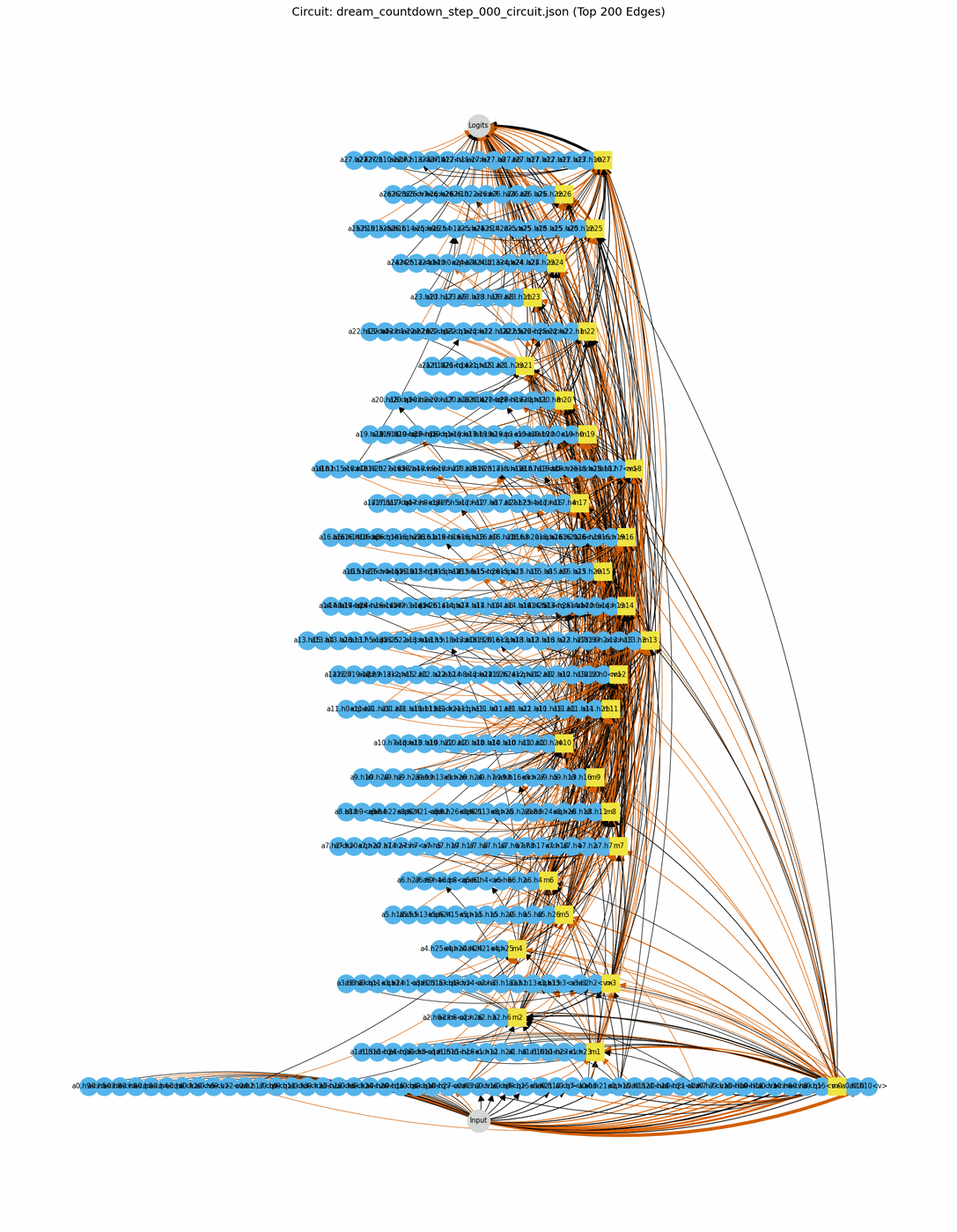}\hfill
    \includegraphics[width=0.24\linewidth]{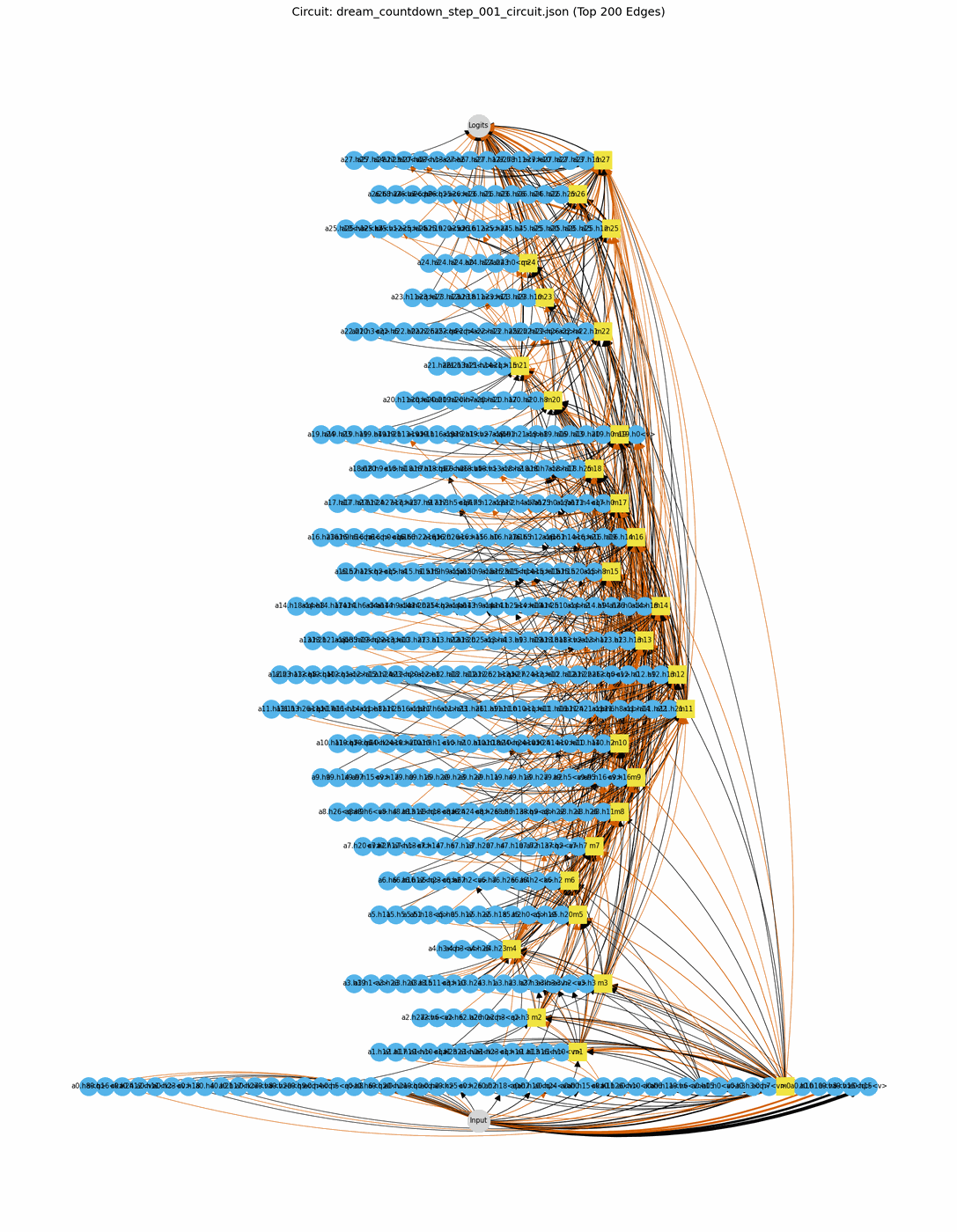}\hfill
    \includegraphics[width=0.24\linewidth]{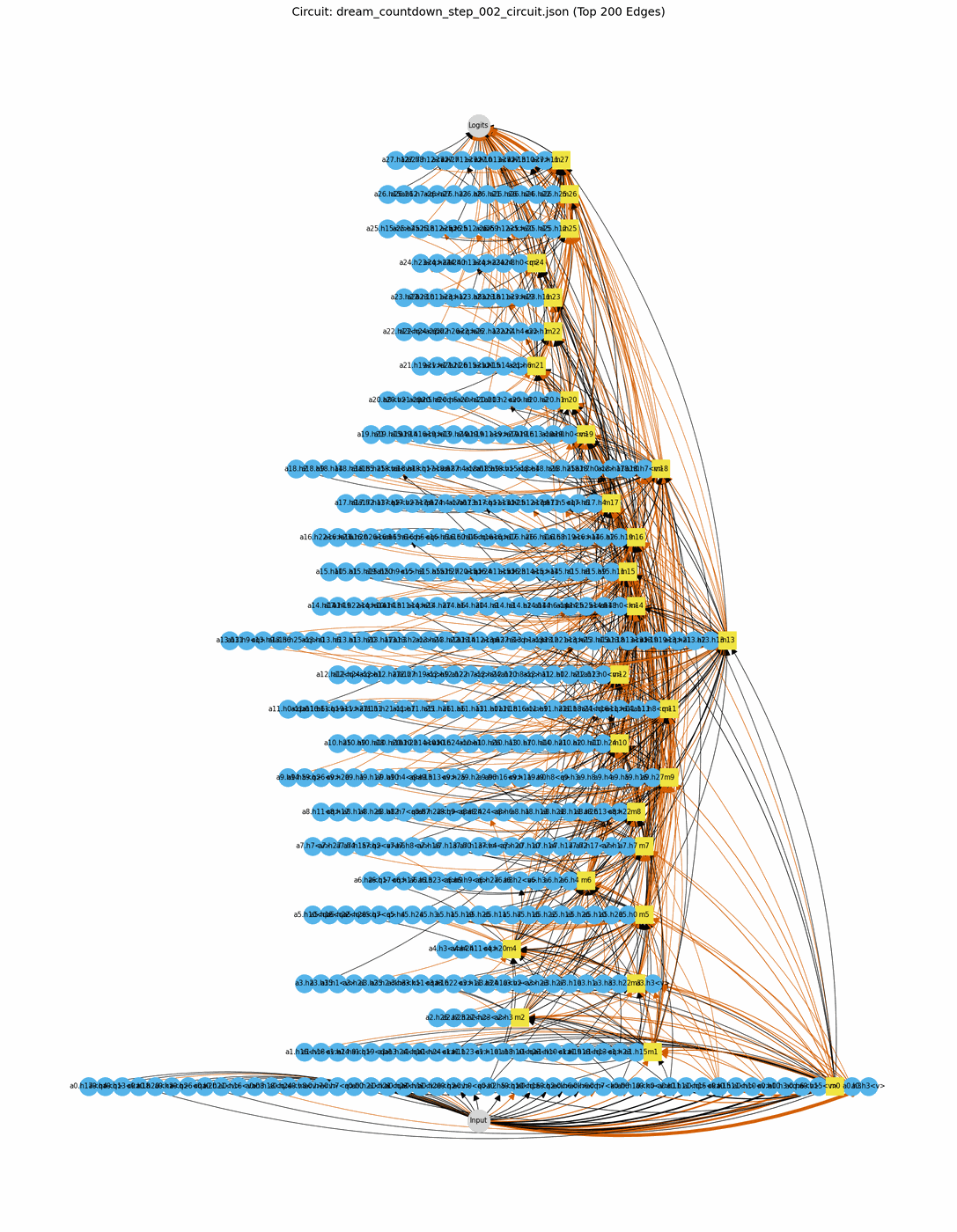}\hfill
    \includegraphics[width=0.24\linewidth]{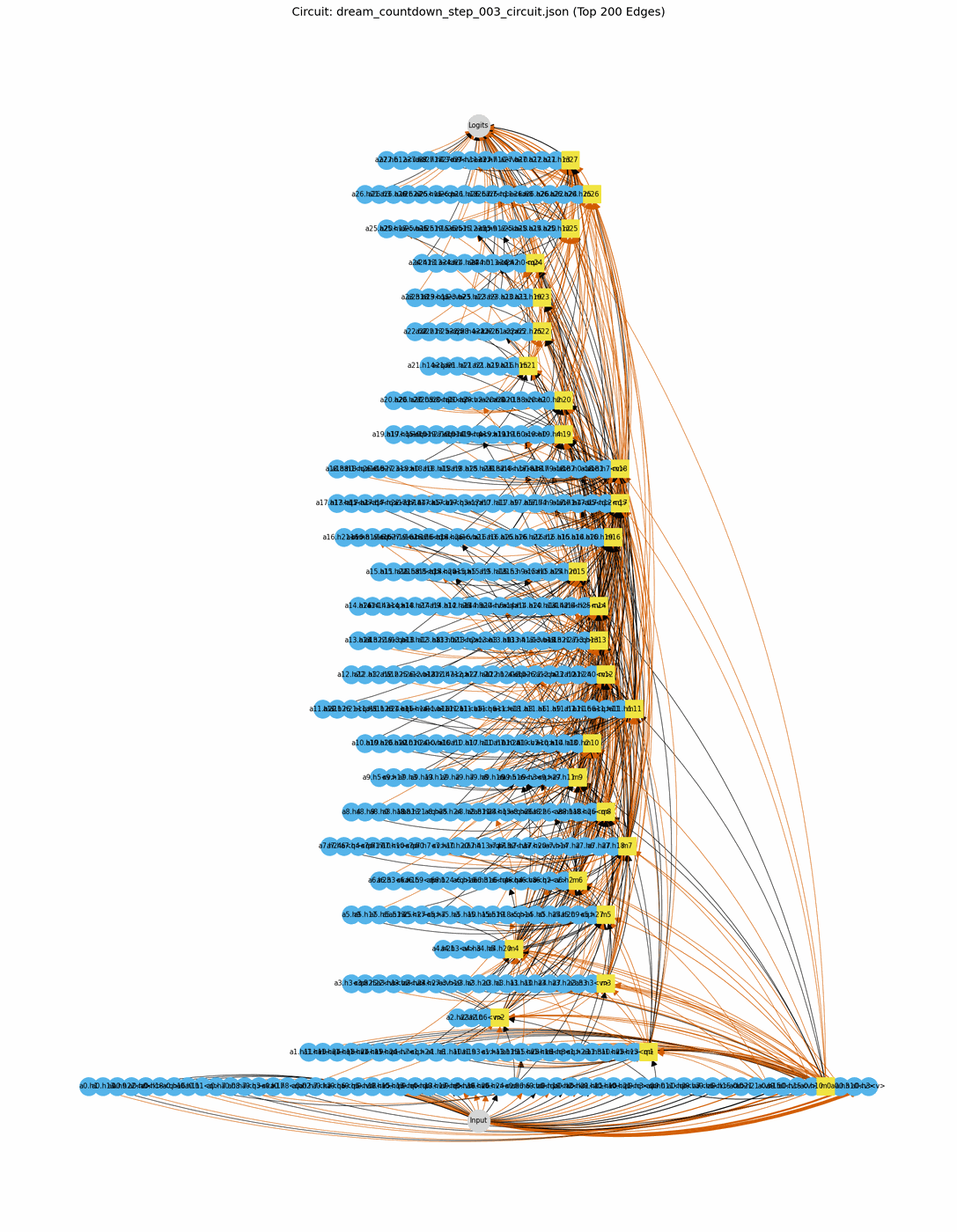}

    \vspace{0.4em}

    \includegraphics[width=0.24\linewidth]{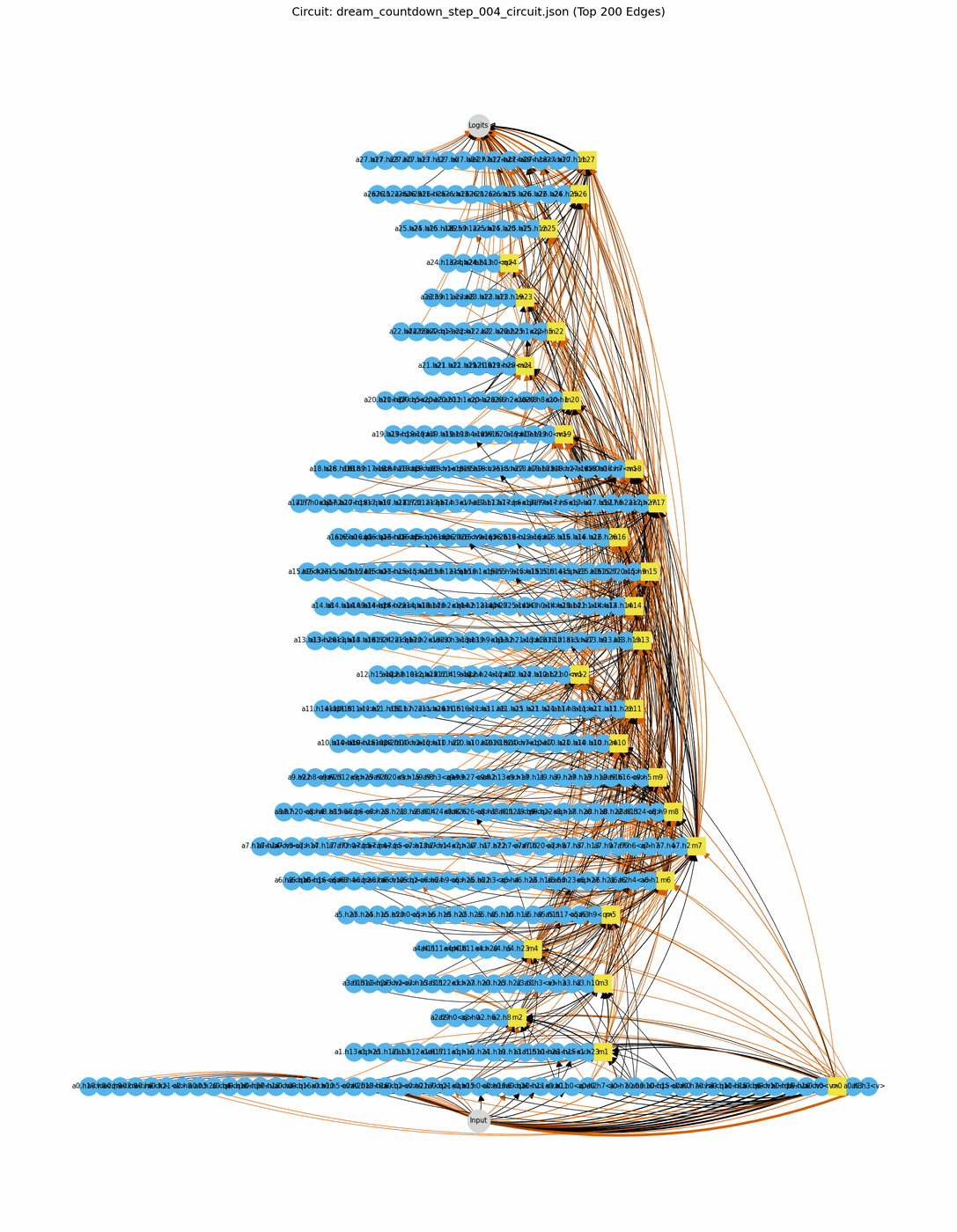}\hfill
    \includegraphics[width=0.24\linewidth]{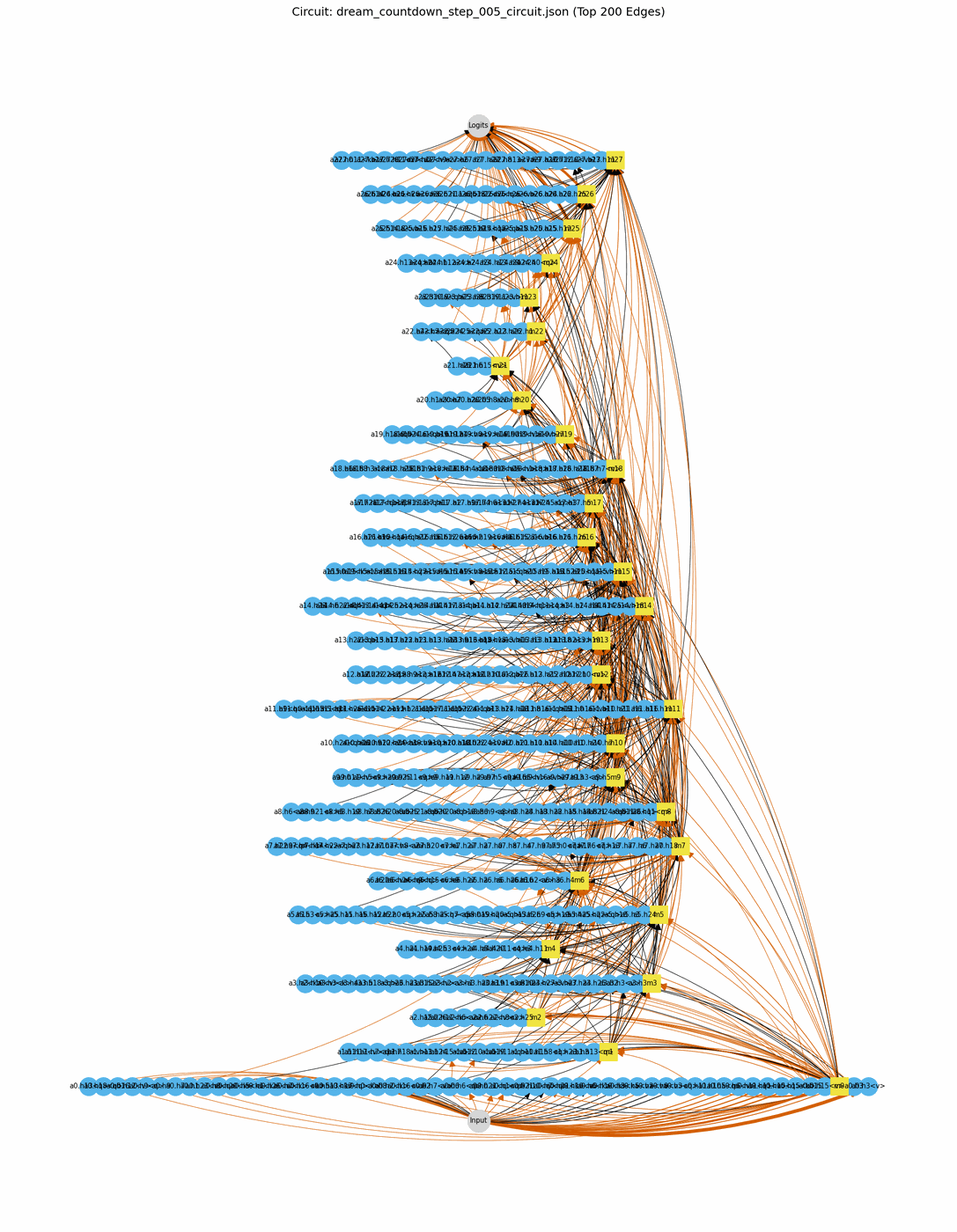}\hfill
    \includegraphics[width=0.24\linewidth]{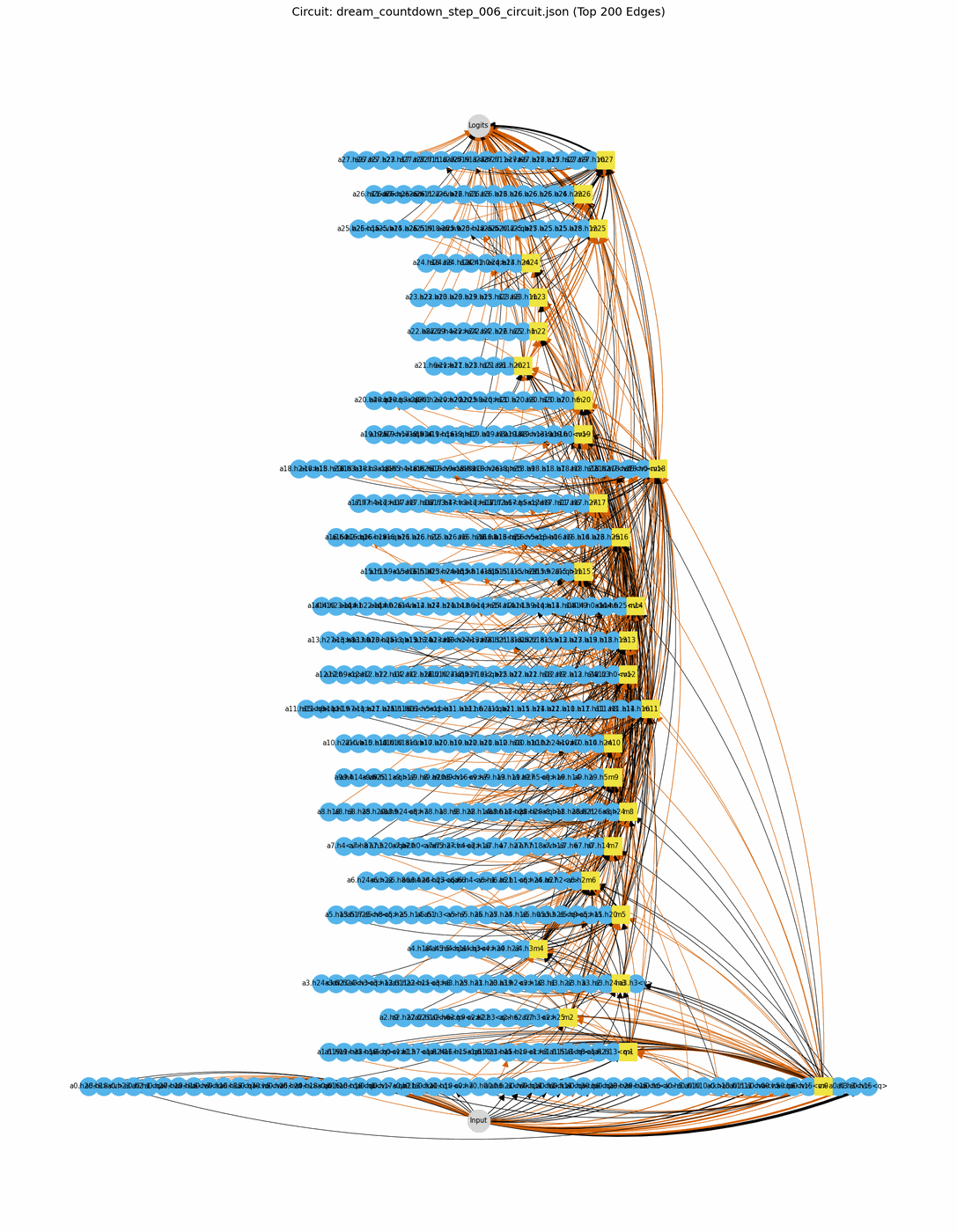}\hfill
    \includegraphics[width=0.24\linewidth]{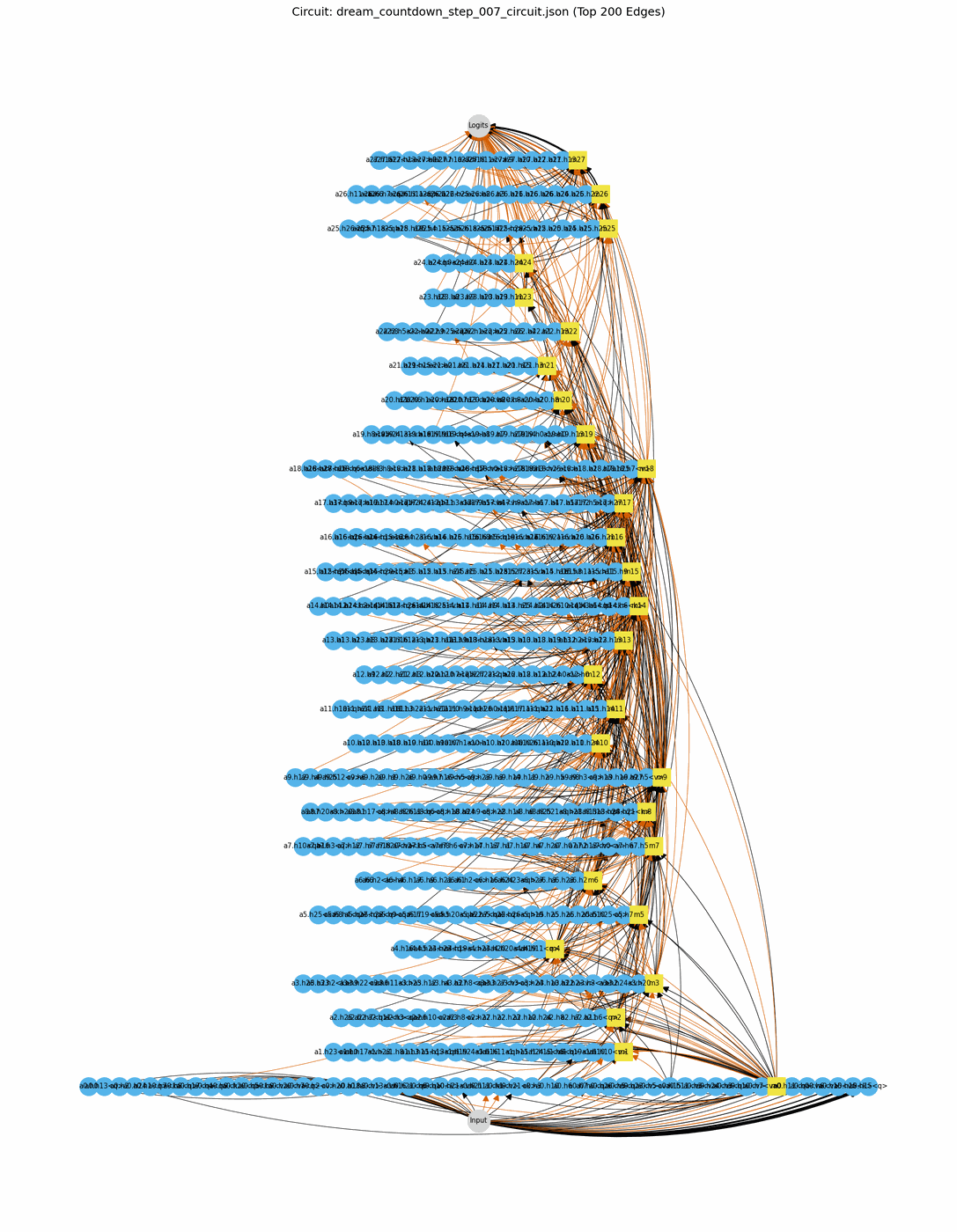}

    \vspace{0.4em}

    \includegraphics[width=0.24\linewidth]{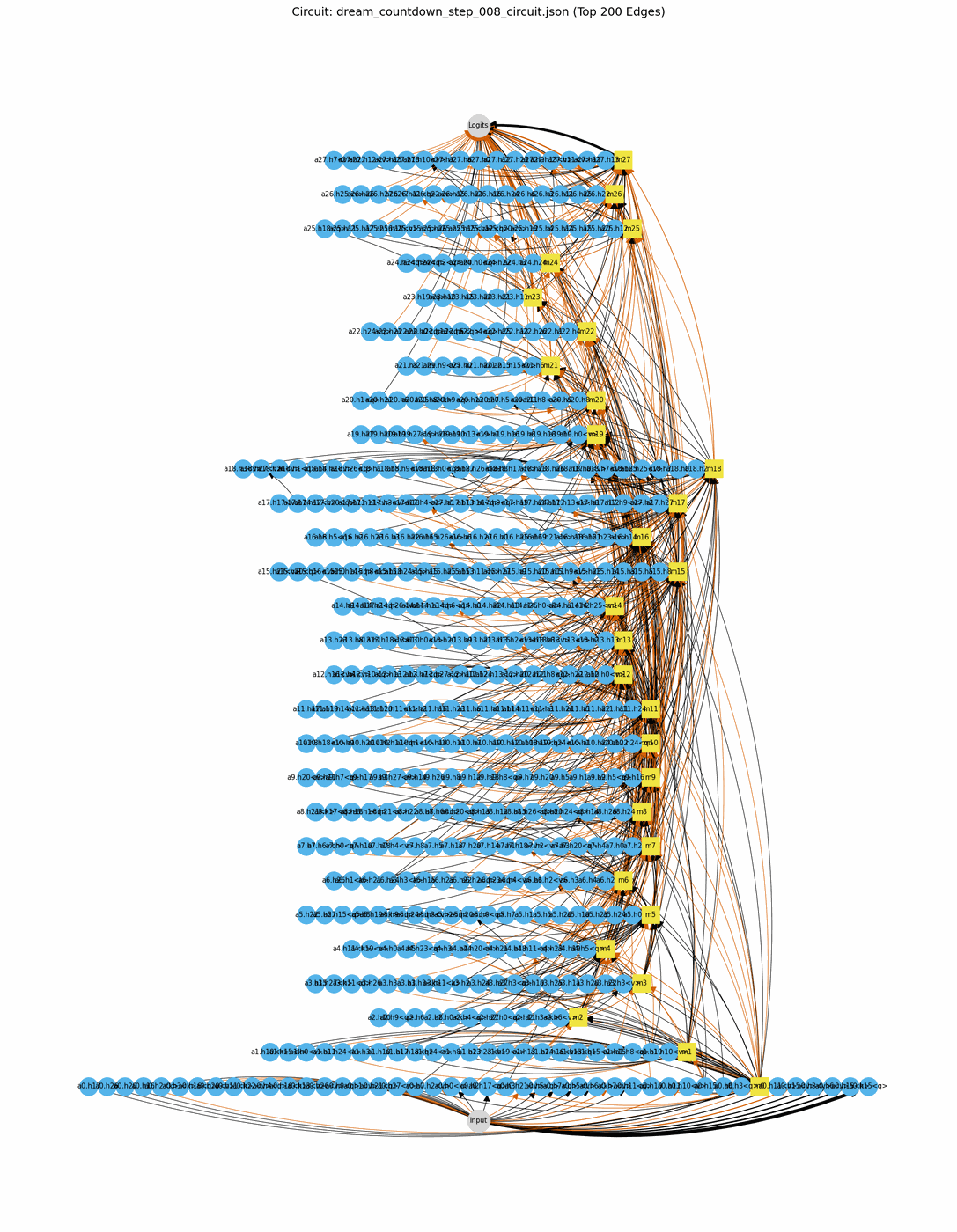}\hfill
    \includegraphics[width=0.24\linewidth]{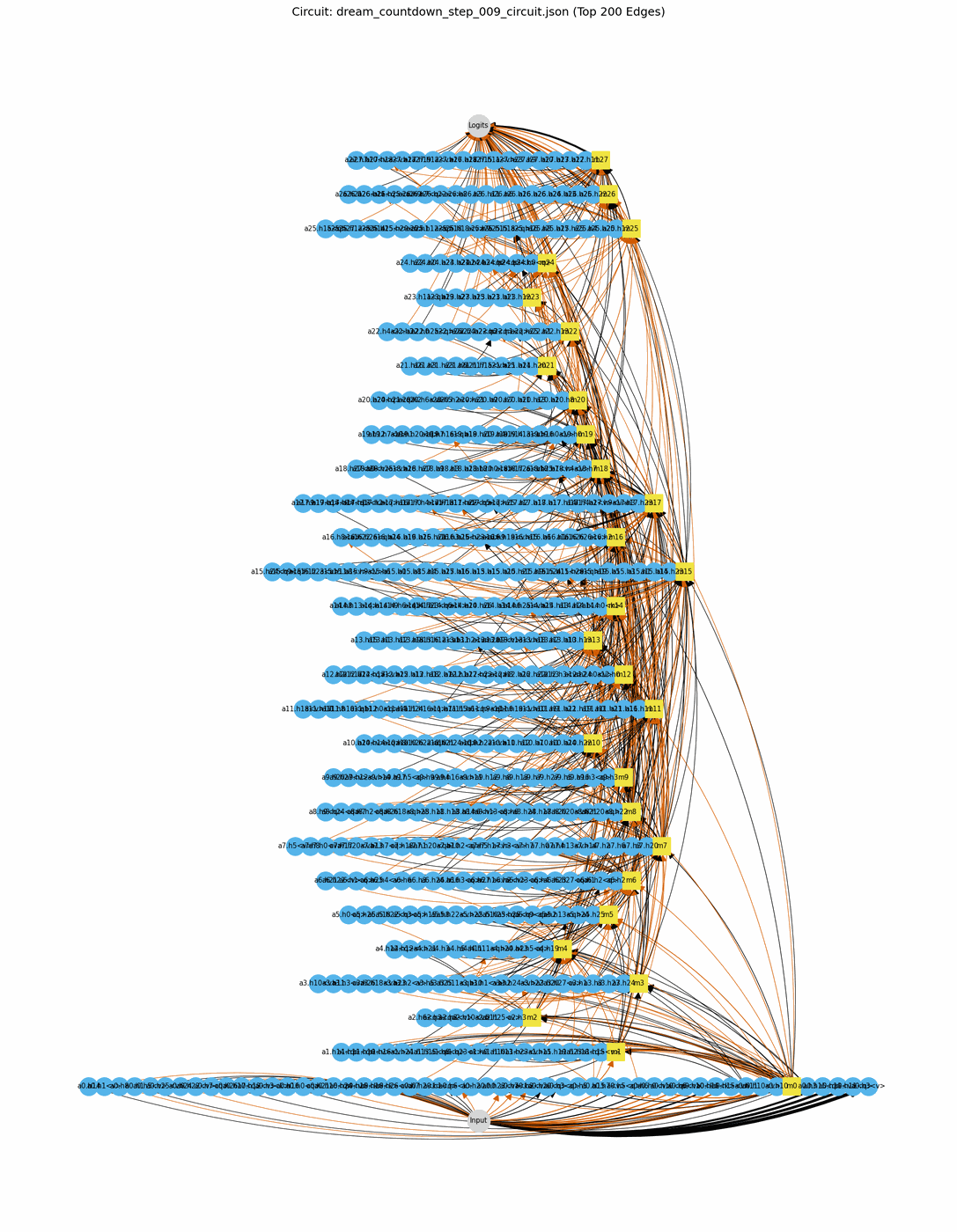}\hfill
    \includegraphics[width=0.24\linewidth]{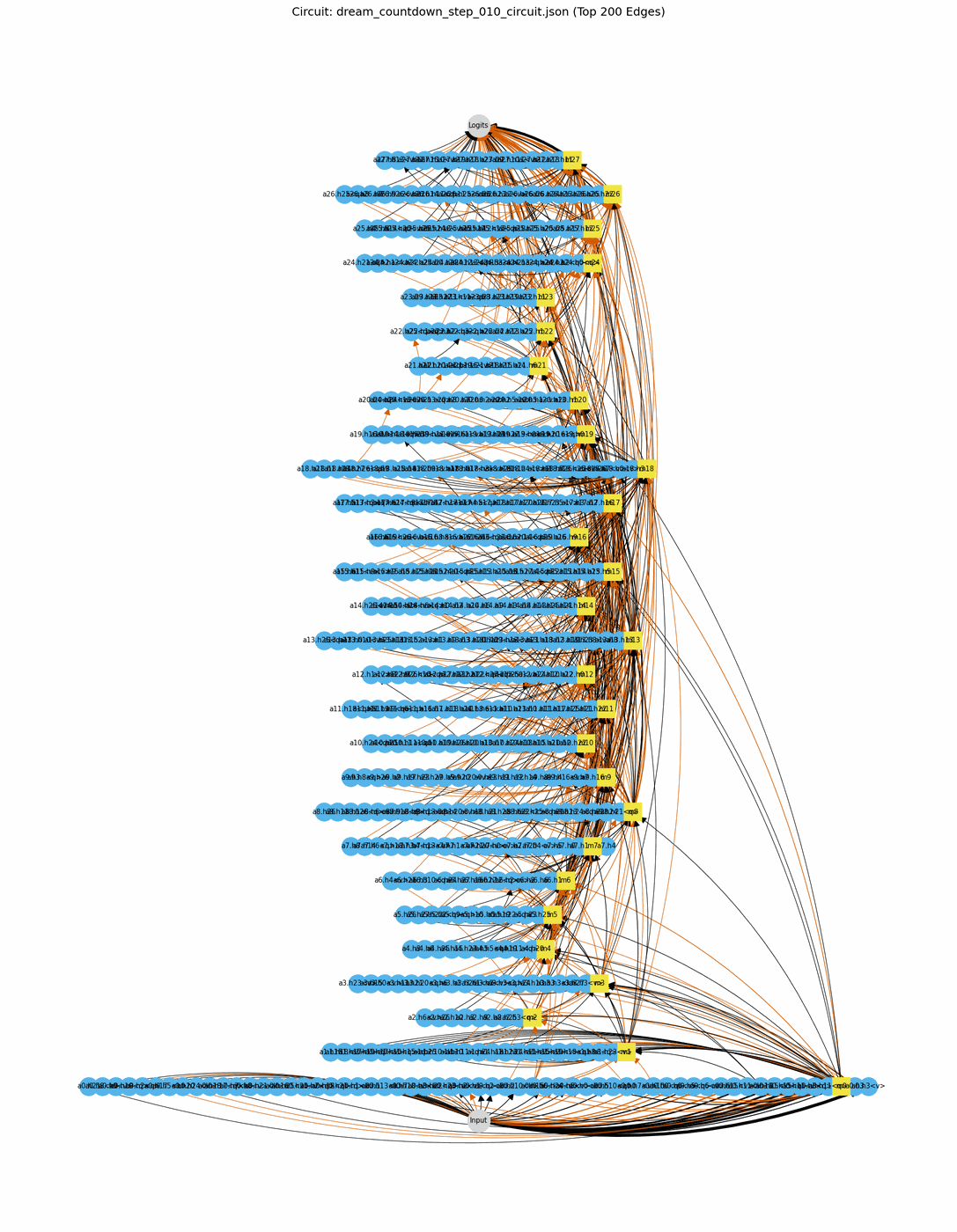}\hfill
    \includegraphics[width=0.24\linewidth]{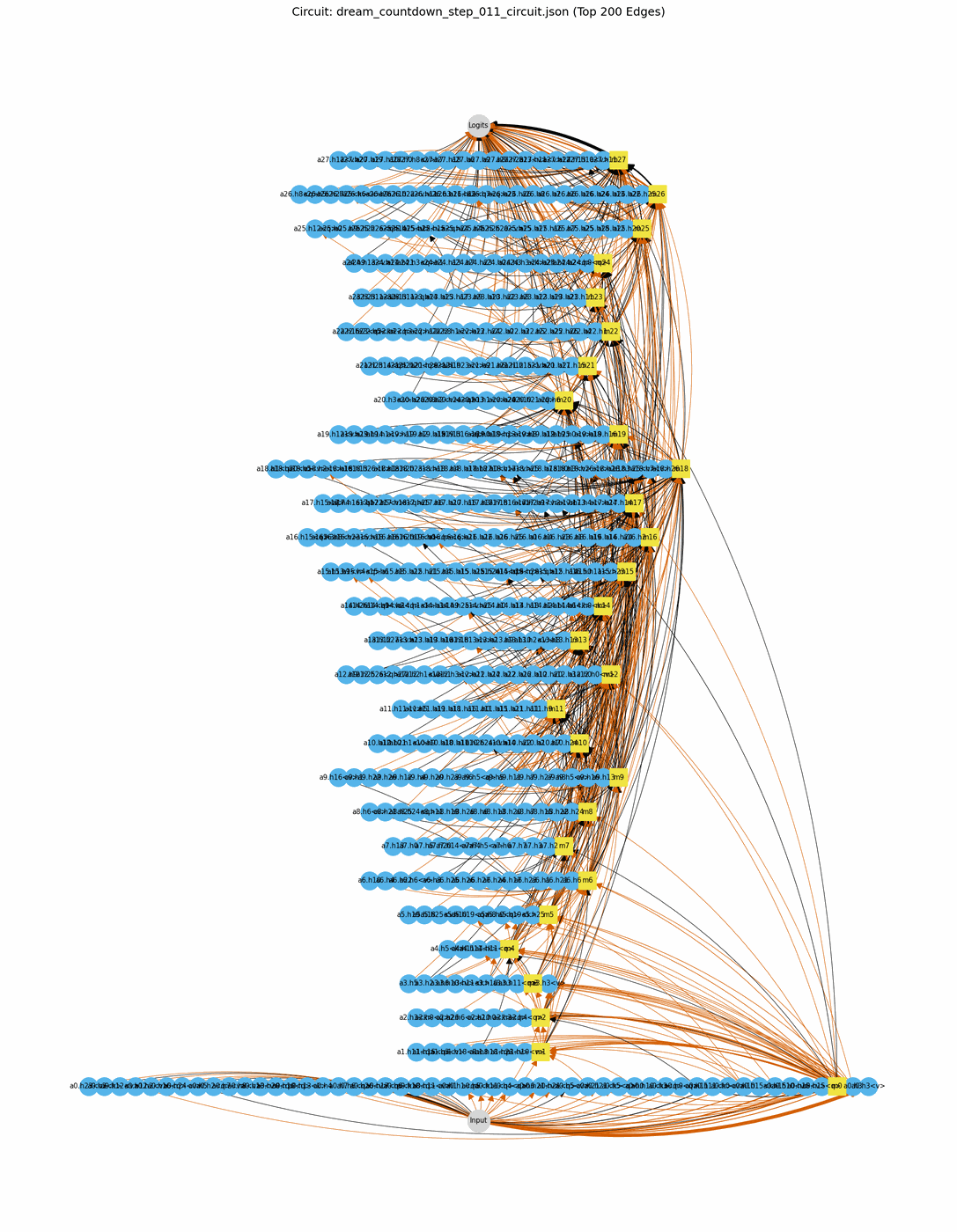}

    \caption{
    Step-wise circuit visualization of Dream on the \textsc{Countdown} task.
    Steps 1--12 are shown from left to right and top to bottom.
    }
    \label{fig:Dream-step-grid-12}
\end{figure*}

\begin{figure*}[!t]
    \centering
    \includegraphics[width=0.24\linewidth]{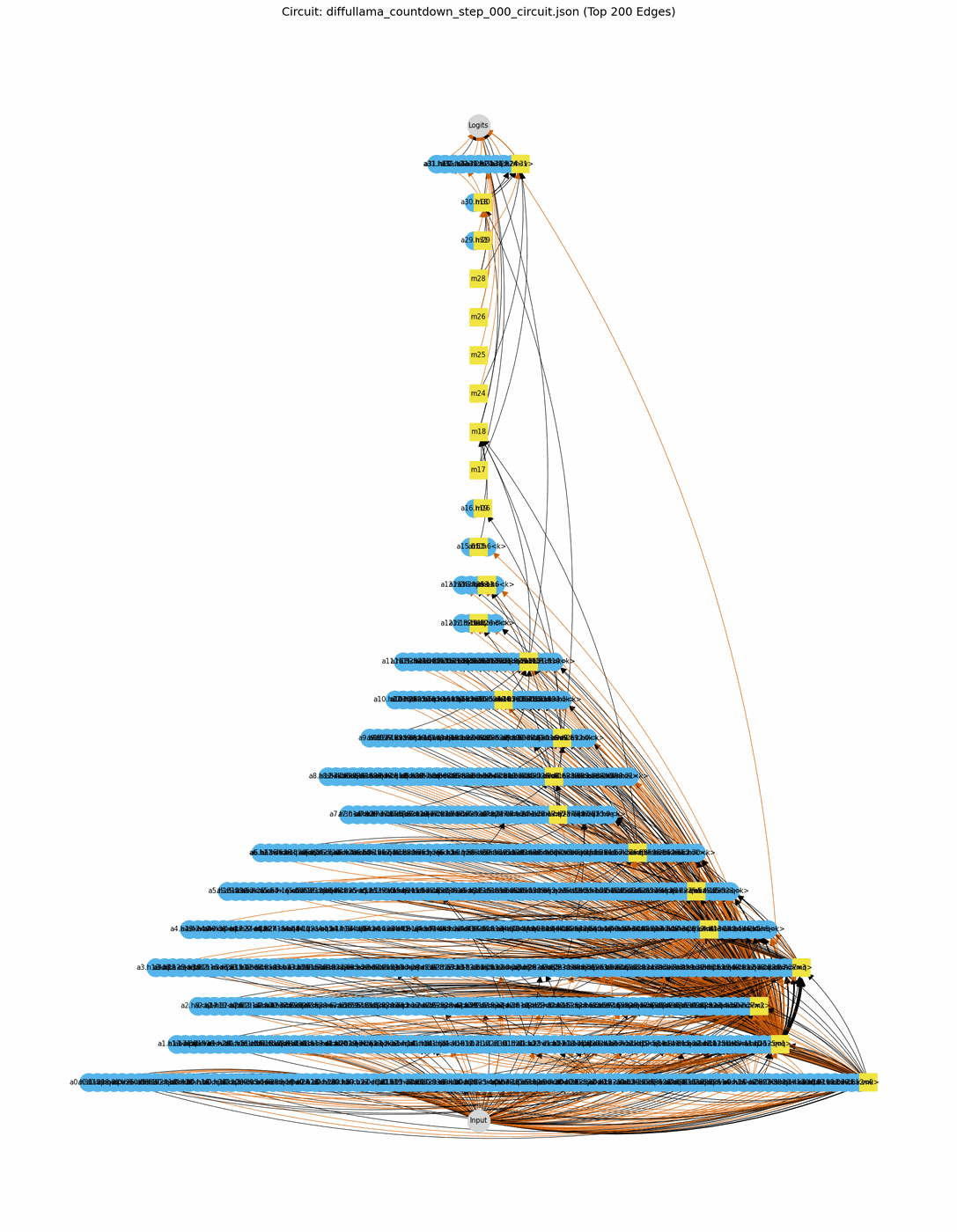}\hfill
    \includegraphics[width=0.24\linewidth]{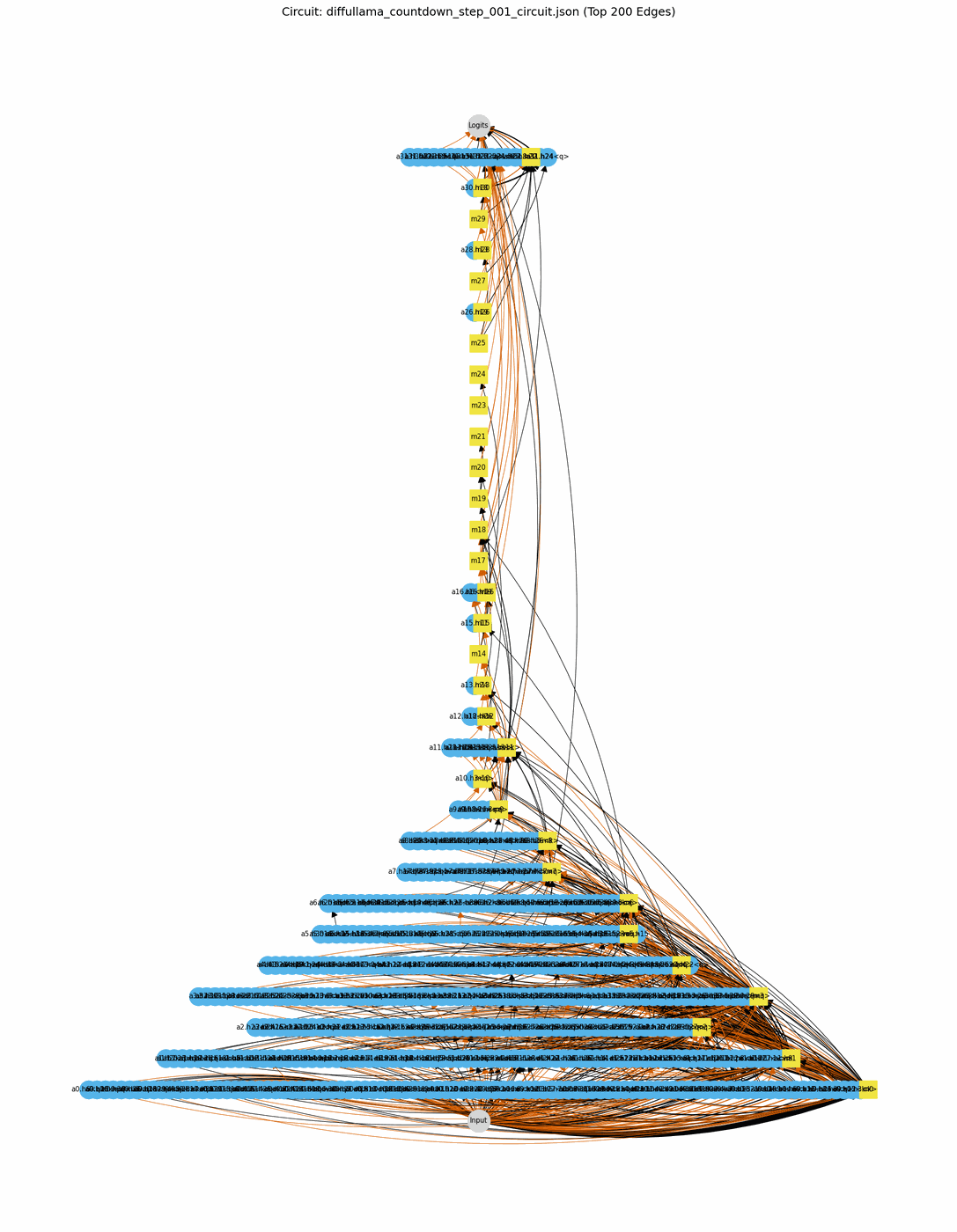}\hfill
    \includegraphics[width=0.24\linewidth]{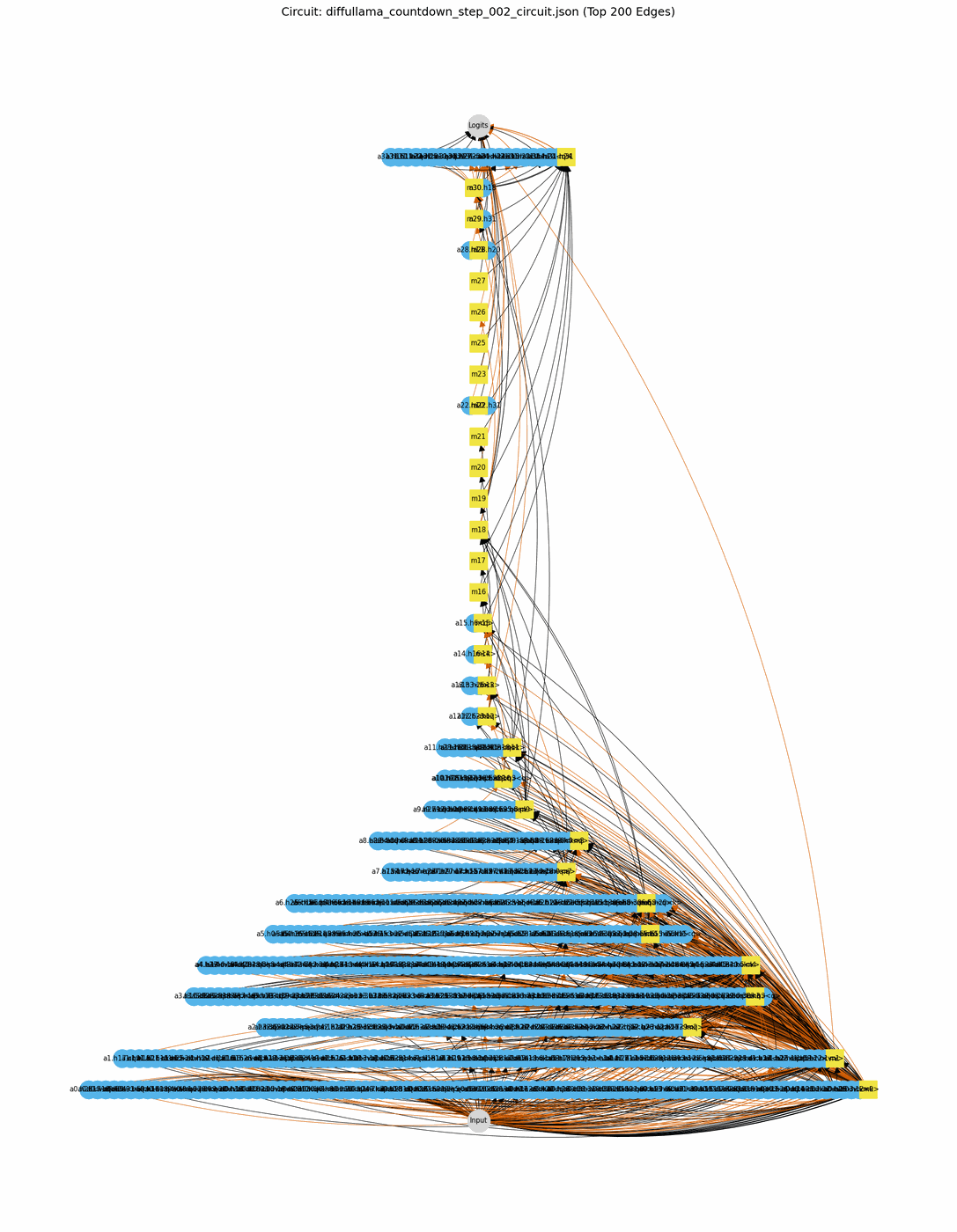}\hfill
    \includegraphics[width=0.24\linewidth]{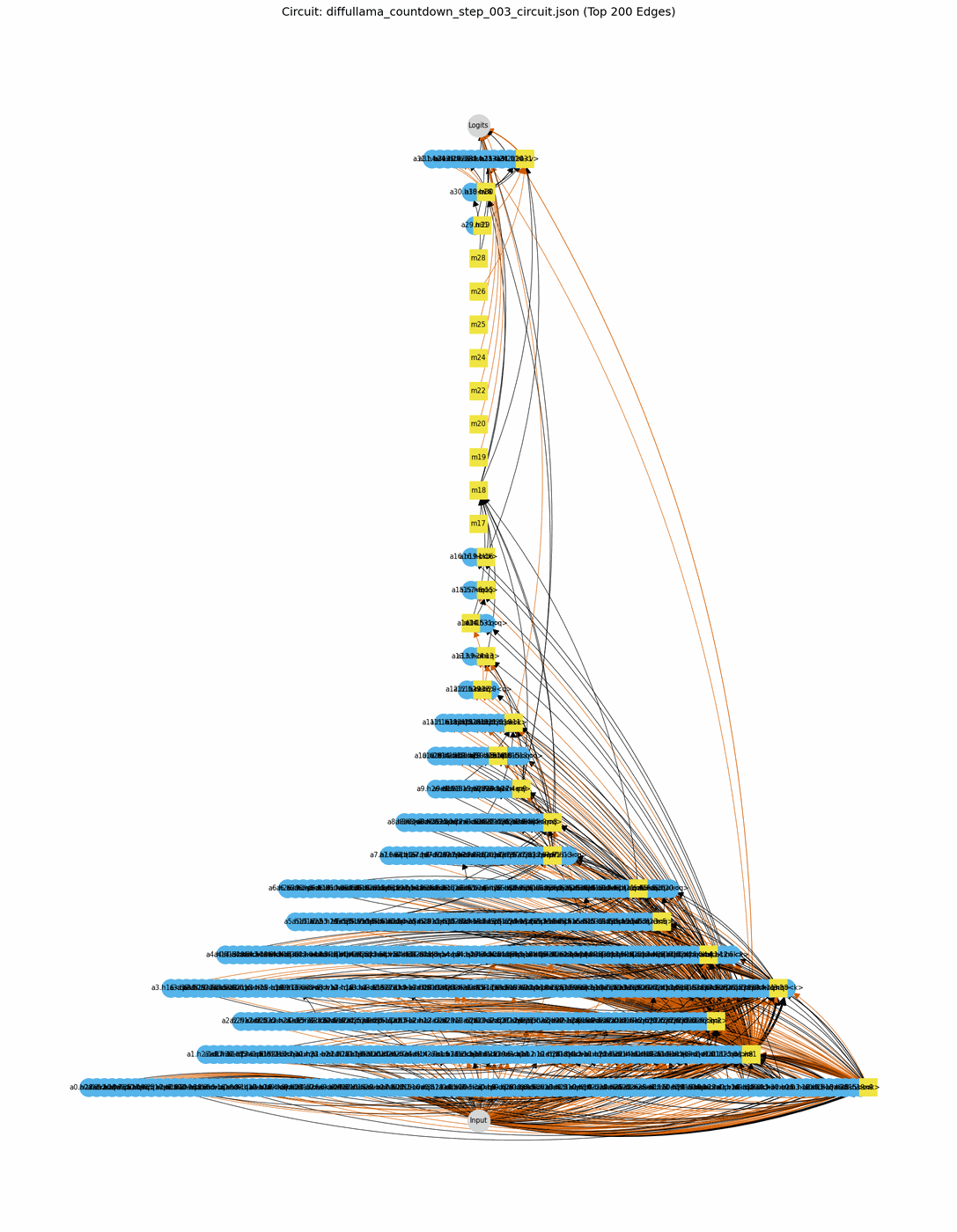}

    \vspace{0.4em}
    \includegraphics[width=0.24\linewidth]{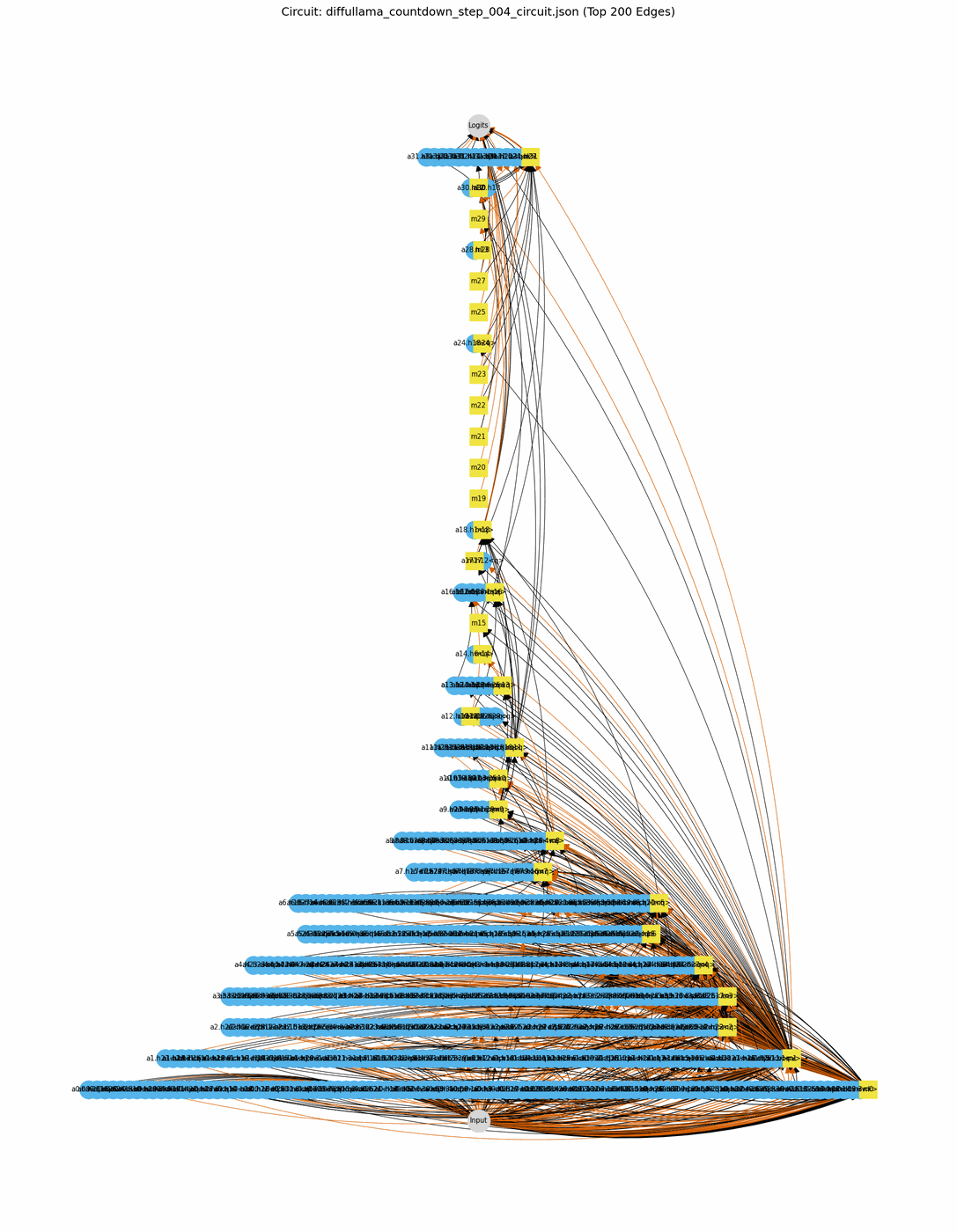}\hfill
    \includegraphics[width=0.24\linewidth]{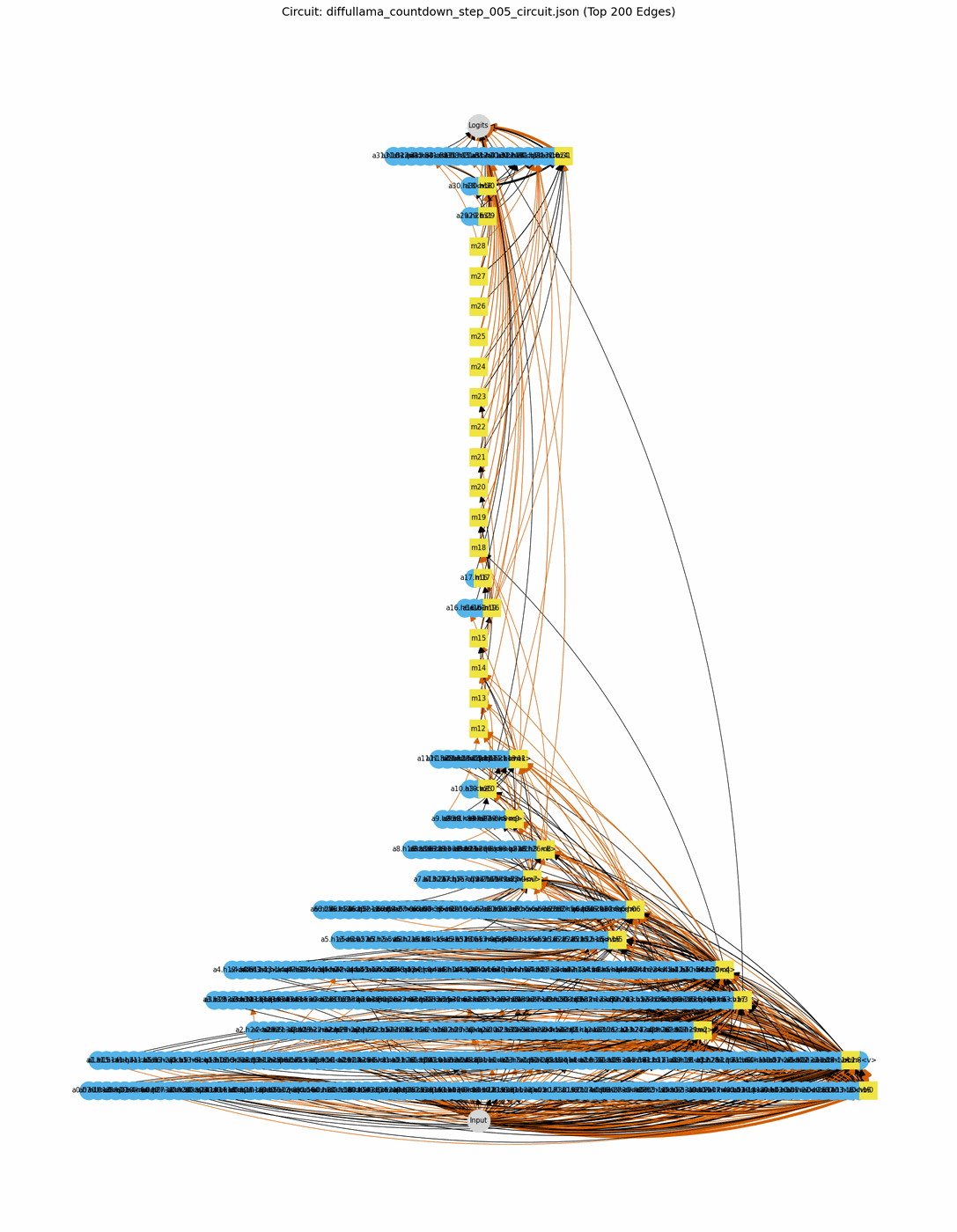}\hfill
    \includegraphics[width=0.24\linewidth]{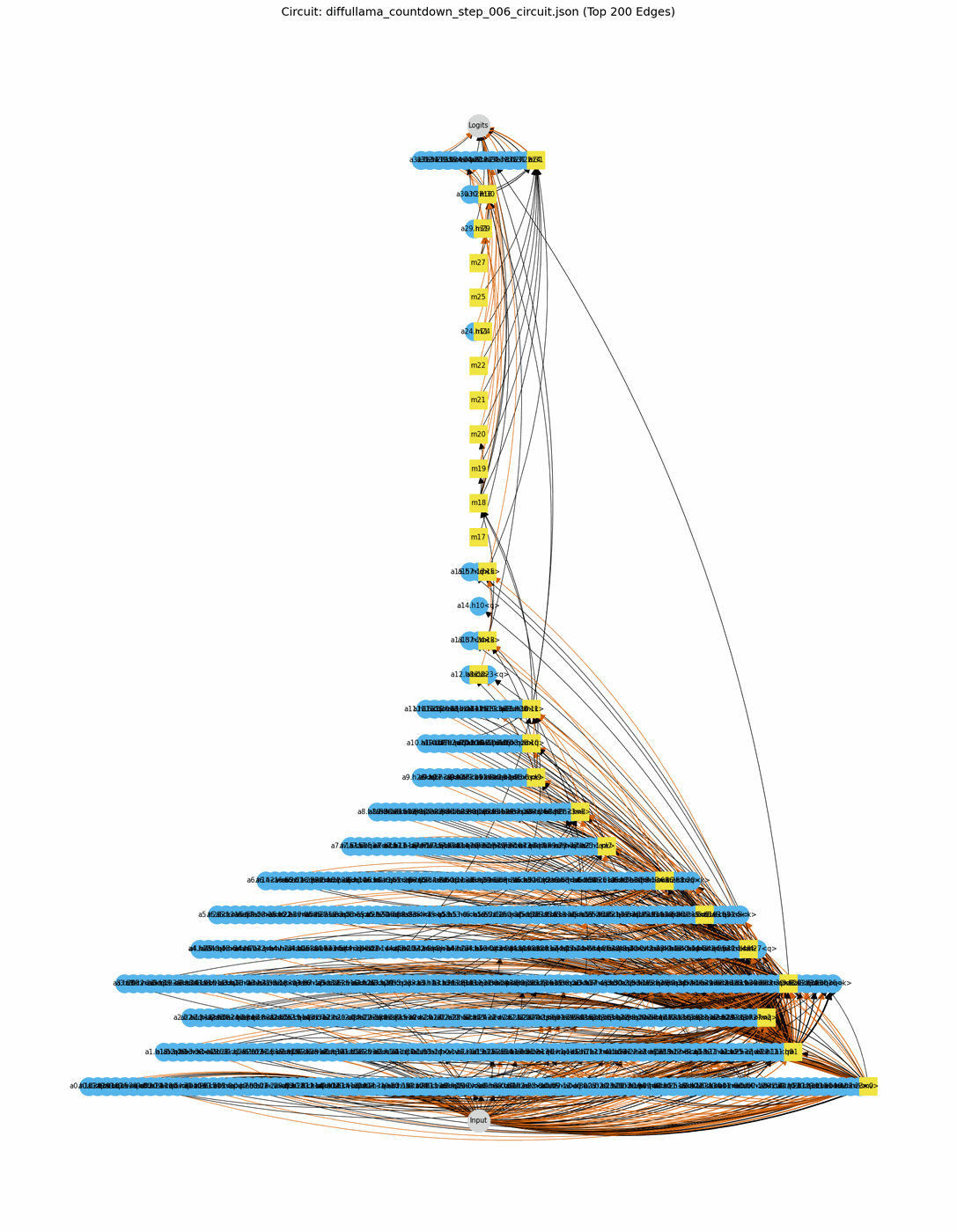}\hfill
    \includegraphics[width=0.24\linewidth]{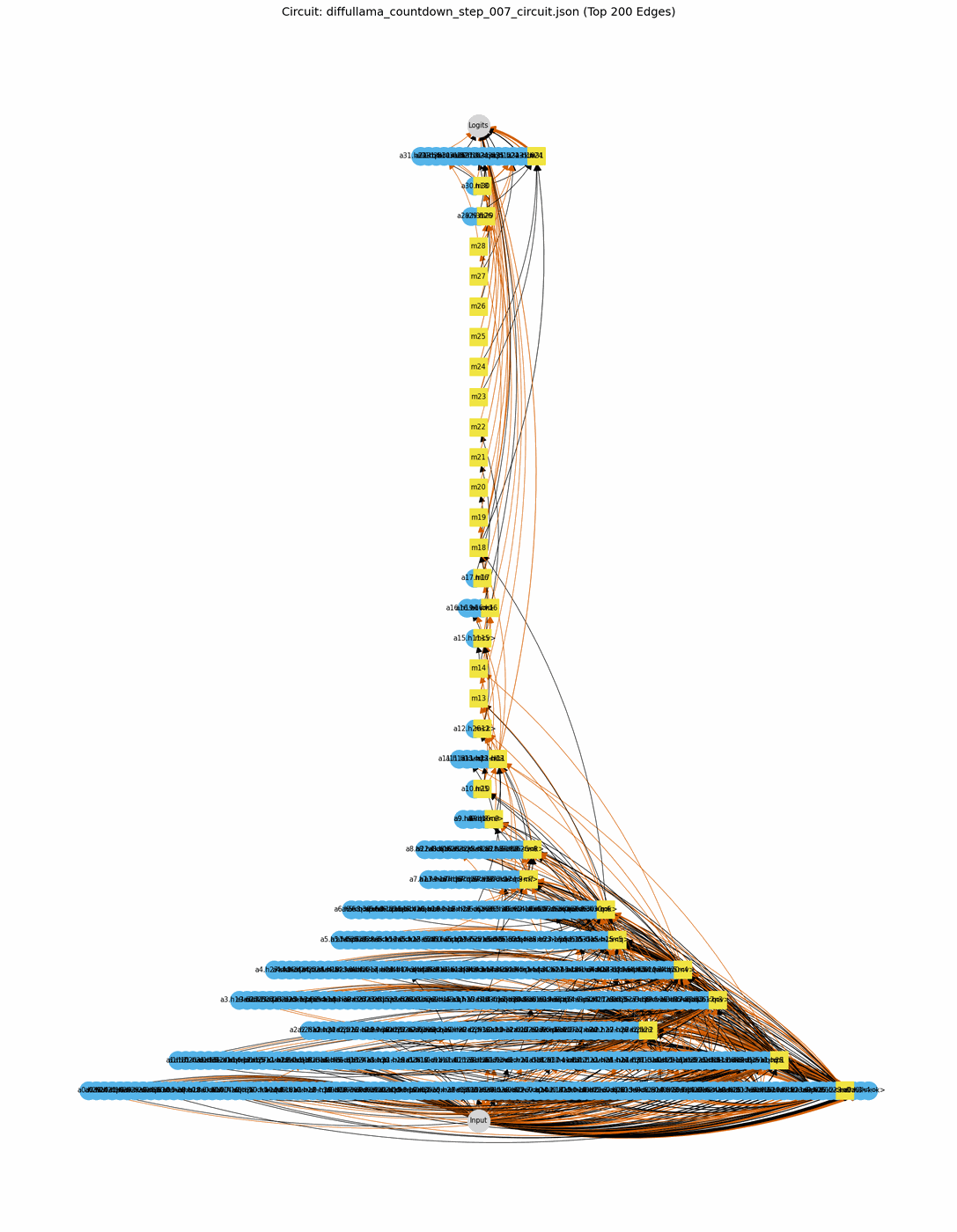}

    \vspace{0.4em}
    \includegraphics[width=0.24\linewidth]{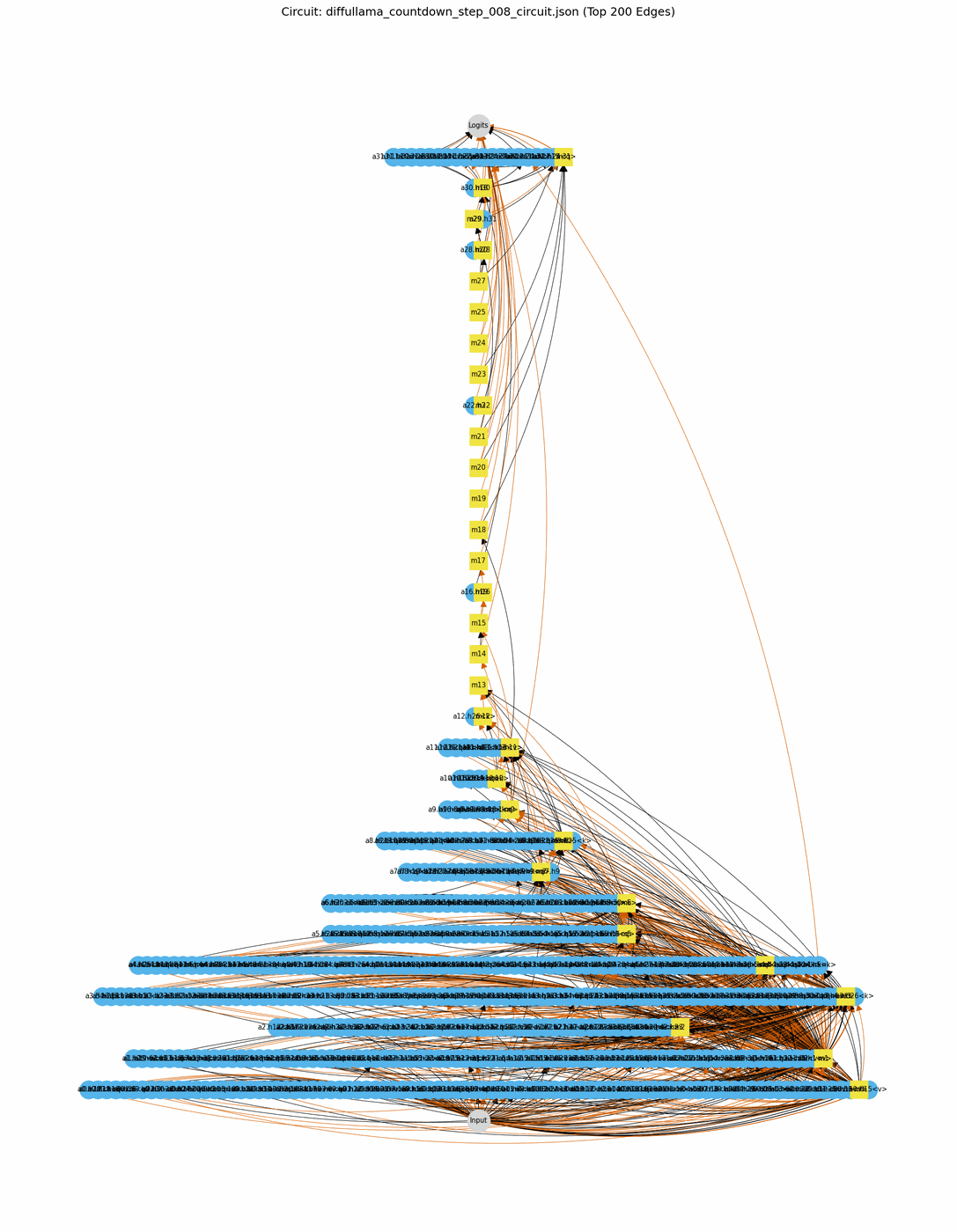}\hfill
    \includegraphics[width=0.24\linewidth]{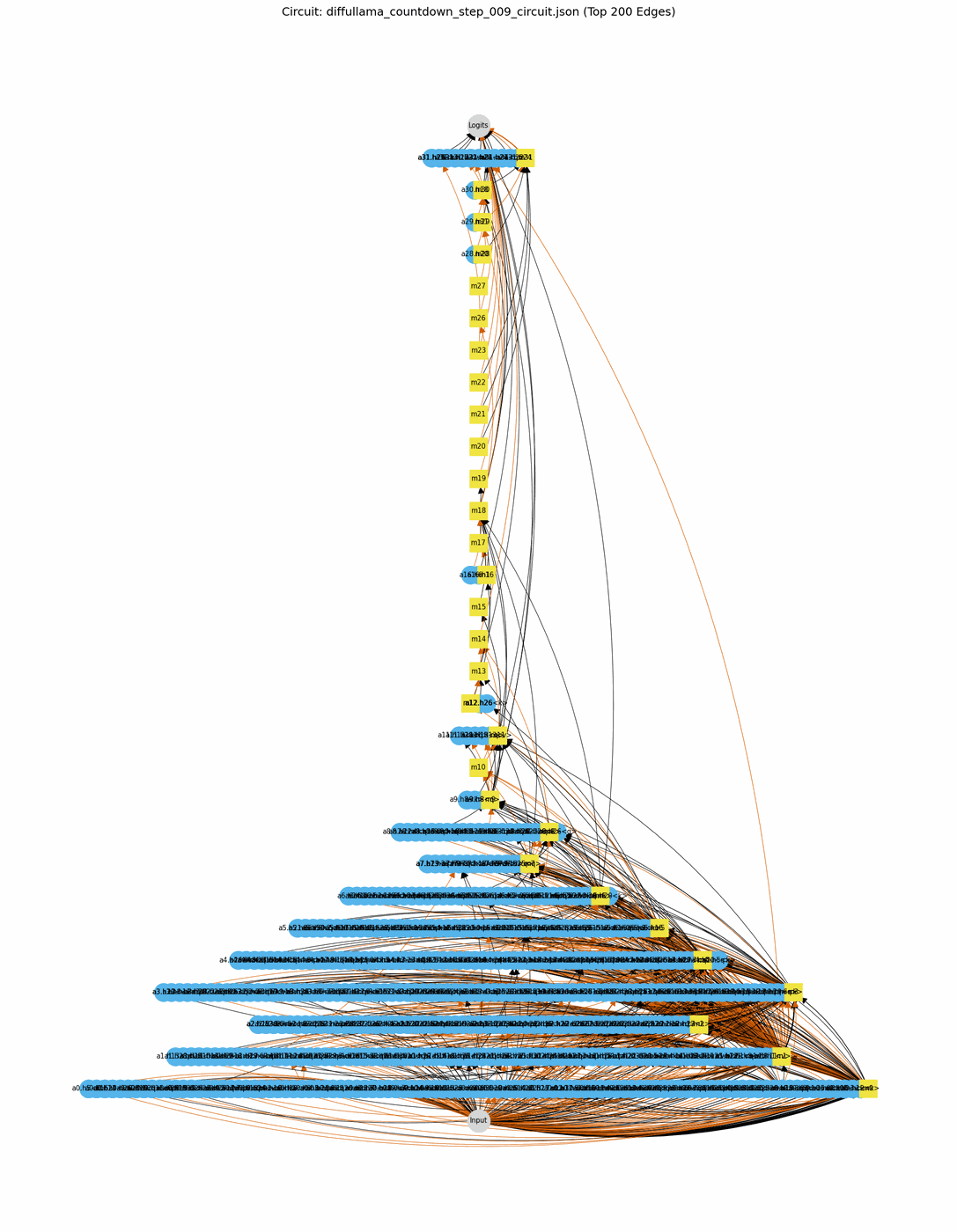}\hfill
    \includegraphics[width=0.24\linewidth]{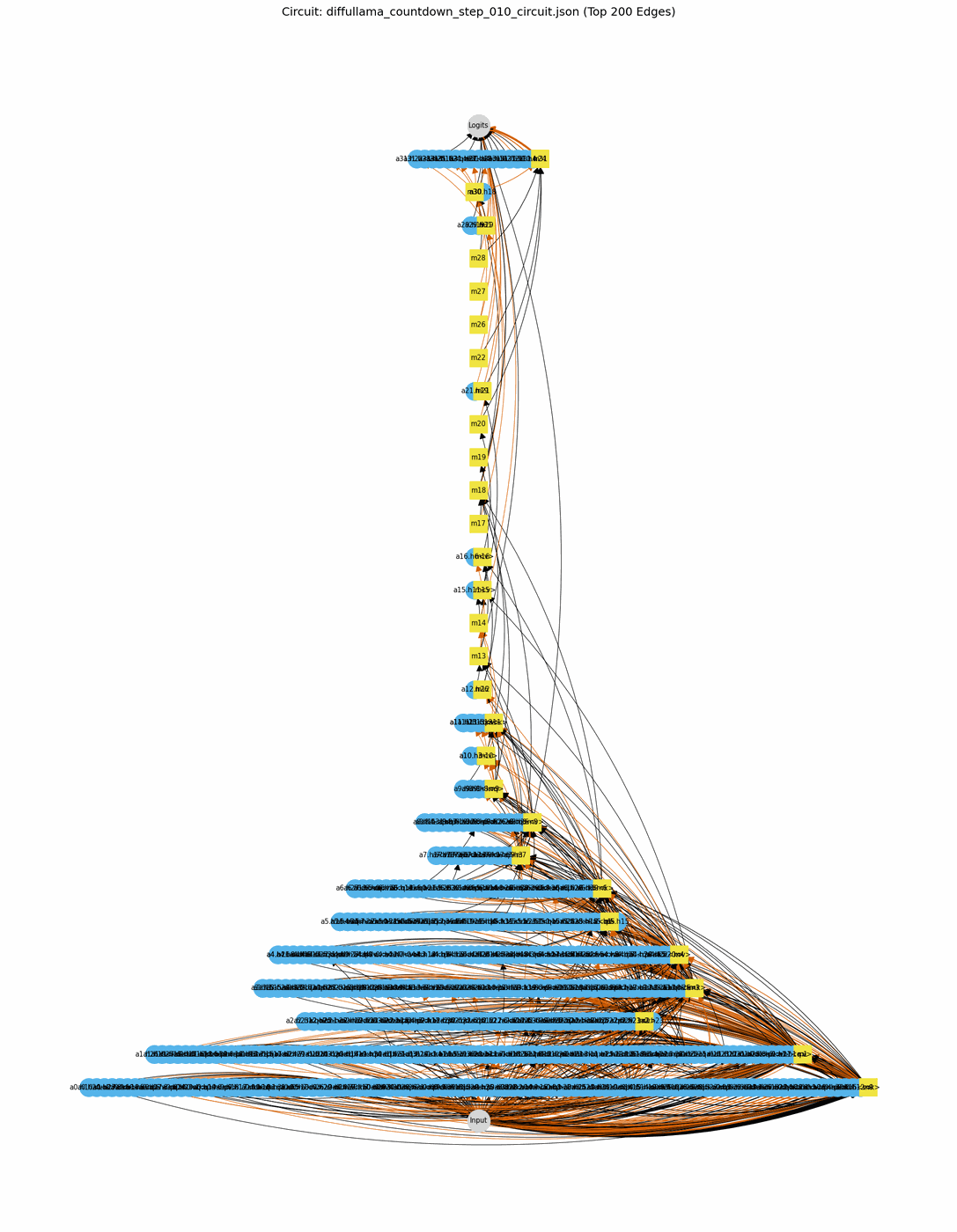}\hfill
    \includegraphics[width=0.24\linewidth]{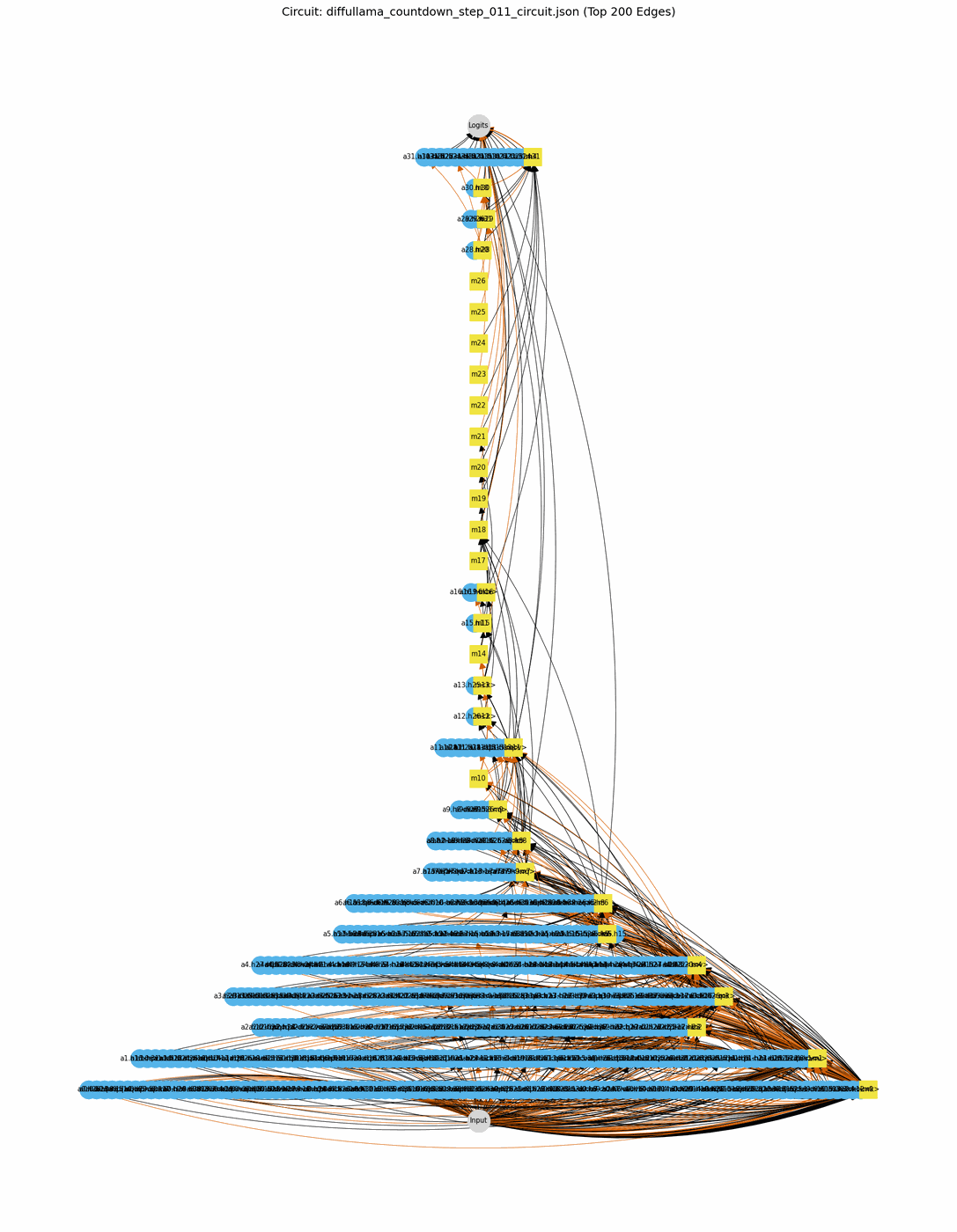}

    \caption{
    Step-wise circuit visualization of DiffuLLaMA on the \textsc{Countdown} task.
    Steps 1--12 are shown from left to right and top to bottom.
    }
    
    \label{fig:Diffullama-step-grid-13}
\end{figure*}

\paragraph{Step-wise Circuit Stability}
For step-wise circuit extraction, we compute circuits at each diffusion step and aggregate attribution scores across steps. The set of participating components (attention heads and MLP layers) remains largely stable over diffusion time, with more than 70\% overlap in selected components between steps. Accordingly, the visualizations in Figures~\ref{fig:Diffullama-step-grid-13} and~\ref{fig:Dream-step-grid-12} represent averaged structures; step-wise variation is reflected mainly in the attributed edges rather than in component identity. This suggests that masked diffusion primarily refines information routing within a largely fixed component set, rather than progressively recruiting new components.

\begin{table*}[t]
\centering
\small
\begin{tabular}{llcp{5.5cm}r}
\toprule
\textbf{Task} & \textbf{Model} & \textbf{Layer} & \textbf{Explanation} & \textbf{Count} \\
\midrule
\multirow{6}{*}[-1ex]{\textsc{IOI}} & \multirow{3}{*}[-0.5ex]{LLaMA} & \multirow{3}{*}[-0.5ex]{0--4} & descriptive adjectives & 2821 \\
& & & expressions of uncertainty or doubt & 104 \\
& & & synonyms for happiness or joy & 50 \\
\cmidrule{2-5}
& \multirow{3}{*}[-0.5ex]{DiffuLLaMA} & \multirow{3}{*}[-0.5ex]{0--4} & synonyms for happiness or joy & 193 \\
& & & expressions of uncertainty or doubt & 80 \\
& & & technical jargon or specialized terminology & 57 \\
\midrule
\multirow{6}{*}[-2ex]{\textsc{Countdown}} & \multirow{3}{*}[-0.5ex]{LLaMA} & \multirow{3}{*}[-0.5ex]{0--4} & numerical values or quantities & 1012 \\
& & & terms associated with technology and innovation & 226 \\
& & & technical terms related to technology & 86 \\
\cmidrule{2-5}
& \multirow{3}{*}[-0.5ex]{DiffuLLaMA} & \multirow{3}{*}[-0.5ex]{0--4} & synonyms for happiness or joy & 207 \\
& & & expressions of uncertainty or doubt & 114 \\
& & & terms associated with technology and innovation & 72 \\
\bottomrule
\end{tabular}
\caption{Top 3 most common semantic explanations for active neurons in early layers (0--4). Autoregressive models (LLaMA) display sharp specialization, dedicating an overwhelming 2,821 neurons to descriptive adjectives in IOI and 1,012 neurons to numerical values in Countdown. In contrast, MDMs (DiffuLLaMA) display a flat, task-agnostic profile, utilizing only $\sim$50--200 active neurons per category for broad concepts regardless of the task.}
\label{tab:neuron_explanations}
\end{table*}



\begin{figure*}[!t]
    \centering

    \begin{minipage}{0.48\linewidth}
        \centering
        \includegraphics[height=0.20\textheight,keepaspectratio]{figures/images/circuits/imgs/Qwen2.5-7B_ioi_eapig_edges.png}
        \caption*{(a) IOI — Qwen-2.5-7B}
    \end{minipage}
    \hfill
    \begin{minipage}{0.48\linewidth}
        \centering
        \includegraphics[height=0.20\textheight,keepaspectratio]{figures/images/circuits/imgs/dream_ioi_step_000_circuit_1.png}
        \caption*{(b) IOI — Dream-Base-7B}
    \end{minipage}

    \vspace{0.3em}

    \begin{minipage}{0.48\linewidth}
        \centering
        \includegraphics[height=0.20\textheight,keepaspectratio]{figures/images/circuits/imgs/Qwen2.5-7B_countdown_edges.png}
        \caption*{(c) Countdown — Qwen-2.5-7B}
    \end{minipage}
    \hfill
    \begin{minipage}{0.48\linewidth}
        \centering
        \includegraphics[height=0.20\textheight,keepaspectratio]{figures/images/circuits/imgs/dream_countdown_avg_circuit.png}
        \caption*{(d) Countdown — Dream-Base-7B}
    \end{minipage}

    \vspace{0.3em}

    \begin{minipage}{0.48\linewidth}
        \centering
        \includegraphics[height=0.20\textheight,keepaspectratio]{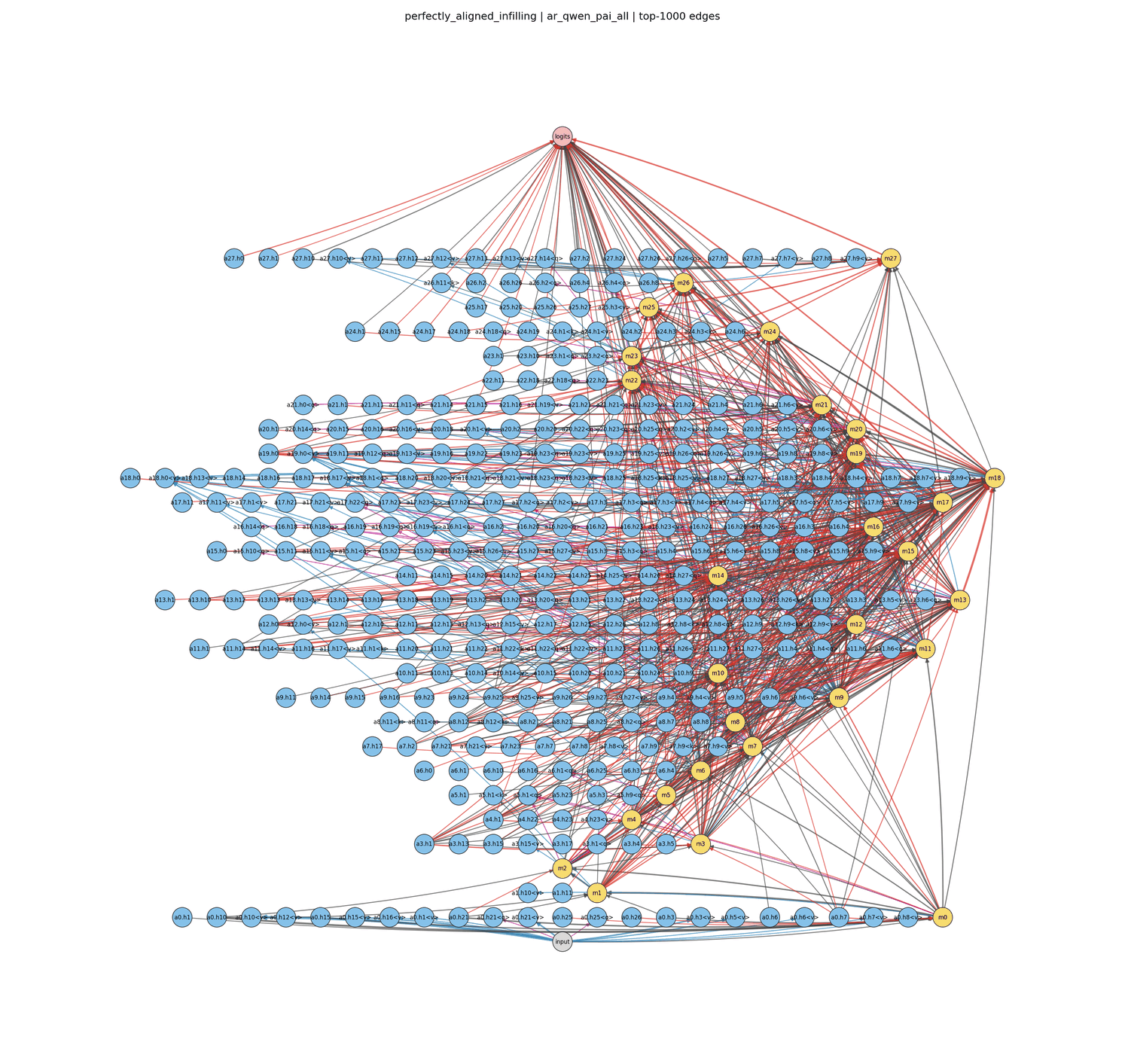}
        \caption*{(e) Semantic Infilling — Qwen-2.5-7B}
    \end{minipage}
    \hfill
    \begin{minipage}{0.48\linewidth}
        \centering
        \includegraphics[height=0.20\textheight,keepaspectratio]{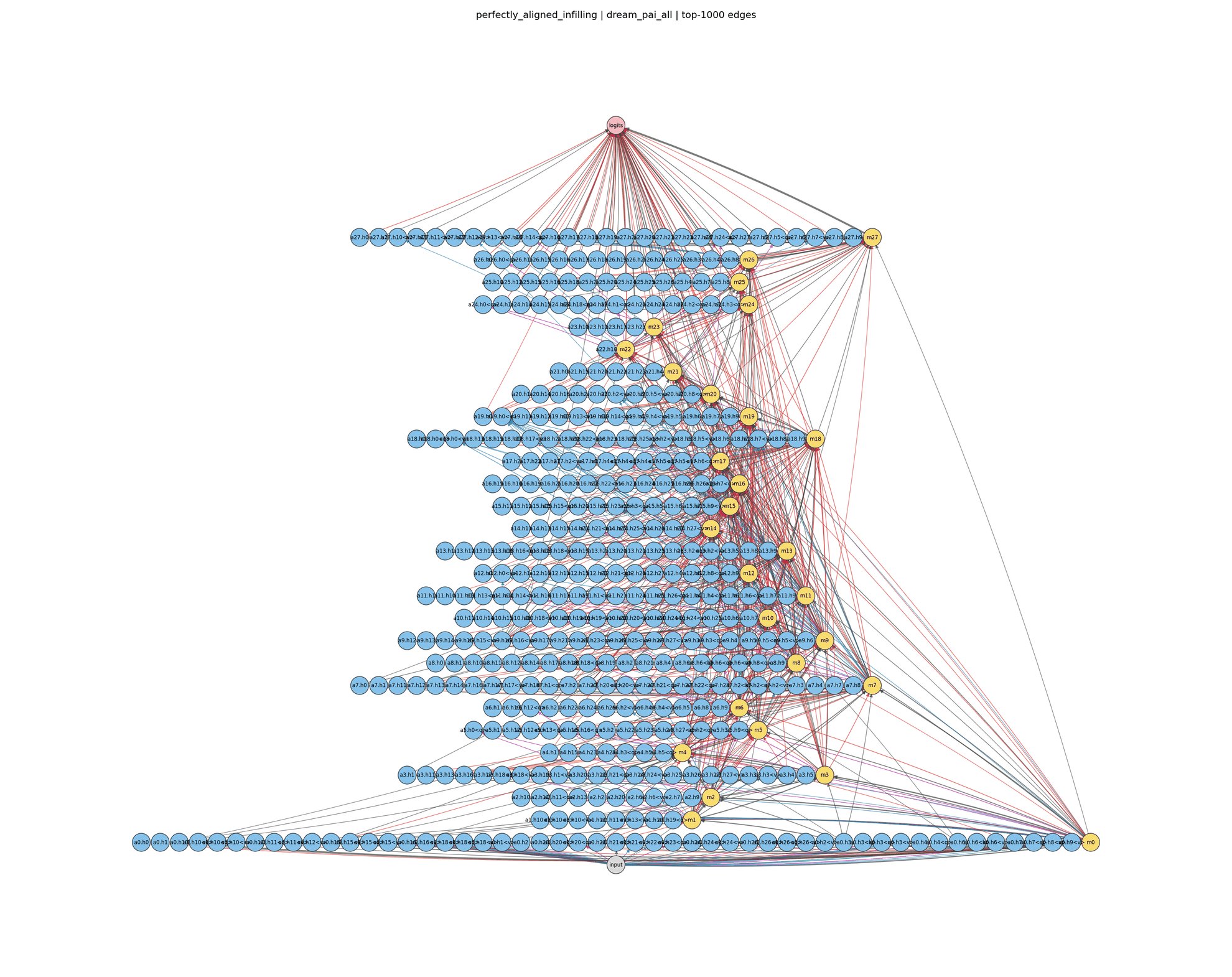}
        \caption*{(f) Semantic Infilling — Dream-Base-7B}
    \end{minipage}

    \vspace{0.3em}

    \begin{minipage}{0.48\linewidth}
        \centering
        \includegraphics[height=0.20\textheight,keepaspectratio]{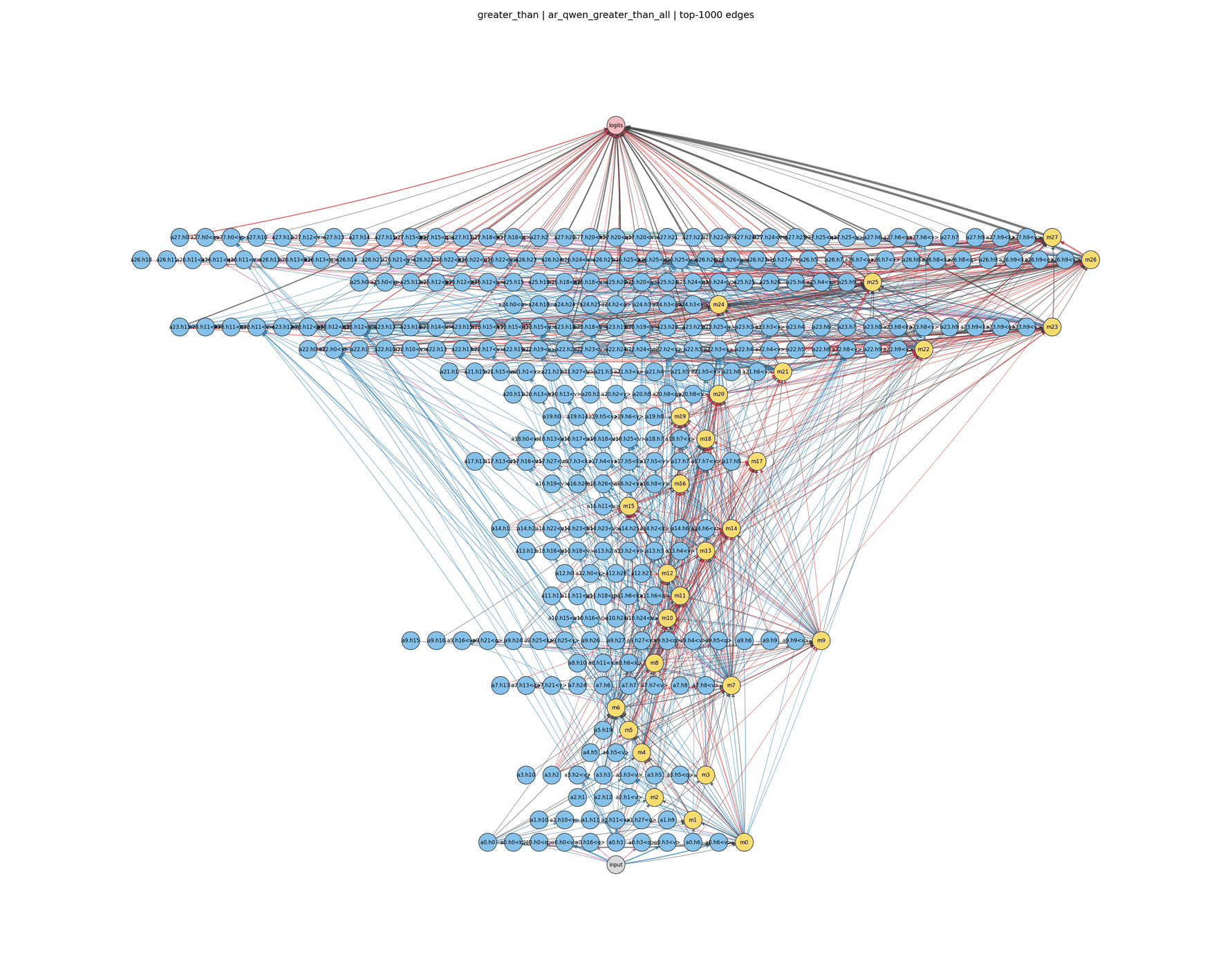}
        \caption*{(g) Greater-Than — Qwen-2.5-7B}
    \end{minipage}
    \hfill
    \begin{minipage}{0.48\linewidth}
        \centering
        \includegraphics[height=0.20\textheight,keepaspectratio]{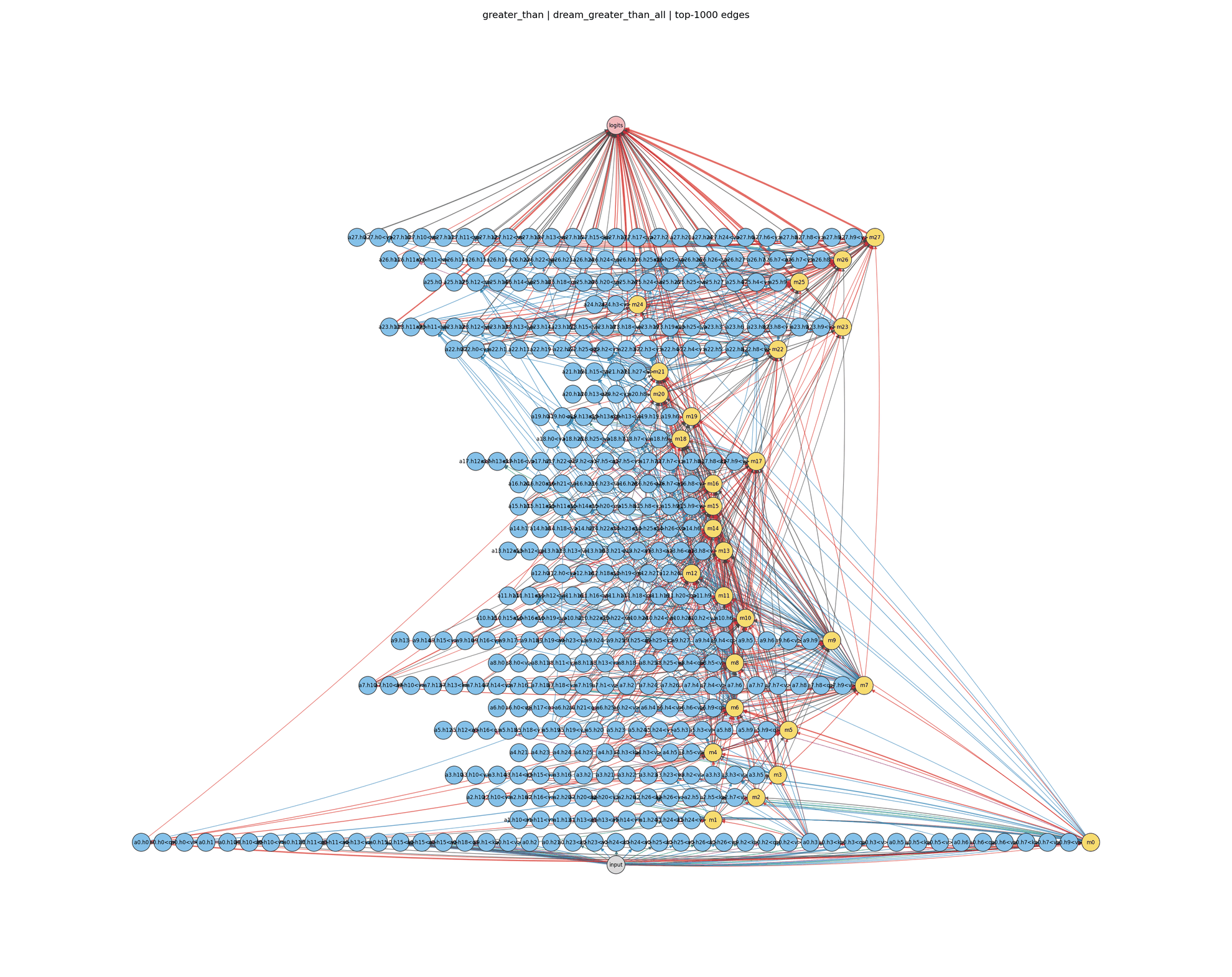}
        \caption*{(h) Greater-Than — Dream-Base-7B}
    \end{minipage}

    \caption{Circuit comparison across tasks and architectures. Left: Autoregressive (Qwen-2.5-7B). Right: Masked Diffusion (Dream-Base-7B). Rows from top to bottom: \textsc{IOI}, \textsc{Countdown}, \textsc{Semantic Infilling}, and \textsc{Greater-Than}.}
    \label{fig:circuit_comparison_expanded}
\end{figure*}


\begin{figure*}[!t]
    \centering

    \begin{minipage}{0.48\linewidth}
        \centering
        \includegraphics[height=0.20\textheight,keepaspectratio]{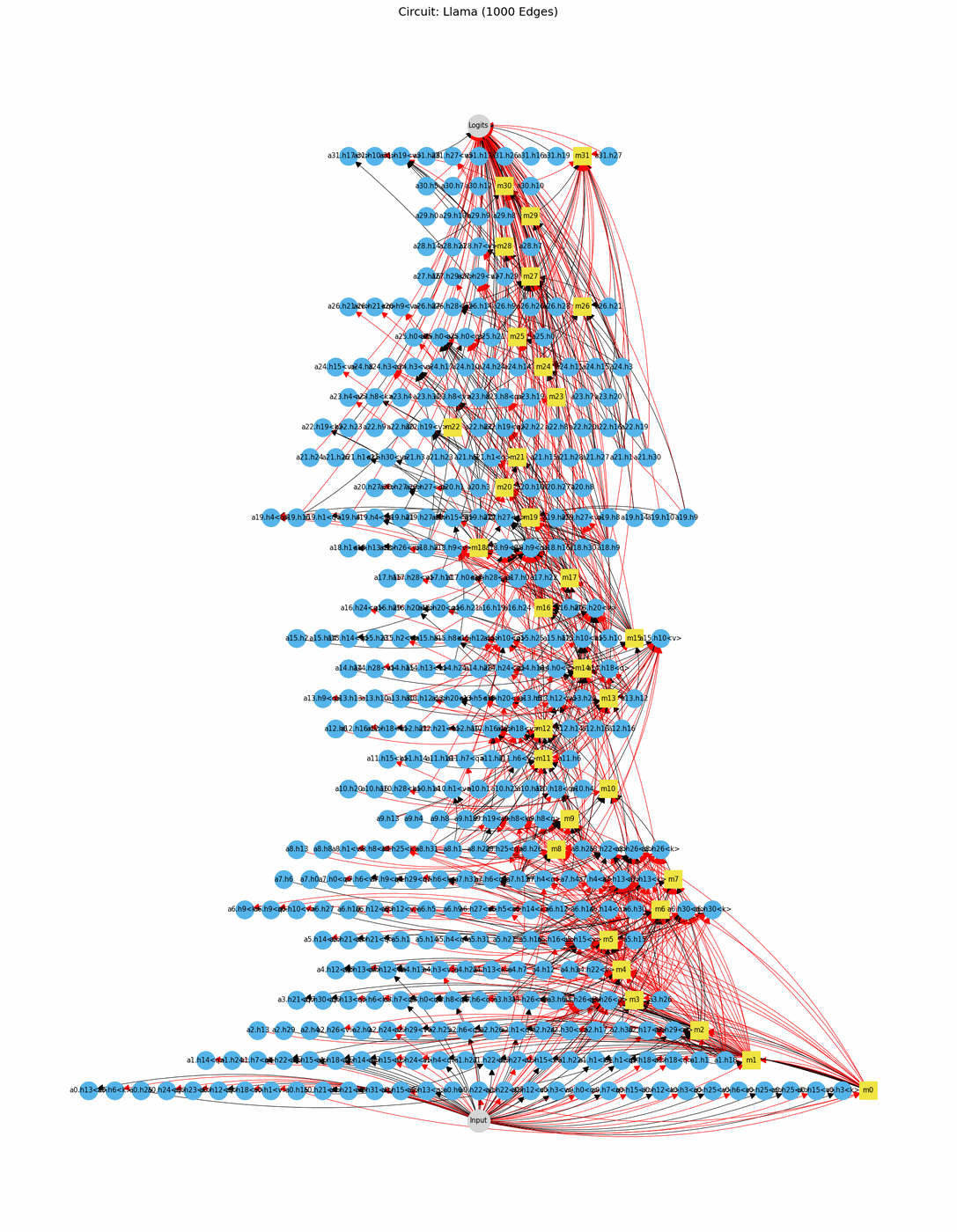}
        \caption*{(a) IOI — LLaMA-2-7B}
    \end{minipage}
    \hfill
    \begin{minipage}{0.48\linewidth}
        \centering
        \includegraphics[height=0.20\textheight,keepaspectratio]{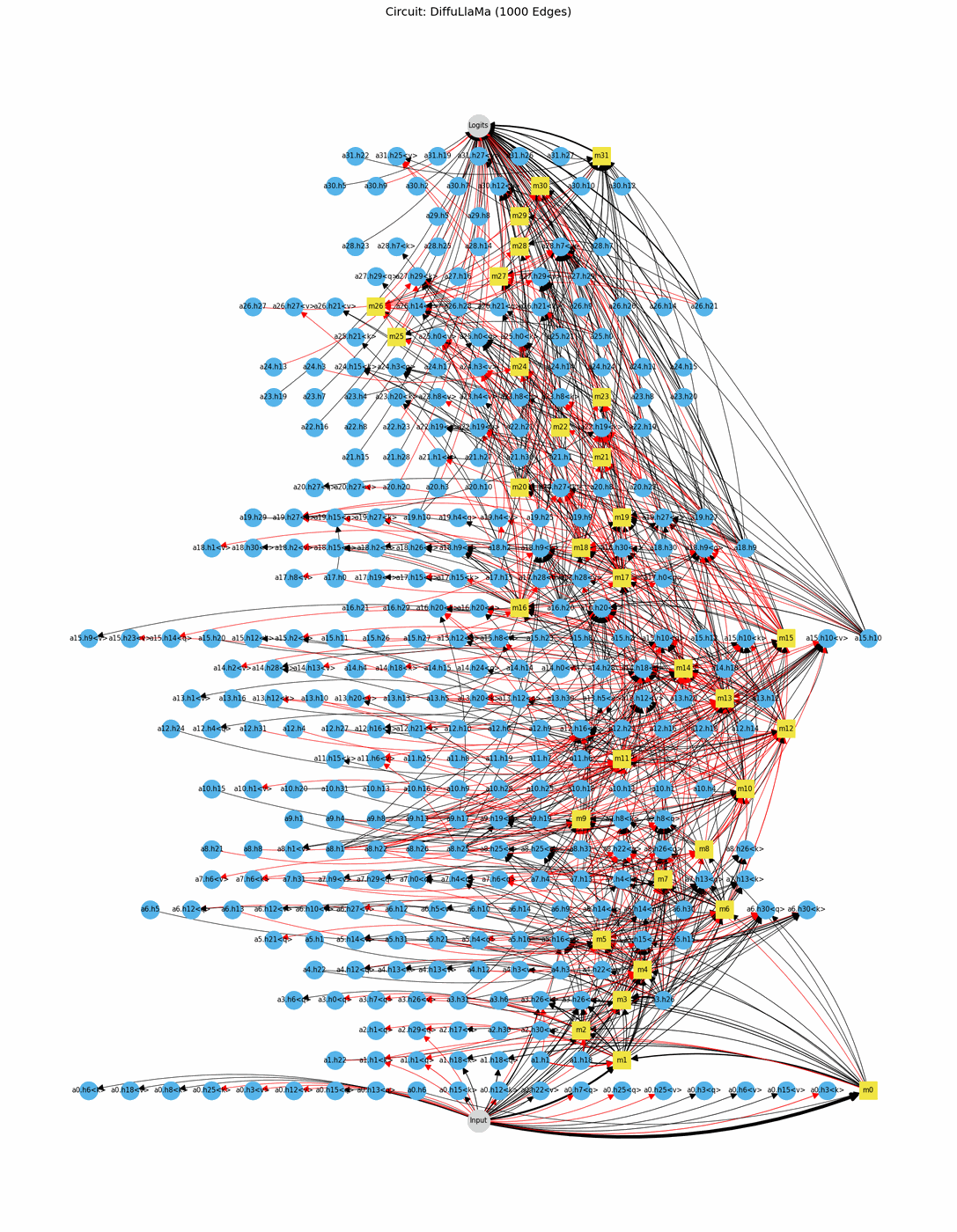}
        \caption*{(b) IOI — DiffuLLaMA-7B}
    \end{minipage}

    \vspace{0.3em}

    \begin{minipage}{0.48\linewidth}
        \centering
        \includegraphics[height=0.20\textheight,keepaspectratio]{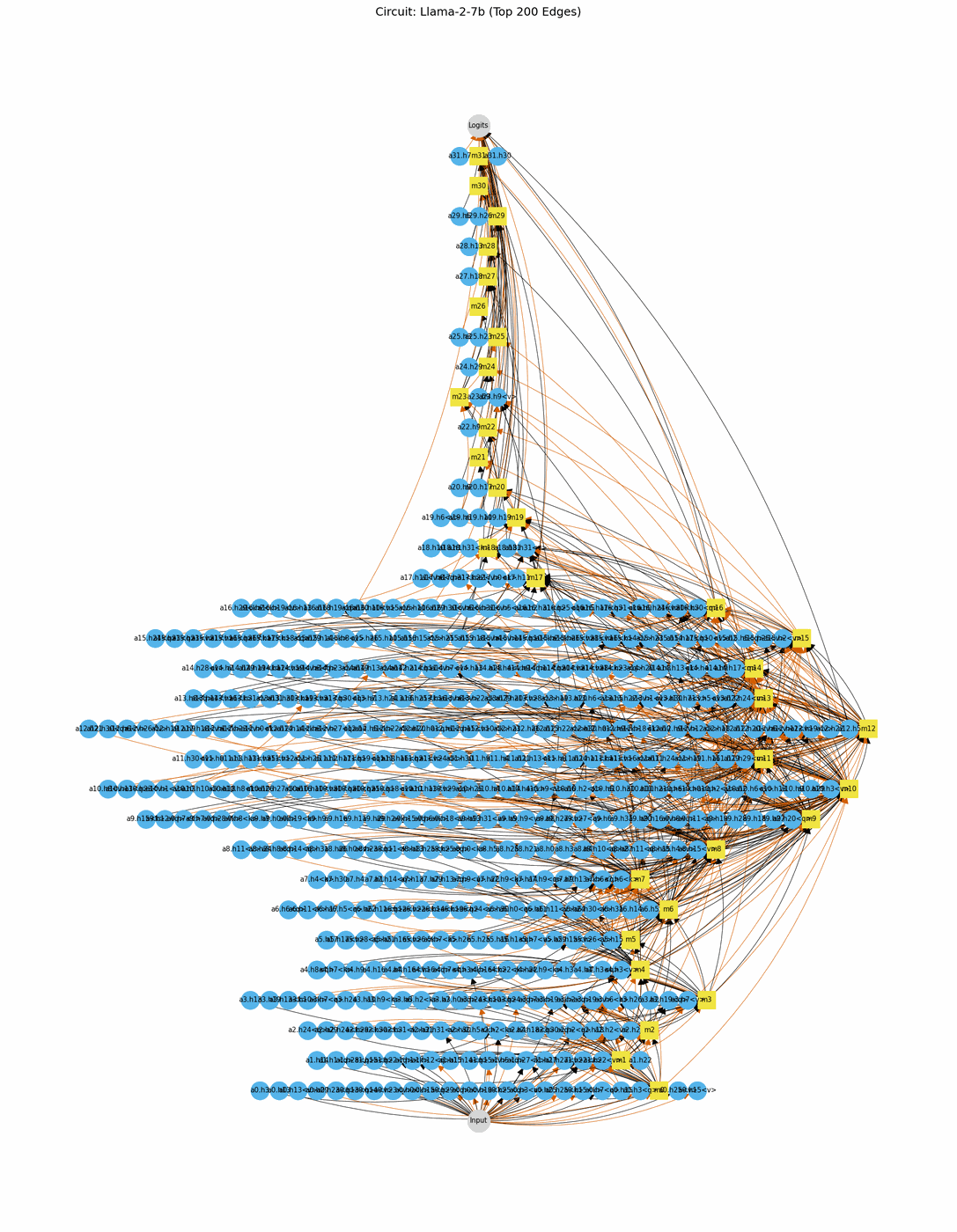}
        \caption*{(c) Countdown — LLaMA-2-7B}
    \end{minipage}
    \hfill
    \begin{minipage}{0.48\linewidth}
        \centering
        \includegraphics[height=0.20\textheight,keepaspectratio]{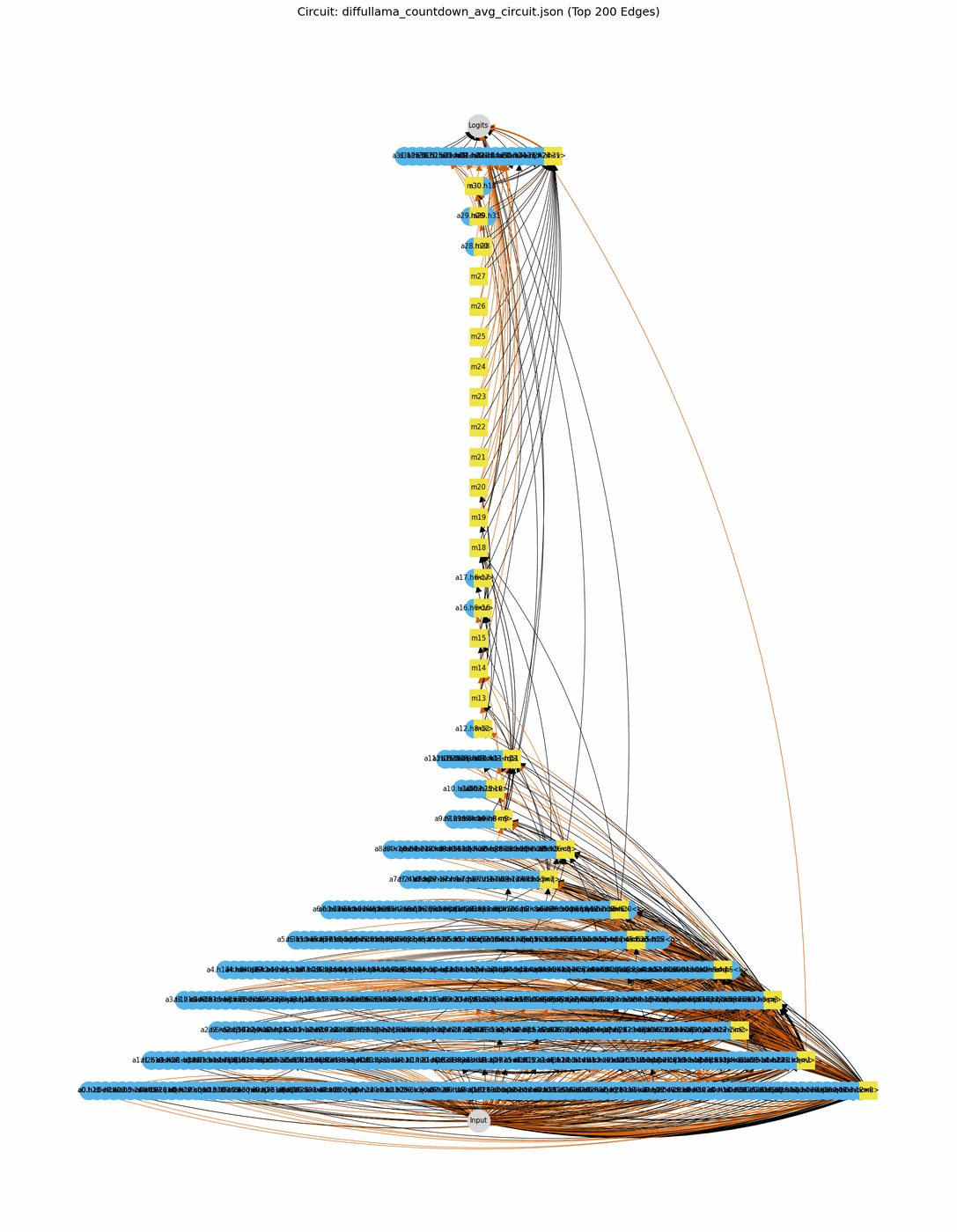}
        \caption*{(d) Countdown — DiffuLLaMA-7B}
    \end{minipage}

    \vspace{0.3em}

    \begin{minipage}{0.48\linewidth}
        \centering
        \includegraphics[height=0.20\textheight,keepaspectratio]{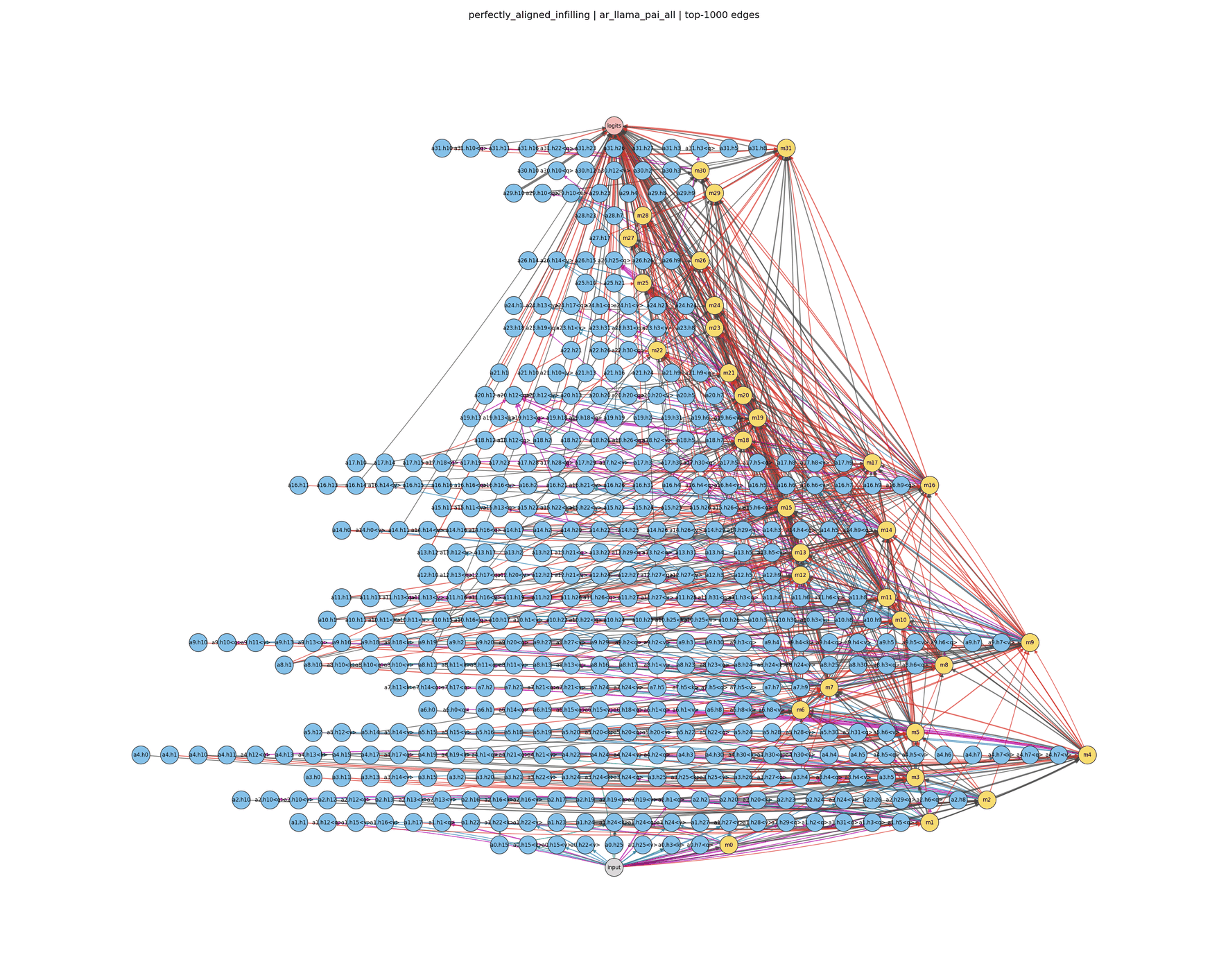}
        \caption*{(e) Semantic Infilling — LLaMA-2-7B}
    \end{minipage}
    \hfill
    \begin{minipage}{0.48\linewidth}
        \centering
        \includegraphics[height=0.20\textheight,keepaspectratio]{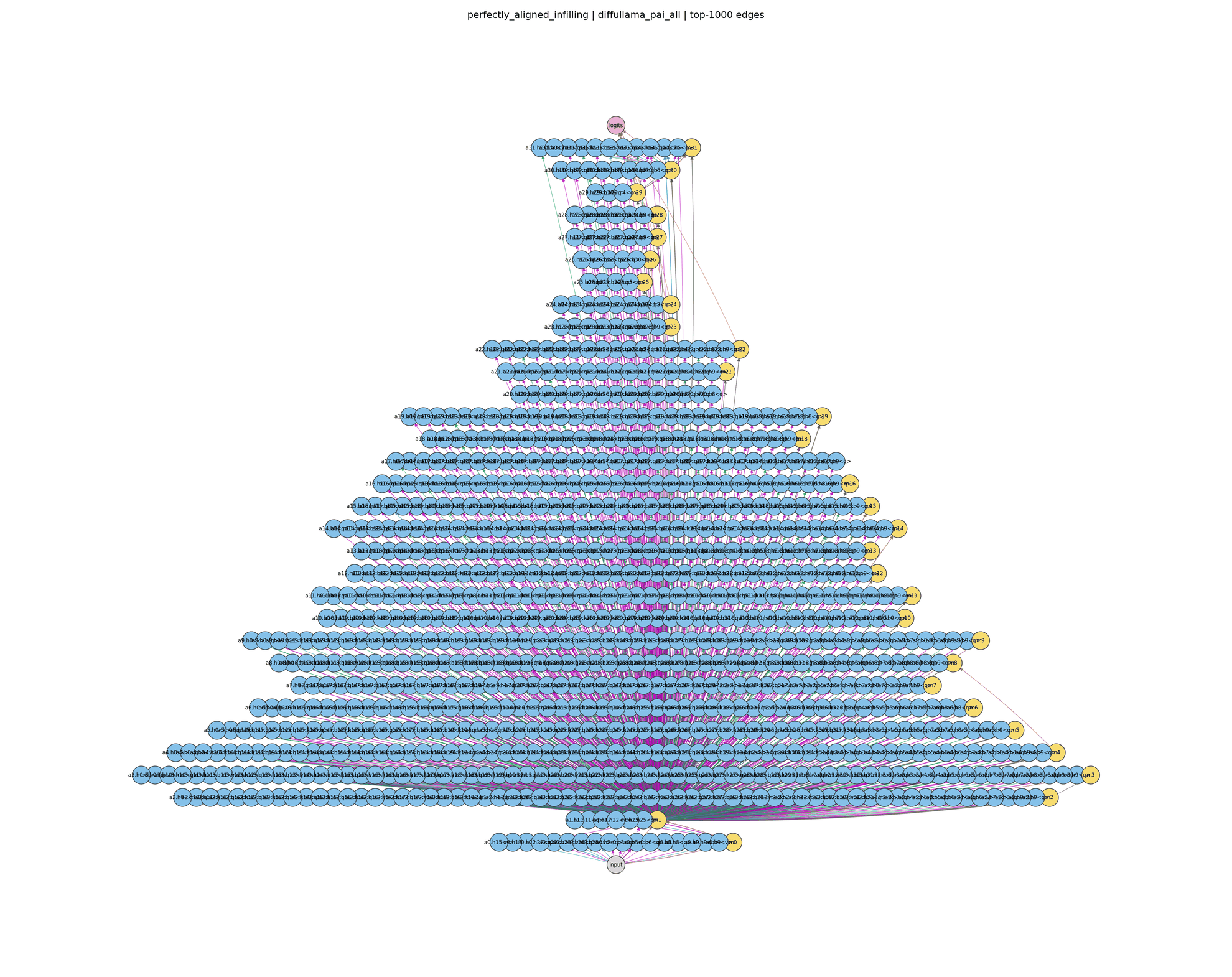}
        \caption*{(f) Semantic Infilling — DiffuLLaMA-7B}
    \end{minipage}

    \vspace{0.3em}

    \begin{minipage}{0.48\linewidth}
        \centering
        \includegraphics[height=0.20\textheight,keepaspectratio]{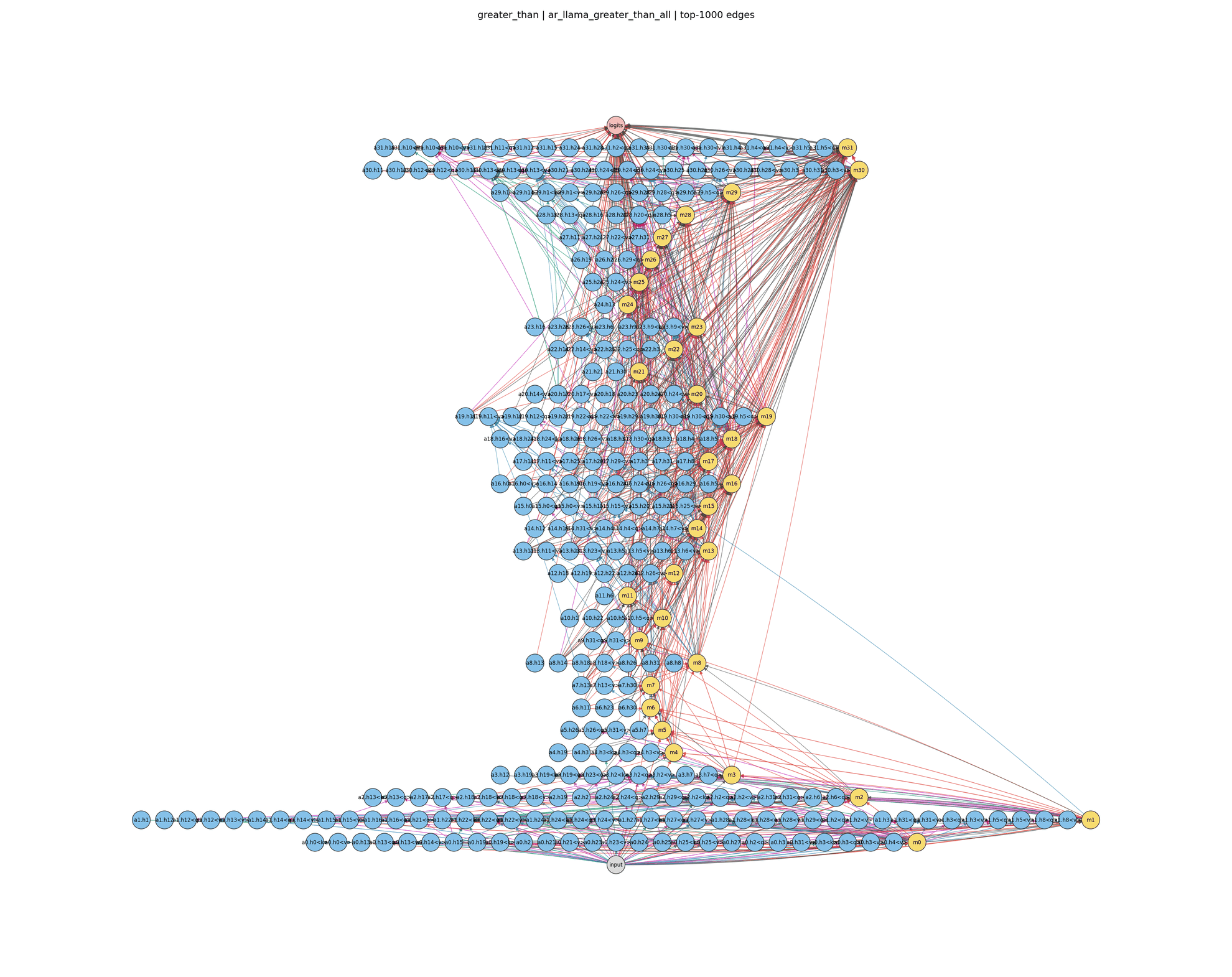}
        \caption*{(g) Greater-Than — LLaMA-2-7B}
    \end{minipage}
    \hfill
    \begin{minipage}{0.48\linewidth}
        \centering
        \includegraphics[height=0.20\textheight,keepaspectratio]{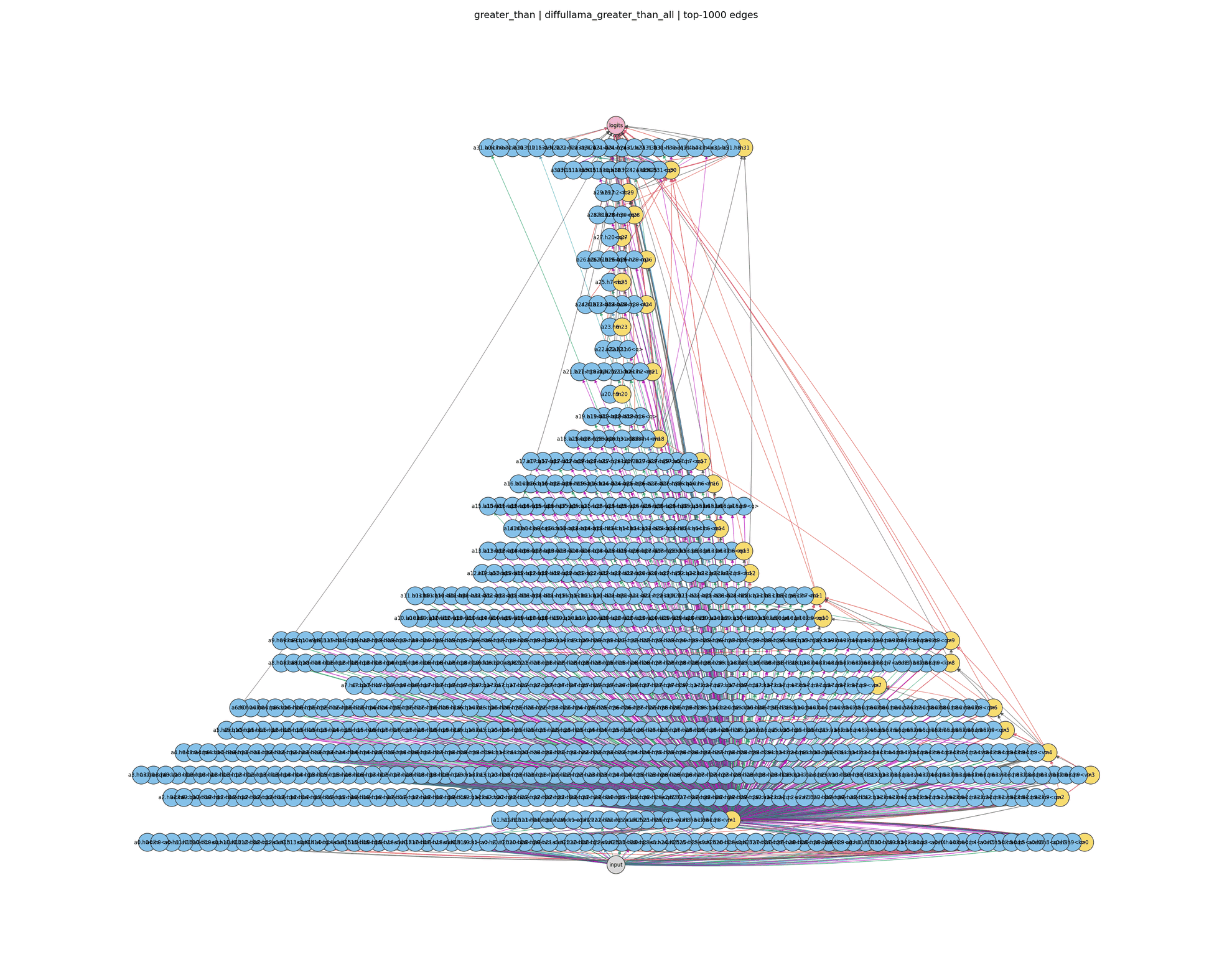}
        \caption*{(h) Greater-Than — DiffuLLaMA-7B}
    \end{minipage}

    \caption{Circuit comparison across tasks and architectures. Left: Autoregressive (LLaMA-2-7B). Right: Masked Diffusion (DiffuLLaMA-7B). Rows from top to bottom: IOI, \textsc{Countdown}, \textsc{Semantic Infilling}, and \textsc{Greater-Than}.}
    \label{fig:big-circuit-expanded}
\end{figure*}



\begin{figure*}[!t]
    \centering
    \hfil
    \includegraphics[width=0.24\linewidth]{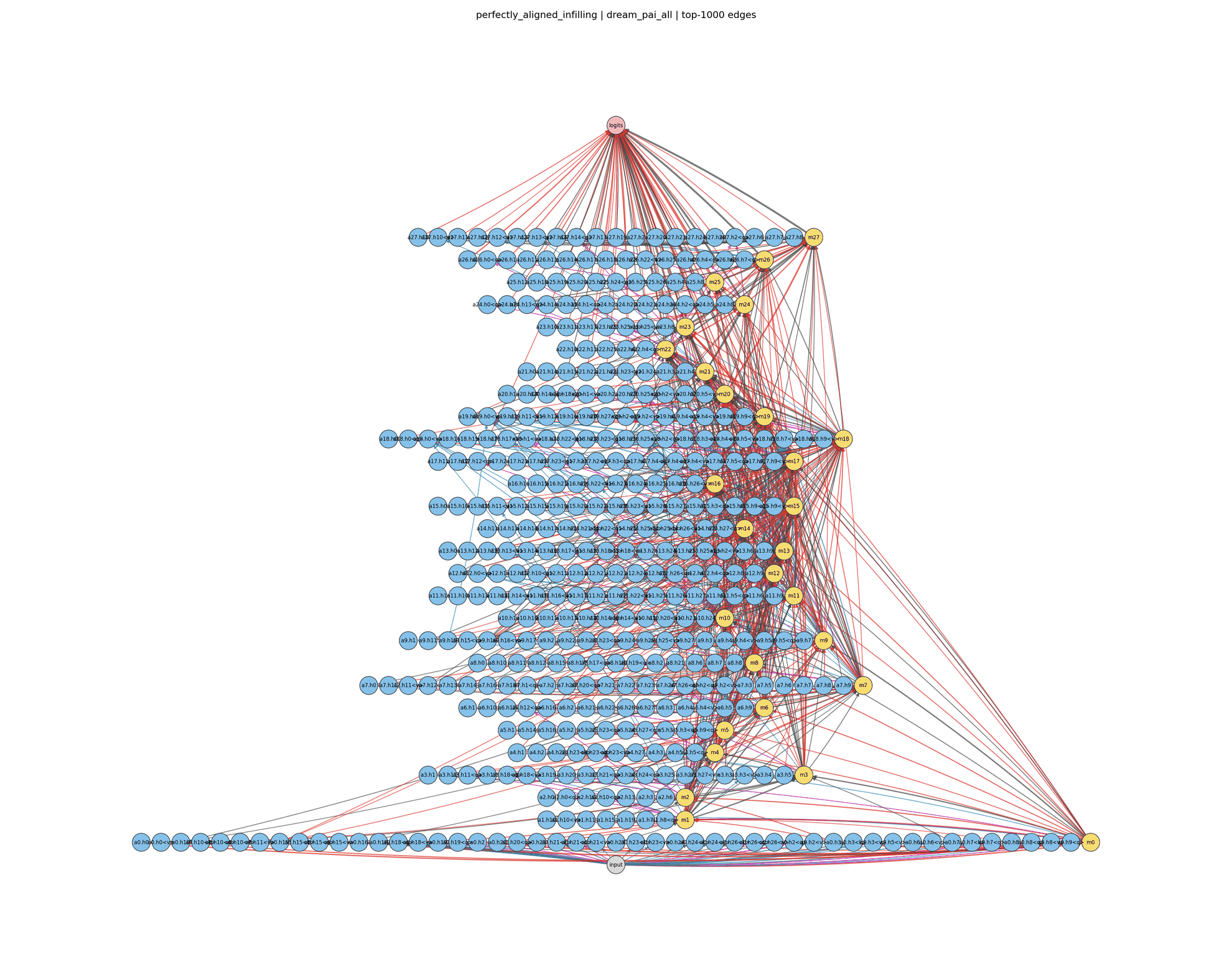}\hfil
    \includegraphics[width=0.24\linewidth]{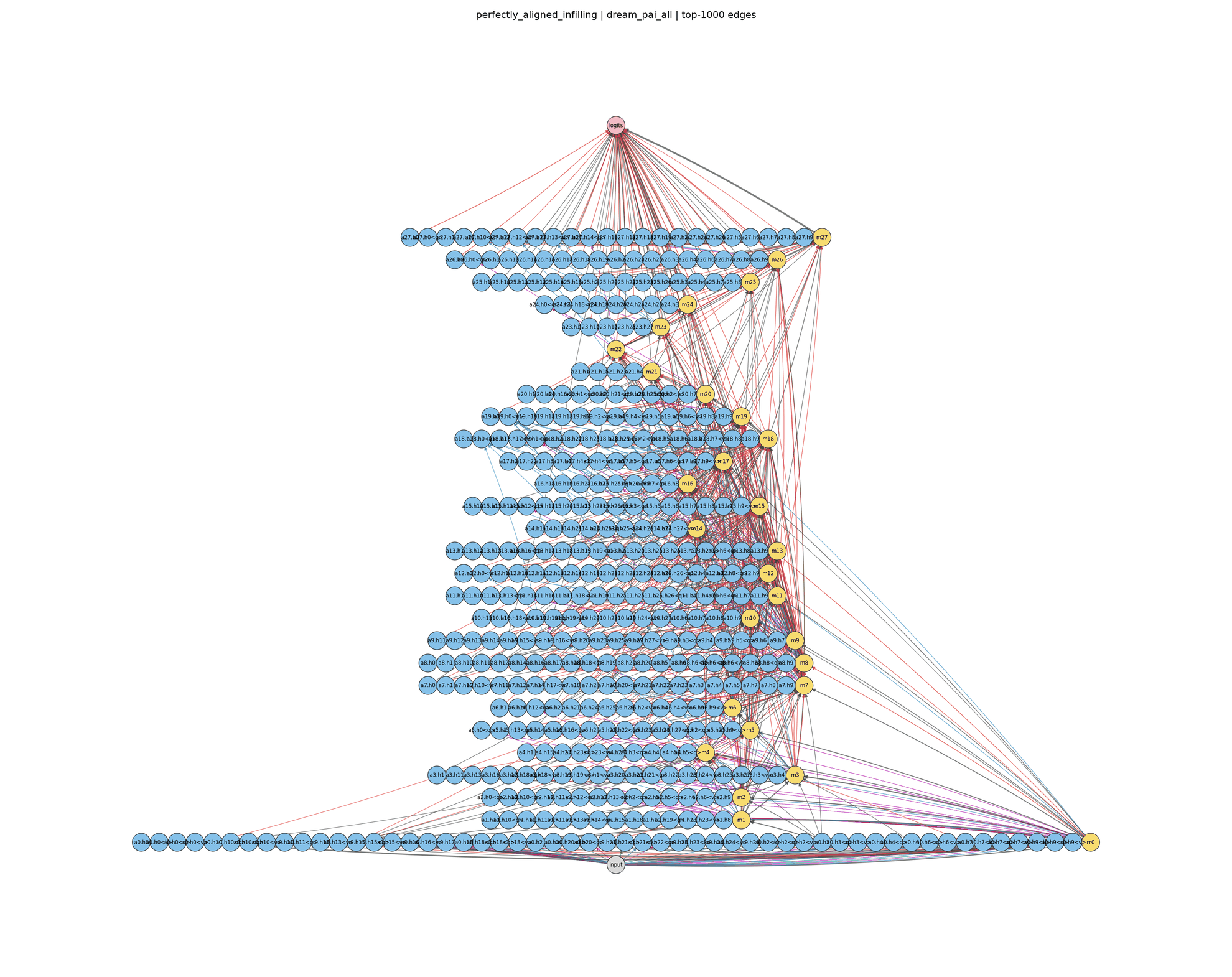}\hfil
    \includegraphics[width=0.24\linewidth]{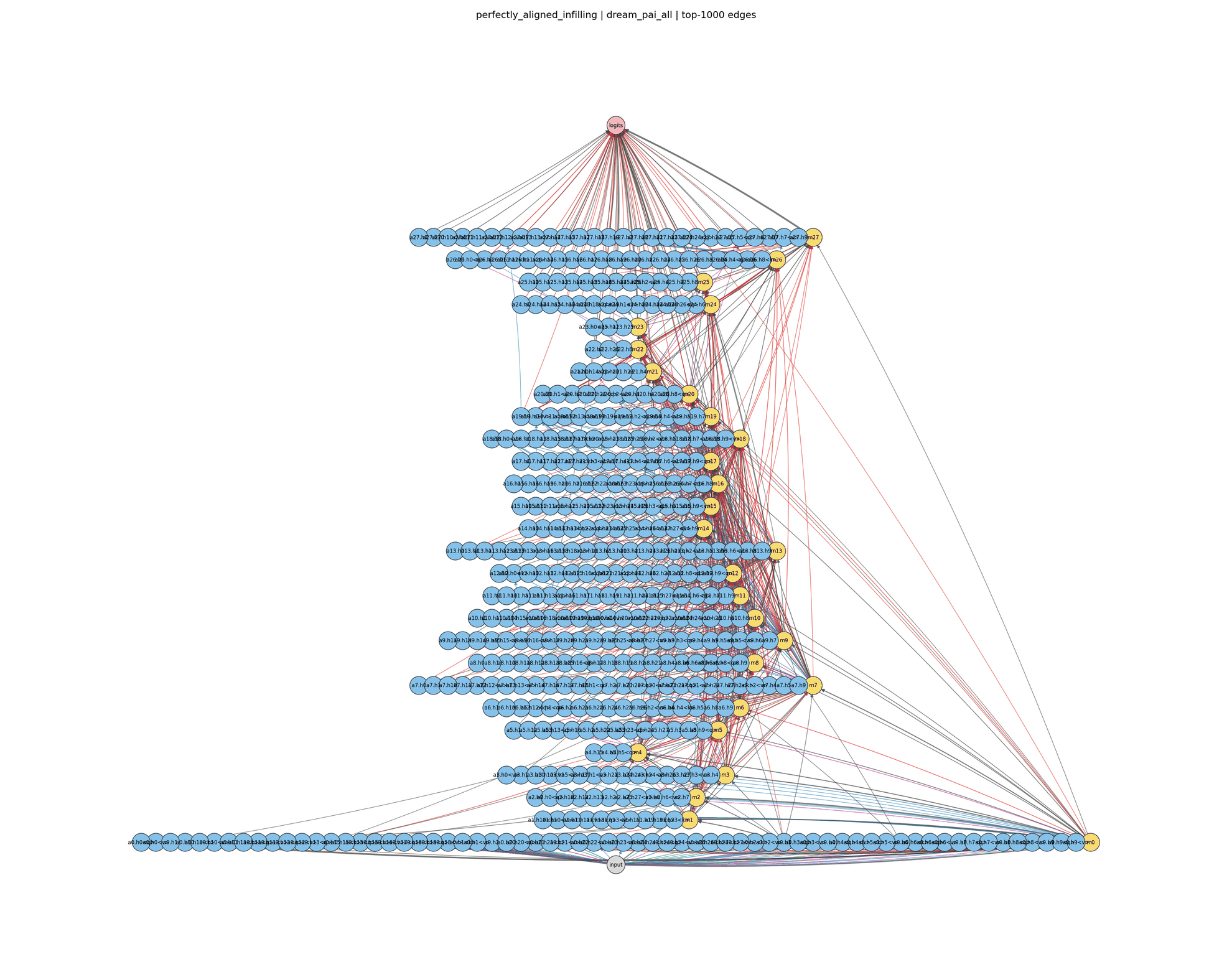}\hfil
    \caption{
    Step-wise circuit visualization of Dream-Base-7B on the Semantic Infilling task. 
    The three panels display the extracted task circuits across its generation intervals from left to right.
    }
    \label{fig:dream-si-steps}
\end{figure*}

\begin{figure*}[!t]
    \centering
    \includegraphics[width=0.24\linewidth]{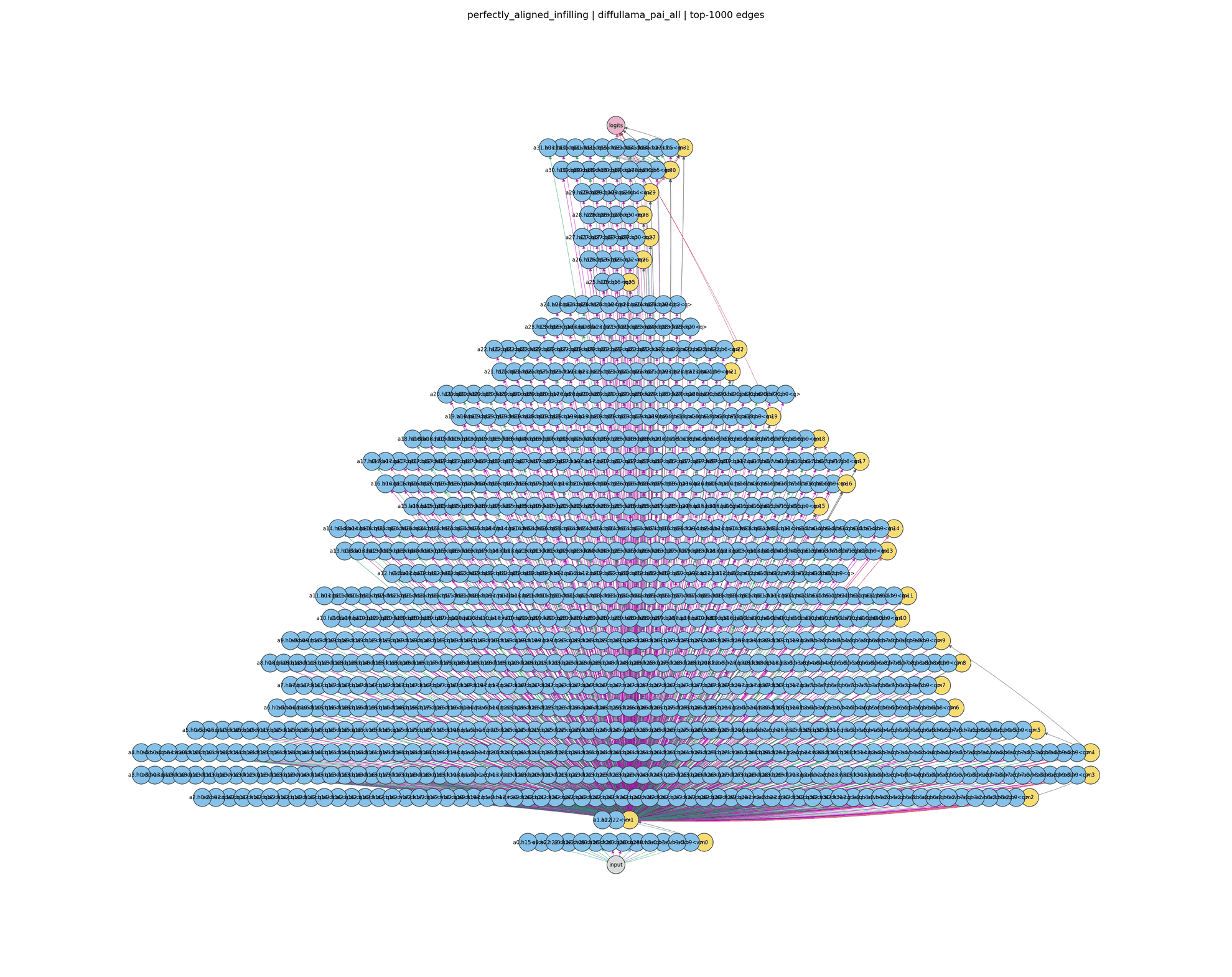}\hfill
    \includegraphics[width=0.24\linewidth]{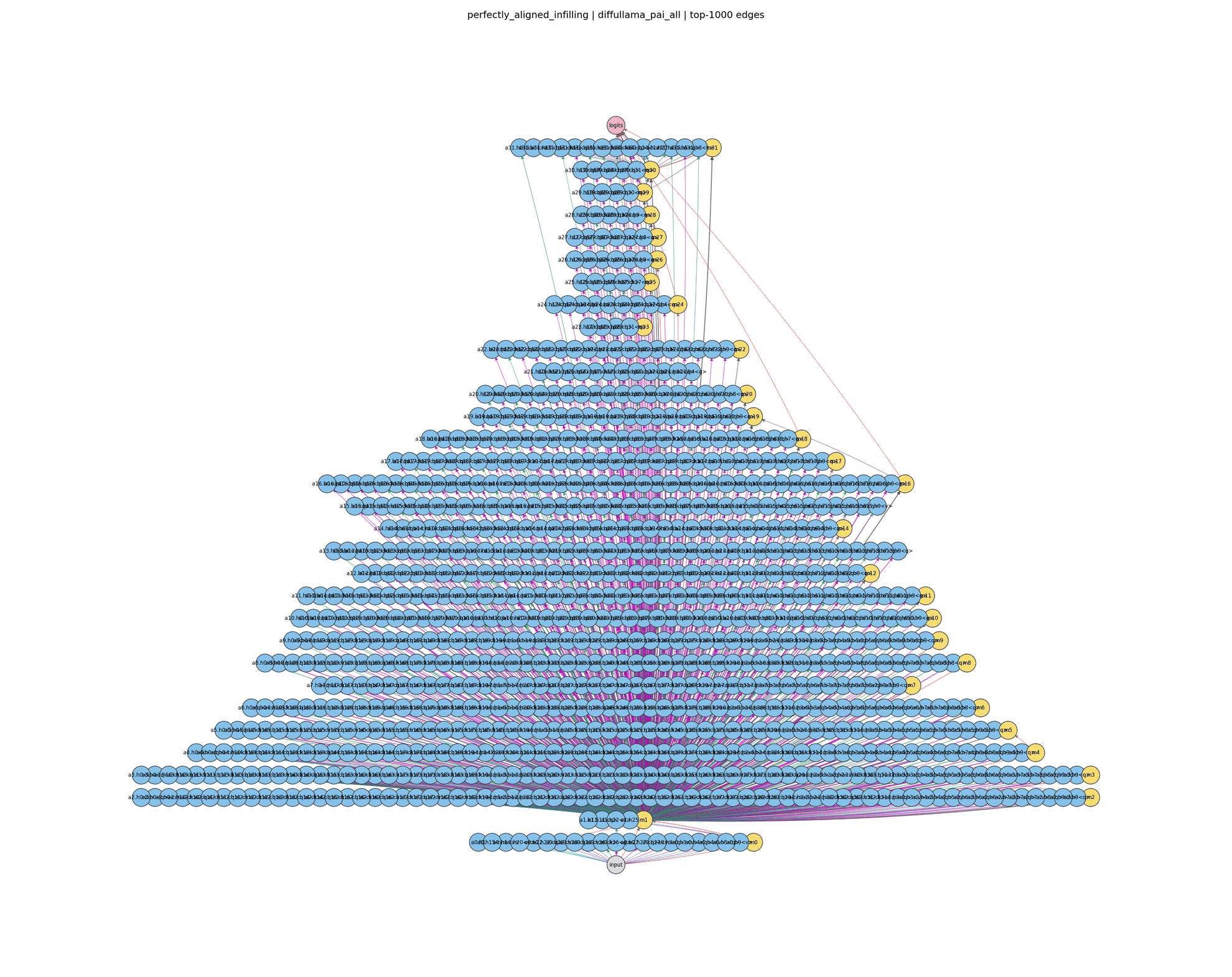}\hfill
    \includegraphics[width=0.24\linewidth]{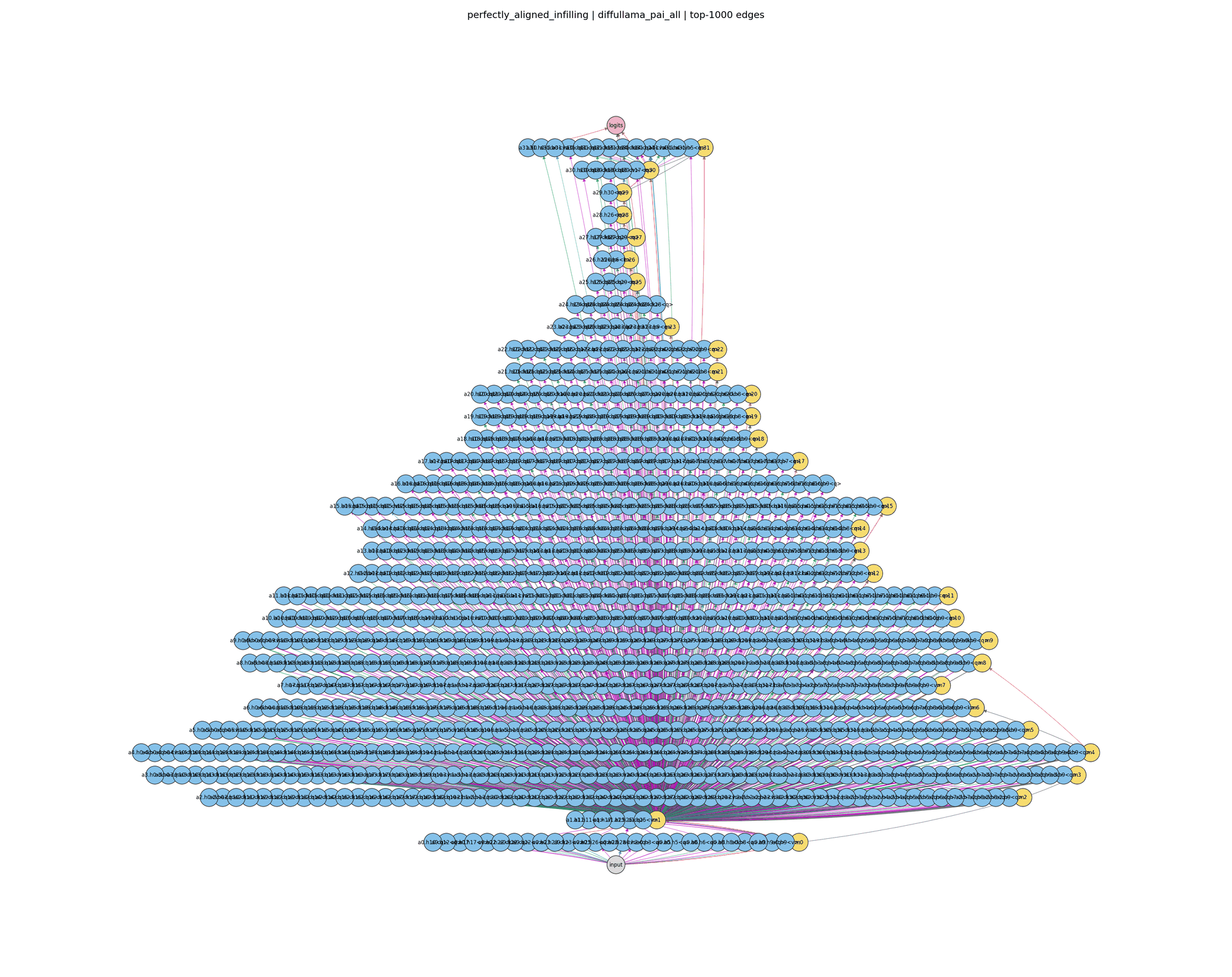}\hfill
    \includegraphics[width=0.24\linewidth]{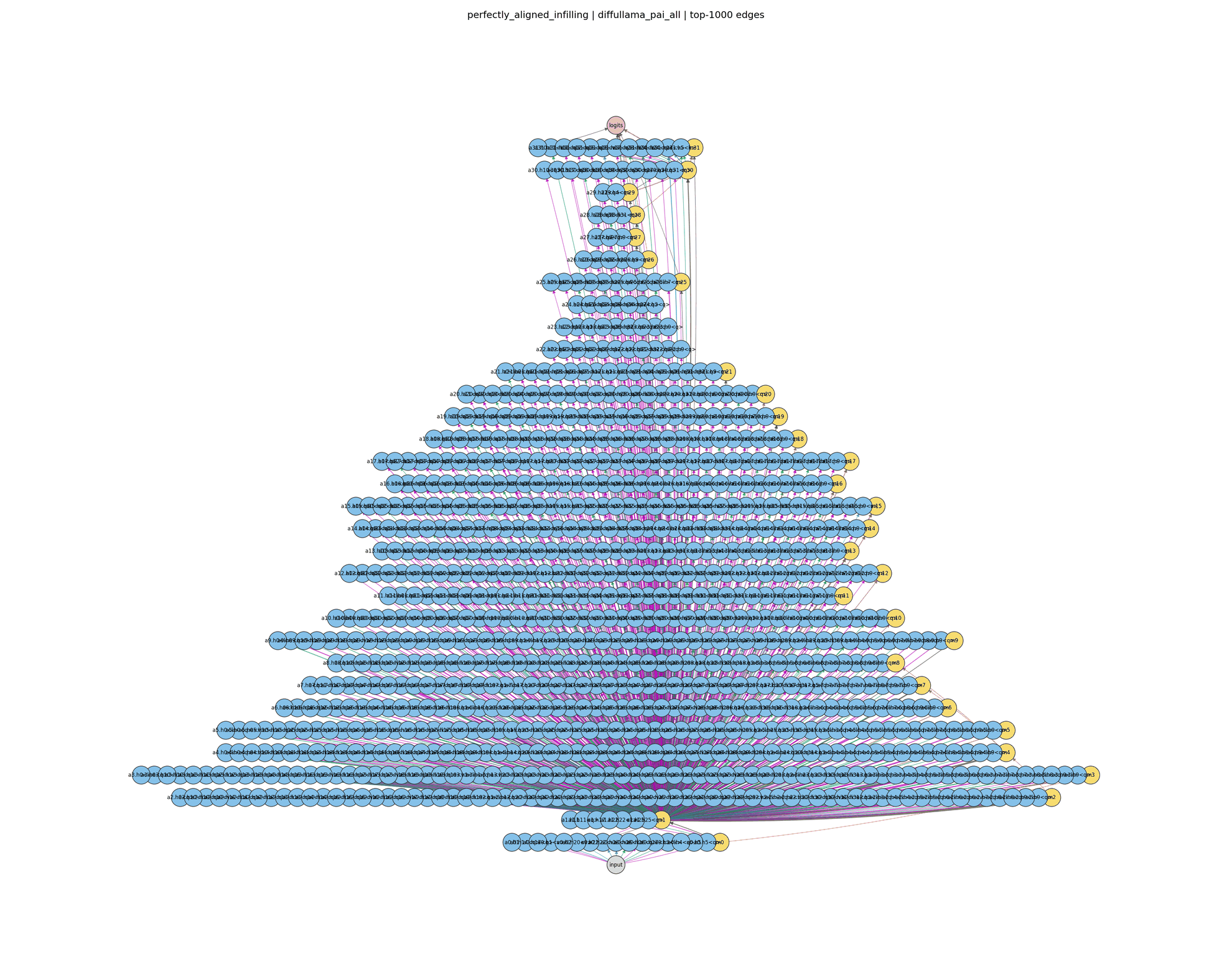}
    \caption{
    Step-wise circuit visualization of DiffuLLaMA-7B on the Semantic Infilling task. 
    The four panels display the extracted task circuits across its generation intervals from left to right.
    }
    \label{fig:diffullama-si-steps}
\end{figure*}


\clearpage

\begin{table*}[t]
    \centering
    \renewcommand{\arraystretch}{1.3}
    \caption{Top 30 Source Components for \textbf{IOI} Task}
    \label{tab:components_ioi}
    \begin{tabular}{l >{\raggedright\arraybackslash}p{13cm}}
        \toprule
        \textbf{Model} & \textbf{Top Source Components (Edges)} \\
        \midrule
        LLaMA-2 & input$\to$m0, input$\to$m1, a3.h26$\to$m3, a26.h21$\to$logits, m1$\to$a3.h26$<q>$, a1.h18$\to$m1, input$\to$a3.h26$<q>$, a25.h0$\to$logits, m27$\to$logits, a27.h29$\to$logits, m0$\to$m1, m1$\to$a3.h26$<k>$, a5.h15$\to$a6.h30$<k>$, input$\to$a3.h26$<k>$, a24.h3$\to$logits, a23.h20$\to$logits, a21.h30$\to$logits, a1.h1$\to$m1, m2$\to$a3.h26$<k>$, m0$\to$a3.h26$<q>$, a18.h9$\to$logits, a21.h1$\to$logits, m2$\to$a3.h26$<q>$, m4$\to$a6.h30$<q>$, m0$\to$a3.h26$<k>$, m0$\to$a1.h18$<q>$, input$\to$a1.h18$<q>$, a1.h1$\to$a3.h26$<q>$, a24.h15$\to$logits, a20.h8$\to$logits \\
        \addlinespace
        Qwen & a24.h24$\to$logits, a26.h15$\to$logits, a23.h11$\to$logits, a27.h21$\to$logits, a27.h4$\to$logits, m27$\to$logits, a27.h1$\to$logits, a26.h26$\to$logits, a26.h22$\to$logits, m20$\to$m22, a27.h18$\to$logits, a27.h17$\to$logits, a27.h5$\to$logits, a27.h14$\to$logits, a24.h23$\to$logits, m24$\to$logits, input$\to$a0.h10$<v>$, a27.h3$\to$logits, a20.h24$\to$m22, a27.h24$\to$logits, a17.h24$\to$a20.h24$<v>$, a26.h5$\to$logits, input$\to$a0.h3$<v>$, a27.h4$\to$m27, m25$\to$logits, a25.h24$\to$logits, a18.h25$\to$a20.h24$<v>$, a18.h27$\to$a20.h24$<v>$, a26.h2$\to$logits, m22$\to$a25.h25$<q>$ \\
        \addlinespace
        DiffuLLaMA & input$\to$m0, input$\to$m1, a3.h26$\to$m3, a28.h7$\to$logits, m31$\to$logits, a26.h21$\to$logits, m0$\to$m1, a1.h18$\to$m1, a27.h29$\to$logits, m1$\to$a3.h26$<q>$, a1.h1$\to$m1, a5.h15$\to$a6.h30$<k>$, m1$\to$a3.h26$<k>$, m1$\to$m4, input$\to$a3.h26$<q>$, input$\to$a3.h26$<k>$, a26.h14$\to$logits, m2$\to$a3.h26$<k>$, a30.h12$\to$logits, m4$\to$a6.h30$<q>$, m0$\to$a3.h26$<q>$, a23.h20$\to$logits, a18.h9$\to$a28.h7$<v>$, input$\to$a1.h18$<q>$, m0$\to$a1.h18$<q>$, input$\to$m4, m2$\to$a3.h26$<q>$, a22.h19$\to$logits, input$\to$m2, a27.h29$\to$a28.h7$<v>$ \\
        \addlinespace
        Dream & a24.h24$\to$logits, m19$\to$m20, a23.h10$\to$logits, m18$\to$m19, m19$\to$m21, m9$\to$m19, m21$\to$m22, a18.h25$\to$m20, m14$\to$m15, m25$\to$logits, m12$\to$m18, m7$\to$m17, a15.h20$\to$m16, m9$\to$m20, m10$\to$m20, a15.h20$\to$m20, m12$\to$m15, m12$\to$m13, m18$\to$m22, m17$\to$m20, m14$\to$m22, a15.h20$\to$m19, a15.h23$\to$m16, m8$\to$m17, m11$\to$m16, m21$\to$logits, a25.h24$\to$logits, a18.h25$\to$m21, m8$\to$m20, m11$\to$m20 \\
        \bottomrule
    \end{tabular}
\end{table*}

\begin{table*}[t]
    \centering
    \renewcommand{\arraystretch}{1.3}
    \caption{Top 30 Source Components for \textbf{COUNTDOWN} Task}
    \label{tab:components_countdown}
    \begin{tabular}{l >{\raggedright\arraybackslash}p{13cm}}
        \toprule
        \textbf{Model} & \textbf{Top Source Components (Edges)} \\
        \midrule
        LLaMA-2 & input$\to$a0.h15$<v>$, m10$\to$m11, m11$\to$m12, a1.h22$\to$m1, m6$\to$m11, m11$\to$m15, input$\to$a0.h25$<v>$, m0$\to$a1.h22$<v>$, m29$\to$logits, m7$\to$m9, m0$\to$m2, m28$\to$logits, m8$\to$m12, m8$\to$a11.h29$<v>$, m8$\to$m11, a2.h2$\to$m3, m3$\to$m5, m14$\to$m15, m0$\to$m1, input$\to$a0.h3$<q>$, m13$\to$m15, a12.h5$\to$m13, m7$\to$m10, a12.h22$\to$m12, a5.h15$\to$a7.h6$<k>$, input$\to$a2.h2$<v>$, m24$\to$logits, m0$\to$a1.h22$<k>$, m12$\to$m14, m7$\to$a8.h15$<v>$ \\
        \addlinespace
        Qwen & m26$\to$logits, m25$\to$logits, m27$\to$logits, input$\to$a0.h3$<v>$, a23.h11$\to$logits, a25.h12$\to$logits, a26.h22$\to$logits, m24$\to$logits, a26.h24$\to$logits, m26$\to$m27, a24.h23$\to$logits, m21$\to$a23.h11$<v>$, a0.h3$\to$m0, a22.h13$\to$logits, a26.h23$\to$logits, a23.h11$\to$m25, m25$\to$m27, a23.h19$\to$logits, a25.h12$\to$m27, a26.h25$\to$logits, m21$\to$a25.h12$<v>$, m20$\to$a23.h11$<v>$, a26.h26$\to$logits, m25$\to$a26.h22$<v>$, a23.h11$\to$m26, a26.h22$\to$m27, a26.h22$\to$m26, a26.h24$\to$m27, a23.h11$\to$a26.h22$<v>$, a24.h23$\to$m25 \\
        \addlinespace
        DiffuLLaMA & m1$\to$m2, input$\to$m0, m1$\to$m3, input$\to$a0.h12$<k>$, m31$\to$logits, m1$\to$a4.h5$<k>$, input$\to$a0.h3$<k>$, input$\to$a0.h15$<v>$, m1$\to$a3.h3$<q>$, m1$\to$a3.h27$<q>$, input$\to$m1, input$\to$a0.h0$<k>$, input$\to$a0.h3$<q>$, input$\to$a0.h1$<v>$, input$\to$a1.h1$<v>$, m1$\to$a3.h26$<k>$, m1$\to$a3.h7$<k>$, m1$\to$a3.h8$<k>$, m1$\to$m4, m0$\to$m1, m1$\to$a4.h5$<q>$, input$\to$a0.h13$<q>$, m1$\to$a2.h2$<k>$, m0$\to$m3, m1$\to$a3.h0$<q>$, input$\to$a0.h3$<v>$, m1$\to$a6.h20$<k>$, input$\to$a0.h13$<v>$, m1$\to$a3.h17$<k>$, m1$\to$a5.h23$<q>$ \\
        \addlinespace
        Dream & m25$\to$logits, m27$\to$logits, m26$\to$logits, input$\to$a0.h15$<q>$, a0.h10$\to$m0, input$\to$a0.h3$<v>$, m24$\to$logits, input$\to$a0.h10$<v>$, m23$\to$logits, input$\to$a0.h11$<q>$, input$\to$a0.h15$<k>$, input$\to$a0.h15$<v>$, a0.h3$\to$m0, a25.h12$\to$logits, m0$\to$m1, m26$\to$m27, input$\to$a0.h0$<v>$, input$\to$a0.h11$<v>$, a27.h11$\to$logits, a26.h22$\to$logits, a26.h25$\to$logits, m22$\to$logits, m21$\to$m27, a0.h15$\to$m0, input$\to$a0.h11$<k>$, m25$\to$m26, m22$\to$m27, a25.h12$\to$m27, input$\to$a0.h10$<k>$, a26.h24$\to$logits \\
        \bottomrule
    \end{tabular}
\end{table*}

\clearpage

\onecolumn
\begin{longtable}{ll l >{\raggedright\arraybackslash}p{9cm}}
\caption{Top interpretable tokens for high-attribution components (excluding components stated in table ~\ref{tab:component-roles}). Components are sorted by confidence (probability of the top token).} \label{tab:logit_lens} \\
\toprule
\textbf{Task} & \textbf{Model} & \textbf{Comp.} & \textbf{Top Tokens (Probability)} \\
\midrule
\endfirsthead
\multicolumn{4}{c}{{\bfseries \tablename\ \thetable{} -- continued from previous page}} \\
\toprule
\textbf{Task} & \textbf{Model} & \textbf{Comp.} & \textbf{Top Tokens (Probability)} \\
\midrule
\endhead
\hline \multicolumn{4}{r}{{Continued on next page}} \\
\endfoot
\bottomrule
\endlastfoot
\textbf{Countdown} & \textbf{DiffuLLaMA} & m1 & \texttt{sierp} (0.936), \texttt{kwiet} (0.045), \texttt{Hinweis} (0.004) \\
 &  & m2 & \texttt{(U+207B)} (5.6e-05), \texttt{nahm} (5.6e-05), \texttt{Hinweis} (5.5e-05) \\
 &  & m3 & \texttt{iftung} (4.1e-05), \texttt{(U+043D) (U+0434) (U+0434)} (4.1e-05), \texttt{iation} (4.1e-05) \\
 &  & input & \texttt{rd} (3.7e-05), \texttt{iation} (3.7e-05), \texttt{ness} (3.7e-05) \\
 &  & m0 & \texttt{mes} (3.5e-05), \texttt{led} (3.5e-05), \texttt{med} (3.5e-05) \\
\addlinespace
 & \textbf{Dream} & m25 & \texttt{ out} (0.012), \texttt{))} (0.001), \texttt{ from} (0.001) \\
 &  & m26 & \texttt{,-} (7.9e-04), \texttt{ A} (7.3e-04), \texttt{ S} (6.3e-04) \\
 &  & m22 & \texttt{ nothing} (8.7e-05), \texttt{ H} (8.6e-05), \texttt{ a} (7.2e-05) \\
 &  & m24 & \texttt{ int} (8.3e-05), \texttt{i} (7.3e-05), \texttt{ commemor} (6.5e-05) \\
 &  & m21 & \texttt{ =} (3.2e-05), \texttt{ by} (3.1e-05), \texttt{ C} (3.0e-05) \\
 &  & a27.h11 & \texttt{ doen} (2.9e-05), \texttt{uteur} (2.6e-05), \texttt{ retali} (2.4e-05) \\
 &  & a25.h12 & \texttt{7} (2.2e-05), \texttt{8} (1.9e-05), \texttt{3} (1.7e-05) \\
 &  & a26.h22 & \texttt{ cosy} (2.0e-05), \texttt{ W} (1.9e-05), \texttt{-ok} (1.8e-05) \\
 &  & a26.h25 & \texttt{ fourth} (1.4e-05), \texttt{five} (1.4e-05), \texttt{IV} (1.4e-05) \\
 &  & a26.h24 & \texttt{ist} (1.2e-05), \texttt{/S} (1.2e-05), \texttt{ SIX} (1.2e-05) \\
 &  & m1 & \texttt{.} (1.0e-05), \texttt{1} (1.0e-05), \texttt{\#} (1.0e-05) \\
 &  & m0 & \texttt{ e} (9.0e-06), \texttt{ J} (9.0e-06), \texttt{ G} (9.0e-06) \\
 &  & a0.h10 & \texttt{ what} (7.0e-06), \texttt{ that} (7.0e-06), \texttt{is} (7.0e-06) \\
 &  & a0.h15 & \texttt{ one} (7.0e-06), \texttt{e} (7.0e-06), \texttt{ the} (7.0e-06) \\
 &  & a0.h3 & \texttt{-} (7.0e-06), \texttt{ in} (7.0e-06), \texttt{ on} (7.0e-06) \\
 &  & input & \texttt{ } (7.0e-06), \texttt{Increment} (7.0e-06), \texttt{1} (7.0e-06) \\
\addlinespace
 & \textbf{LLaMA-2} & m1 & \texttt{sierp} (0.896), \texttt{Unterscheidung} (0.074), \texttt{kwiet} (0.027) \\
 &  & m24 & \texttt{them} (0.009), \texttt{ihnen} (0.004), \texttt{they} (0.003) \\
 &  & m28 & \texttt{—} (0.006), \texttt{,} (0.005), \texttt{–} (0.004) \\
 &  & m14 & \texttt{/-} (7.7e-04), \texttt{-+} (2.2e-04), \texttt{ỳ} (2.0e-04) \\
 &  & m15 & \texttt{by} (2.7e-04), \texttt{look} (2.1e-04), \texttt{>} (1.5e-04) \\
 &  & m13 & \texttt{Halle} (1.7e-04), \texttt{Hook} (1.2e-04), \texttt{wa} (1.1e-04) \\
 &  & m11 & \texttt{attan} (1.4e-04), \texttt{iore} (1.1e-04), \texttt{(U+0442)(U+043A)(U+0443)} (1.0e-04) \\
 &  & m12 & \texttt{ieder} (1.1e-04), \texttt{ikai} (1.1e-04), \texttt{sail} (1.1e-04) \\
 &  & m7 & \texttt{bek} (9.7e-05), \texttt{untime} (7.7e-05), \texttt{mina} (7.5e-05) \\
 &  & m9 & \texttt{opsis} (8.7e-05), \texttt{PO} (8.2e-05), \texttt{zug} (7.7e-05) \\
 &  & m10 & \texttt{ador} (8.4e-05), \texttt{rok} (8.3e-05), \texttt{keit} (8.3e-05) \\
 &  & m8 & \texttt{concrete} (8.3e-05), \texttt{OST} (8.1e-05), \texttt{Chor} (8.0e-05) \\
 &  & m6 & \texttt{idenote} (7.2e-05), \texttt{ischof} (7.1e-05), \texttt{asm} (7.1e-05) \\
 &  & m0 & \texttt{bolds} (6.6e-05), \texttt{sce} (6.0e-05), \texttt{hina} (5.7e-05) \\
 &  & m5 & \texttt{kop} (6.6e-05), \texttt{Ő} (6.3e-05), \texttt{Sug} (6.2e-05) \\
 &  & m2 & \texttt{nobody} (6.5e-05), \texttt{nahm} (6.0e-05), \texttt{everybody} (6.0e-05) \\
 &  & m3 & \texttt{uche} (5.2e-05), \texttt{Chronology} (5.1e-05), \texttt{emer} (4.9e-05) \\
 &  & a12.h5 & \texttt{extension} (3.9e-05), \texttt{oba} (3.9e-05), \texttt{extensions} (3.9e-05) \\
 &  & a5.h15 & \texttt{Campbell} (3.8e-05), \texttt{beskre} (3.7e-05), \texttt{:(U+2009)} (3.7e-05) \\
 &  & input & \texttt{/-} (3.7e-05), \texttt{igny} (3.6e-05), \texttt{Extern} (3.6e-05) \\
 &  & a1.h22 & \texttt{(U+045A)y} (3.4e-05), \texttt{(U+4E0B)} (3.4e-05), \texttt{unci} (3.4e-05) \\
 &  & a12.h22 & \texttt{tel} (3.4e-05), \texttt{} (3.3e-05), \texttt{guez} (3.3e-05) \\
 &  & a2.h2 & \texttt{zik} (3.2e-05), \texttt{Muse} (3.2e-05), \texttt{èn} (3.2e-05) \\
\addlinespace
 & \textbf{Qwen} & m26 & \texttt{(U+6027)(U+4EF7)} (1.000), \texttt{B} (1.000), \texttt{ \&} (0.999) \\
 &  & m27 & \texttt{Human} (1.000), \texttt{\^{}K} (1.000), \texttt{ derive} (0.999) \\
 &  & m24 & \texttt{ (U+62EC)} (0.973), \texttt{(U+5973)(U+6027)(U+670B)(U+53CB)} (0.955), \texttt{.ImageAlign} (0.914) \\
 &  & m21 & \texttt{aeda} (0.608), \texttt{ so} (0.599), \texttt{ to} (0.306) \\
 &  & a26.h24 & \texttt{ A} (0.032), \texttt{A} (0.006), \texttt{(G} (0.004) \\
 &  & a26.h23 & \texttt{ make} (0.008), \texttt{(make} (0.008), \texttt{.make} (0.006) \\
 &  & a26.h26 & \texttt{-await} (4.0e-04), \texttt{XMLElement} (3.2e-04), \texttt{etail} (2.7e-04) \\
 &  & m0 & \texttt{ fkk} (3.5e-04), \texttt{ libertine} (2.8e-04), \texttt{[];\textbackslash{}n} (2.3e-04) \\
 &  & a26.h25 & \texttt{ (U+81EA)(U+52A8)(U+751F)(U+6210)} (3.1e-04), \texttt{/Dk} (2.5e-04), \texttt{line} (2.4e-04) \\
 &  & a26.h22 & \texttt{(U+5341)(U+56DB)} (1.9e-04), \texttt{(U+5341)(U+4E09)} (1.8e-04), \texttt{(U+80B2)(U+4EBA)} (1.6e-04) \\
 &  & a25.h12 & \texttt{4} (1.5e-04), \texttt{5} (1.4e-04), \texttt{ Fifth} (1.2e-04) \\
 &  & a23.h19 & \texttt{ Five} (7.1e-05), \texttt{ five} (5.0e-05), \texttt{5} (4.1e-05) \\
 &  & a24.h23 & \texttt{num} (2.5e-05), \texttt{ Gall} (2.5e-05), \texttt{	num} (2.4e-05) \\
 &  & a0.h3 & \texttt{teenth} (2.4e-05), \texttt{ bénéficie} (1.9e-05), \texttt{ Noticed} (1.9e-05) \\
 &  & a23.h11 & \texttt{...";\textbackslash{}n} (1.3e-05), \texttt{\#ac} (1.2e-05), \texttt{ąż} (1.2e-05) \\
 &  & input & \texttt{(U+304D)(U+3061)(U+3093)} (9.0e-06), \texttt{(U+4E26)(U+4E14)} (9.0e-06), \texttt{(U+6362)(U+53E5)(U+8BDD)} (9.0e-06) \\
\midrule
\textbf{IOI} & \textbf{DiffuLLaMA} & m1 & \texttt{sierp} (0.550), \texttt{kwiet} (0.155), \texttt{Hinweis} (0.032) \\
 &  & m31 & \texttt{in} (0.204), \texttt{to} (0.044), \texttt{\textbackslash{}n} (0.039) \\
 &  & a23.h20 & \texttt{Sarah} (3.8e-04), \texttt{Vir} (1.1e-04), \texttt{sar} (1.1e-04) \\
 &  & a28.h7 & \texttt{V} (2.9e-04), \texttt{K} (2.0e-04), \texttt{Kim} (2.0e-04) \\
 &  & a26.h14 & \texttt{David} (1.7e-04), \texttt{David} (1.4e-04), \texttt{dav} (1.2e-04) \\
 &  & a30.h12 & \texttt{William} (1.3e-04), \texttt{Will} (1.2e-04), \texttt{Will} (1.1e-04) \\
 &  & m4 & \texttt{disambiguation} (1.1e-04), \texttt{printStackTrace} (9.8e-05), \texttt{-} (9.5e-05) \\
 &  & m2 & \texttt{nahm} (5.5e-05), \texttt{-} (5.5e-05), \texttt{Hinweis} (5.4e-05) \\
 &  & a27.h29 & \texttt{urn} (5.4e-05), \texttt{urr} (5.3e-05), \texttt{enis} (4.9e-05) \\
 &  & a18.h9 & \texttt{owo} (5.1e-05), \texttt{ceu} (4.3e-05), \texttt{fen} (4.3e-05) \\
 &  & m3 & \texttt{(U+4F1D)} (3.5e-05), \texttt{tu} (3.5e-05), \texttt{adr} (3.5e-05) \\
 &  & a5.h15 & \texttt{temps} (3.4e-05), \texttt{vend} (3.4e-05), \texttt{cancel} (3.4e-05) \\
 &  & a3.h26 & \texttt{lex} (3.3e-05), \texttt{ongodb} (3.3e-05), \texttt{huvudstaden} (3.3e-05) \\
 &  & input & \texttt{ô} (3.3e-05), \texttt{Horn} (3.3e-05), \texttt{roid} (3.3e-05) \\
 &  & m0 & \texttt{Einzeln} (3.3e-05), \texttt{(U+800C)} (3.3e-05), \texttt{atri} (3.3e-05) \\
 &  & a1.h1 & \texttt{(U+4EA4)} (3.2e-05), \texttt{gate} (3.2e-05), \texttt{Moc} (3.2e-05) \\
 &  & a1.h18 & \texttt{} (3.2e-05), \texttt{jsp} (3.2e-05), \texttt{epen} (3.2e-05) \\
\addlinespace
 & \textbf{Dream} & m21 & \texttt{pliers} (1.4e-05), \texttt{Tap} (1.4e-05), \texttt{ynom} (1.3e-05) \\
 &  & m13 & \texttt{ doen} (1.1e-05), \texttt{ bourgeois} (1.1e-05), \texttt{ upholstery} (1.0e-05) \\
 &  & m8 & \texttt{ wooded} (1.1e-05), \texttt{ curt} (1.0e-05), \texttt{ Genius} (1.0e-05) \\
 &  & m9 & \texttt{ melodies} (1.1e-05), \texttt{ interpolate} (1.0e-05), \texttt{ Infantry} (1.0e-05) \\
 &  & m10 & \texttt{ forestry} (1.0e-05), \texttt{ rhet} (1.0e-05), \texttt{ doen} (1.0e-05) \\
 &  & m15 & \texttt{orestation} (1.0e-05), \texttt{ secluded} (9.0e-06), \texttt{ cosy} (9.0e-06) \\
 &  & m18 & \texttt{ Races} (1.0e-05), \texttt{weets} (9.0e-06), \texttt{ife} (9.0e-06) \\
 &  & m19 & \texttt{ adjud} (1.0e-05), \texttt{ enchanted} (1.0e-05), \texttt{ instantiate} (1.0e-05) \\
 &  & m20 & \texttt{ blot} (1.0e-05), \texttt{ oval} (1.0e-05), \texttt{ blinking} (1.0e-05) \\
 &  & m7 & \texttt{ glimps} (1.0e-05), \texttt{ seeding} (1.0e-05), \texttt{ sadd} (1.0e-05) \\
 &  & m11 & \texttt{Tweet} (9.0e-06), \texttt{ Intr} (9.0e-06), \texttt{ enchanted} (8.0e-06) \\
 &  & m12 & \texttt{ milit} (9.0e-06), \texttt{ bourgeois} (9.0e-06), \texttt{slashes} (9.0e-06) \\
 &  & m14 & \texttt{ enam} (9.0e-06), \texttt{ upholstery} (9.0e-06), \texttt{ vener} (9.0e-06) \\
 &  & m16 & \texttt{ lan} (9.0e-06), \texttt{part} (9.0e-06), \texttt{ocal} (9.0e-06) \\
 &  & m17 & \texttt{ Israelis} (9.0e-06), \texttt{ commemor} (9.0e-06), \texttt{ driv} (9.0e-06) \\
 &  & a23.h10 & \texttt{ } (8.0e-06), \texttt{-} (8.0e-06), \texttt{0} (8.0e-06) \\
 &  & a24.h24 & \texttt{i} (8.0e-06), \texttt{ in} (8.0e-06), \texttt{ on} (8.0e-06) \\
 &  & a15.h20 & \texttt{'s} (7.0e-06), \texttt{ home} (7.0e-06), \texttt{ half} (7.0e-06) \\
 &  & a15.h23 & \texttt{ doen} (7.0e-06), \texttt{ Packages} (7.0e-06), \texttt{ classy} (7.0e-06) \\
 &  & a18.h25 & \texttt{ropy} (7.0e-06), \texttt{liner} (7.0e-06), \texttt{Bio} (7.0e-06) \\
 &  & a25.h24 & \texttt{ mailed} (7.0e-06), \texttt{RSS} (7.0e-06), \texttt{ masturbating} (7.0e-06) \\
\addlinespace
 & \textbf{LLaMA-2} & m1 & \texttt{sierp} (0.865), \texttt{Unterscheidung} (0.110), \texttt{kwiet} (0.022) \\
 &  & m27 & \texttt{too} (0.394), \texttt{her} (0.042), \texttt{e} (0.031) \\
 &  & a26.h21 & \texttt{Marian} (0.076), \texttt{Pat} (0.008), \texttt{Anne} (0.008) \\
 &  & a25.h0 & \texttt{Richard} (0.050), \texttt{William} (0.035), \texttt{David} (0.033) \\
 &  & a24.h3 & \texttt{Rosa} (0.007), \texttt{Williams} (6.4e-04), \texttt{Alice} (6.1e-04) \\
 &  & a20.h8 & \texttt{Susan} (0.004), \texttt{sus} (4.6e-04), \texttt{suspect} (1.4e-04) \\
 &  & a23.h20 & \texttt{Sarah} (0.004), \texttt{Vir} (0.001), \texttt{vir} (0.001) \\
 &  & a21.h30 & \texttt{Lee} (6.1e-04), \texttt{Kelly} (3.3e-04), \texttt{ee} (1.5e-04) \\
 &  & m4 & \texttt{vy} (2.1e-04), \texttt{disambiguation} (2.1e-04), \texttt{-} (2.0e-04) \\
 &  & a27.h29 & \texttt{arta} (1.3e-04), \texttt{ML} (1.3e-04), \texttt{ignon} (1.3e-04) \\
 &  & a18.h9 & \texttt{Blue} (9.5e-05), \texttt{cyk} (9.2e-05), \texttt{nja} (8.8e-05) \\
 &  & m0 & \texttt{bolds} (6.4e-05), \texttt{sce} (6.0e-05), \texttt{partiellement} (5.6e-05) \\
 &  & m2 & \texttt{nobody} (6.4e-05), \texttt{nahm} (5.9e-05), \texttt{everybody} (5.9e-05) \\
 &  & m3 & \texttt{ime} (5.4e-05), \texttt{(U+82B1)} (5.2e-05), \texttt{ña} (5.1e-05) \\
 &  & input & \texttt{ny} (4.0e-05), \texttt{ten} (4.0e-05), \texttt{eral} (3.9e-05) \\
 &  & a5.h15 & \texttt{Chronology} (3.7e-05), \texttt{://} (3.6e-05), \texttt{Extern} (3.6e-05) \\
 &  & a3.h26 & \texttt{erea} (3.6e-05), \texttt{ząt} (3.6e-05), \texttt{Songs} (3.6e-05) \\
 &  & a1.h1 & \texttt{(U+4EA4)} (3.4e-05), \texttt{Indep} (3.3e-05), \texttt{gate} (3.3e-05) \\
 &  & a1.h18 & \texttt{Bek} (3.3e-05), \texttt{arguments} (3.3e-05), \texttt{Millionen} (3.3e-05) \\
\addlinespace
 & \textbf{Qwen} & m27 & \texttt{Human} (1.000), \texttt{ Rossi} (0.990), \texttt{ ``} (0.978) \\
 &  & m25 & \texttt{ Alexander} (1.000), \texttt{ shall} (0.986), \texttt{zá} (0.970) \\
 &  & m24 & \texttt{ court} (0.983), \texttt{(U+5973)(U+6027)(U+670B)(U+53CB)} (0.955), \texttt{\})();\textbackslash{}n} (0.941) \\
 &  & m22 & \texttt{thought} (0.958), \texttt{ during} (0.921), \texttt{term} (0.914) \\
 &  & m20 & \texttt{(U+6027)(U+4EF7)} (0.893), \texttt{",\_\_} (0.247), \texttt{ynos} (0.101) \\
 &  & a27.h17 & \texttt{ Christina} (0.444), \texttt{ Jessica} (0.339), \texttt{ Crystal} (0.330) \\
 &  & a27.h18 & \texttt{ Lisa} (0.418), \texttt{ Elizabeth} (0.282), \texttt{Nic} (0.228) \\
 &  & a27.h1 & \texttt{ Jamie} (0.400), \texttt{ Nathan} (0.345), \texttt{ Mary} (0.275) \\
 &  & a27.h21 & \texttt{ Amy} (0.360), \texttt{ Amber} (0.331), \texttt{ Adam} (0.331) \\
 &  & a26.h5 & \texttt{ Jesse} (0.344), \texttt{Nich} (0.217), \texttt{ Rebecca} (0.203) \\
 &  & a27.h3 & \texttt{ Katie} (0.305), \texttt{ Ken} (0.295), \texttt{ Brittany} (0.189) \\
 &  & a27.h14 & \texttt{ Heather} (0.252), \texttt{ Steven} (0.244), \texttt{ Sean} (0.223) \\
 &  & a27.h24 & \texttt{ Scott} (0.233), \texttt{Brad} (0.221), \texttt{ Kris} (0.220) \\
 &  & a26.h2 & \texttt{Brad} (0.227), \texttt{ Megan} (0.194), \texttt{ brand} (0.180) \\
 &  & a27.h4 & \texttt{ Mary} (0.224), \texttt{Ben} (0.223), \texttt{ Mark} (0.212) \\
 &  & a26.h15 & \texttt{ Danielle} (0.215), \texttt{ Alicia} (0.182), \texttt{ Dustin} (0.176) \\
 &  & a24.h23 & \texttt{ John} (0.013), \texttt{ Thomas} (0.013), \texttt{ Kenneth} (0.008) \\
 &  & a25.h24 & \texttt{ ch} (0.002), \texttt{ William} (0.001), \texttt{	w} (0.001) \\
 &  & a24.h24 & \texttt{ Gad} (0.001), \texttt{ogen} (9.4e-04), \texttt{ Aqu} (8.7e-04) \\
 &  & a26.h22 & \texttt{.AppSettings} (6.5e-04), \texttt{ azt} (6.0e-04), \texttt{.d} (4.3e-04) \\
 &  & a26.h26 & \texttt{(U+0625)(U+0639)(U+062F)(U+0627)(U+062F)} (1.9e-04), \texttt{.TRAILING} (1.8e-04), \texttt{inx} (1.8e-04) \\
 &  & a18.h25 & \texttt{\_locator} (8.1e-05), \texttt{="'.} (8.1e-05), \texttt{.instrument} (7.8e-05) \\
 &  & a20.h24 & \texttt{setChecked} (7.5e-05), \texttt{anmar} (6.5e-05), \texttt{CAF} (6.4e-05) \\
 &  & a18.h27 & \texttt{(U+0623)(U+063A)(U+0644)(U+0628)} (4.6e-05), \texttt{ dealloc} (4.5e-05), \texttt{(U+FFFD)(U+FFFD)} (4.4e-05) \\
 &  & a17.h24 & \texttt{.setCharacter} (3.0e-05), \texttt{-urlencoded} (2.8e-05), \texttt{entious} (2.7e-05) \\
 &  & input & \texttt{(U+304D)(U+3061)(U+3093)} (9.0e-06), \texttt{(U+6362)(U+53E5)(U+8BDD)} (9.0e-06), \texttt{(U+4E26)(U+4E14)} (9.0e-06) \\
\bottomrule
\end{longtable}

\begin{table*}[t]
\centering
\footnotesize
\begin{tabular}{lllccl}
\toprule
\textbf{Task} & \textbf{Model} & \textbf{Component} & \textbf{Mean Logit} & \textbf{Top Tokens} & \textbf{Role} \\
\midrule
\textsc{GT} & Qwen & m26 & 198.04 & \texttt{2}, \texttt{1}, \texttt{(U+6027)(U+4EF7)} & Digit-related component \\
& Qwen & m27 & 143.05 & \texttt{nineteen}, \texttt{Kremlin}, \texttt{2} & General lexical component \\
& LLaMA & m31 & 23.97 & \texttt{roughly}, \texttt{Sunday}, \texttt{1} & General lexical component \\
& LLaMA & m30 & 17.35 & \texttt{1}, \texttt{(U+042A)}, \texttt{I} & Proper-noun-related component \\
& Dream & m27 & 53.85 & \texttt{doen} & General lexical component \\
& Dream & m26 & 7.56 & \texttt{doen}, \texttt{6}, \texttt{9} & Digit-related component \\
& DiffuLLaMA & m1 & 70.80 & \texttt{sierp}, \texttt{kwiet}, \texttt{Hinweis} & General lexical component \\
\midrule
\textsc{SI} & Qwen & m26 & 176.40 & \texttt{S}, \texttt{(U+6027)(U+4EF7)}, \texttt{H} & Proper-noun-related component \\
& Qwen & m27 & 138.79 & \texttt{pedest}, \texttt{Human}, \texttt{Jeremy} & Person-name-related component \\
& LLaMA & m31 & 23.13 & \texttt{Tags}, \texttt{Royal}, \texttt{ibile} & Person-name-related component \\
& LLaMA & m30 & 17.11 & \texttt{and}, \texttt{(U+042A)}, \texttt{the} & General lexical component \\
& Dream & m27 & 31.43 & \texttt{1} & Digit-related component \\
& Dream & m26 & 9.54 & \texttt{courteous}, \texttt{renewables}, \texttt{chops} & General lexical component \\
& DiffuLLaMA & m1 & 65.30 & \texttt{sierp}, \texttt{kwiet}, \texttt{Hinweis} & General lexical component \\
\bottomrule
\end{tabular}
\caption{Representative high-attribution components for \textsc{GT} and \textsc{SI} from the completed Top-100 component-filtered logit-lens runs. Non-ASCII tokens are rendered as Unicode code-point sequences to avoid LaTeX encoding ambiguity. Roles are descriptive labels summarizing the observed top-token patterns rather than definitive functional labels.}
\label{tab:component-roles-gt-si}
\end{table*}

\paragraph{Expanded GT/SI logit-lens metrics.}
The following table aggregates the completed Top-100 component-filtered logit-lens outputs for \textsc{GT} and \textsc{SI}, including both Dream and DiffuLLaMA. Token strings in the representative-token appendix are rendered with Unicode code-point escapes when needed; this quantitative table contains only scalar summaries.

\begin{table*}[t]
\centering
\small
\caption{Component-level logit-lens metrics on \textsc{GT} and \textsc{SI}, aggregated over all Top-100 components per model. 
\textsc{NameFrac}@10 is the fraction of name-like tokens among the top-10 aligned tokens; \textsc{LogitGap}, entropy, and \(\Delta\mathrm{LME}\) are averaged, not median, across components. 
Across both regimes, MDMs, especially Dream and DiffuLLaMA, show smaller \textsc{LogitGap} and higher entropy than Qwen, consistent with the reduced single-component dominance reported in the main text.}
\label{tab:gt_si_logitlens_metrics}
\begin{tabular}{llrrrr}
\toprule
\textbf{Task} & \textbf{Model} & \textbf{NameFrac@10} & \textbf{LogitGap} & \textbf{Entropy} & \textbf{$\Delta$LME}\\
\midrule
\multirow{4}{*}{\textsc{GT}}
& Qwen & 0.0642 & 3.1070 & 2.1713 & -0.1456 \\
& LLaMA & 0.1323 & 0.1058 & 2.2628 & -0.0299 \\
& Dream & 0.1449 & 0.0852 & 2.2884 & -0.0430 \\
& DiffuLLaMA & 0.1210 & 0.0374 & 2.2722 & -0.0510 \\
\midrule
\multirow{4}{*}{\textsc{SI}}
& Qwen & 0.0565 & 3.0503 & 2.1502 & -0.2015 \\
& LLaMA & 0.1447 & 0.1001 & 2.2670 & -0.0342 \\
& Dream & 0.1650 & 0.0423 & 2.2891 & -0.0047 \\
& DiffuLLaMA & 0.1450 & 0.0759 & 2.2633 & -0.1433 \\
\bottomrule
\end{tabular}
\end{table*}

\section{Quantitative Mechanism Metrics \& Results}
\label{sec:metrics}

To formalize specialization and dominance in our component-level logit lens analysis, and to provide objective support for our qualitative interpretations, we introduce several quantitative probes. Let $l_i(c)$ denote the logit assigned by component $c$ to token $i$, and let $T_K(c)$ denote the set of top-$K$ tokens ranked by logit. 

\subsection{Semantic Alignment and Dominance Probes}

\paragraph{Name Alignment Frequency (NameFrac@K):}
To quantitatively assess whether semantic alignment extends beyond the top-1 token, we measure the fraction of person-name tokens among the top-$K$ aligned tokens for each component:
\begin{equation}
    \text{NameFrac}@K(c) = \frac{|T_K(c) \cap \mathcal{N}|}{|T_K(c)|}
\end{equation}
where $\mathcal{N}$ denotes the set of person-name tokens, identified using a pretrained BERT-based Named Entity Recognition (NER) model (filtered for the \texttt{PERSON} label).

\paragraph{Selective Amplification (Logit Gap):}
To quantify whether a component strongly favors a single token, we compute the logit gap between the highest and second-highest logits. Because softmax probabilities depend exponentially on logit differences, larger logit gaps indicate stronger selective amplification:
\begin{equation}
    \text{LogitGap}(c) = l_{(1)}(c) - l_{(2)}(c)
\end{equation}
where $l_{(1)}$ and $l_{(2)}$ represent the maximum and second-maximum logits, respectively.

\paragraph{Log-Mean-Exp Dominance Gap ($\Delta$LME):}
Because frequency alone does not measure predictive dominance, we quantify whether name tokens exert stronger influence when they appear by computing a log-mean-exp dominance gap between name and non-name tokens within the top-$K$ set:
\begin{equation}
    \Delta\text{LME}(c) = \text{LME}(T_K(c) \cap \mathcal{N}) - \text{LME}(T_K(c) \setminus \mathcal{N})
\end{equation}
where the LME for a subset of tokens $S$ is defined as:
\begin{equation}
    \text{LME}(S) = \log \left( \frac{1}{|S|} \sum_{i \in S} \exp(l_i(c)) \right)
\end{equation}
This metric controls for group size and directly measures selective amplification.

\paragraph{Distributional Sharpness (Entropy):}
To formalize whether semantic alignment is concentrated or distributed, we compute the entropy over the normalized top-$K$ logits:
\begin{equation}
    H(c) = - \sum_{i \in T_K(c)} p_i \log p_i, \quad \text{where} \quad p_i = \frac{\exp(l_i(c))}{\sum_{j \in T_K(c)} \exp(l_j(c))}
\end{equation}
Lower entropy indicates sharper specialization (typical of autoregressive models), while higher entropy indicates more distributed alignment (typical of masked diffusion models).

\subsection{Architectural Depth Metric}

\paragraph{Center of Gravity (CoG):}
To provide an objective, scalar measure of the ``Mechanism Shift'' independent of graph layout and visual density, we calculate the Center of Gravity (CoG). The CoG represents the attribution-weighted average layer index of the discovered circuit, identifying where the core computations are localized:
\begin{equation}
    \text{CoG} = \frac{\sum_l l \cdot A_l}{\sum_l A_l}
\end{equation}
where $l$ is the layer index and $A_l$ is the sum of EAP-IG attribution scores for all components in layer $l$.

The tables below provide the empirical results derived from the metrics defined above. We observe that autoregressive models exhibit larger logit gaps (e.g., Qwen: 2.12; LLaMA: 0.70) than their masked diffusion counterparts (DiffuLLaMA: 0.49), indicating a stronger concentration of probability mass on a single dominant token. Furthermore, DiffuLLaMA exhibits higher entropy (2.01) than autoregressive models (LLaMA: 1.81; Qwen: 0.97), indicating more distributed semantic alignment. Together, these metrics demonstrate the differences in specialization, selective amplification, and architectural depth between ARMs and MDMs.

\begin{table}[h]
\centering
\small
\caption{Component-level dominance, logit gap, and entropy metrics on \textsc{IOI}. 
In contrast to Table~\ref{tab:gt_si_logitlens_metrics}, values are reported as medians over Top-100 components, following the protocol used for the \textsc{IOI} analysis. 
Name Token Proportion (Top-1) is the fraction of components whose top-1 aligned token is a person name. 
Although MDMs occasionally show a higher raw proportion of name tokens, ARMs exhibit substantially stronger selective amplification, with higher \(\Delta\mathrm{LME}\) and \textsc{LogitGap}, and sharper specialization, with lower entropy. 
Dashes indicate metrics not computed in the original \textsc{IOI} run.}
\label{tab:dominance_metrics}
\begin{tabular}{lcccc}
\toprule
\textbf{Model} & \textbf{Name Token Proportion (Top-1)} & \textbf{Median $\Delta$LME} & \textbf{Logit Gap} & \textbf{Entropy} \\
\midrule
Dream & 19.7\% & 0.31 & -- & -- \\
DiffuLLaMA & 25.2\% & 0.05 & 0.49 & 2.01 \\
Qwen & 6.0\% & 0.90 & 2.12 & 0.97 \\
LLaMA & 13.4\% & 0.89 & 0.70 & 1.81 \\
\bottomrule
\end{tabular}
\end{table}

\begin{table}[h]
\centering
\small
\caption{Center of Gravity (CoG) for IOI and Countdown tasks. The sharp drop in CoG for Dream on the Countdown task provides quantitative validation of the mechanism shift (front-loading) into earlier layers during global reasoning.}
\label{tab:cog_results}
\begin{tabular}{llcc}
\toprule
\textbf{Task} & \textbf{Model} & \textbf{CoG (Layer Index)} & \textbf{Relative CoG (0--1)} \\
\midrule
\multirow{2}{*}{\textsc{IOI}} & Qwen2.5-7B & 17.528 & 0.548 \\
 & Dream-Base-7B & 20.432 & 0.638 \\
\midrule
\multirow{2}{*}{\textsc{Countdown}} & Qwen2.5-7B & 16.582 & 0.518 \\
 & Dream-Base-7B & 4.856 & 0.152 \\
\bottomrule
\end{tabular}
\end{table}

\twocolumn

\end{document}